\documentclass[10pt,twocolumn,letterpaper]{article}

\usepackage{cvpr}              

\usepackage{graphicx}
\usepackage{amsmath}
\usepackage{amssymb}
\usepackage{booktabs}

\usepackage[pagebackref,breaklinks,colorlinks]{hyperref}

\usepackage{subcaption}
\usepackage{multirow}
\usepackage{ltablex}
\usepackage{makecell}
\usepackage{nicematrix}
\usepackage{enumitem}
\usepackage{xspace}
\usepackage{array}
\usepackage{verbatim}
\usepackage{color, colortbl}
\usepackage{pdflscape}

\usepackage{xparse}
\usepackage{etoolbox}
\usepackage{pgfplots,xifthen,bm,tikz,tikz-cd,adjustbox,afterpage,algorithmic}
\usetikzlibrary{calc, positioning}
\usepackage{stackengine}
\usepackage{arydshln}
\usepackage{euscript}
\DeclareMathAlphabet{\mathpzc}{T1}{pzc}{m}{it}

\usepackage[ruled,vlined]{algorithm2e}
\usepackage{algorithmic}
\usepackage{amsmath}
\usepackage{amsfonts}
\SetAlFnt{\footnotesize}
\SetKwInput{KwInit}{Initialization}

\usepackage{tikz}
\usepackage{pgfplots}

\usepackage[capitalize]{cleveref}
\crefname{section}{Sec.}{Secs.}
\Crefname{section}{Section}{Sections}
\Crefname{table}{Table}{Tables}
\crefname{table}{Tab.}{Tabs.}

\def\cvprPaperID{10062} 
\def\confName{CVPR}
\def\confYear{2023}

\begin{document}

\title{Class-Incremental Mixture of Gaussians for Deep Continual Learning}

\author{Lukasz Korycki\\
Virginia Commonwealth University\\
{\tt\small koryckil@vcu.edu}
\and
Bartosz Krawczyk\\
Virginia Commonwealth University\\
{\tt\small bkrawczyk@vcu.edu}
}
\maketitle

\begin{abstract}
   Continual learning models for stationary data focus on learning and retaining concepts coming to them in a sequential manner. In the most generic class-incremental environment, we have to be ready to deal with classes coming one by one, without any higher-level grouping. This requirement invalidates many previously proposed methods and forces researchers to look for more flexible alternative approaches. In this work, we follow the idea of centroid-driven methods and propose end-to-end incorporation of the mixture of Gaussians model into the continual learning framework. By employing the gradient-based approach and designing losses capable of learning discriminative features while avoiding degenerate solutions, we successfully combine the mixture model with a deep feature extractor allowing for joint optimization and adjustments in the latent space. Additionally, we show that our model can effectively learn in memory-free scenarios with fixed extractors. In the conducted experiments, we empirically demonstrate the effectiveness of the proposed solutions and exhibit the competitiveness of our model when compared with state-of-the-art continual learning baselines evaluated in the context of image classification problems.
\end{abstract}

\section{Introduction}
\label{sec:intro}

While the initial research done in the domain of continual learning from stationary data was, in large part, oriented towards task-incremental solutions, more recent works attempt to address generalized cases consisting of purely class-incremental and data-incremental (also known as domain-incremental) settings \cite{Masana:2020, Ven:2019three}. These scenarios are usually more universal but also more challenging and restrictive mainly due to the lack of task or even class labels. Such settings make many of the previously proposed solutions practically useless, for example, the methods based on memory-free regularization \cite{Lange:2019}, which are not capable of discriminating between older and new classes, even if they address the catastrophic forgetting problem \cite{Maltoni:2019sit, Ven:2019three}. Although the most standard experience replay methods can be effectively applied in the class-incremental scenarios \cite{Buzzega2020:dark, Verwimp:2021}, there has been also a search for alternative approaches that could provide natural capabilities required for such cases. A significant group of methods can be identified based on their reliance on centroids (or prototypes) combined with the nearest-centroid classification methods \cite{Mai:2021contr}. Since centroids can be independently added to the classifier, they are examples of methods that can be very smoothly incorporated into class-incremental scenarios, offering almost no interference in the latent space.

In this work, we explore an advanced version of these alternatives by proposing integration of the gradient-based Gaussian mixture model with a class-incremental deep continual learning framework, called \textbf{MIX}. In fact, it requires us to tackle three major problems at the same time: (i) gradient-based mixture training, (ii) combining it with a trainable deep feature extractor and, finally, (iii) making it suitable for class-incremental scenarios. To achieve these goals, we introduce a set of dedicated losses, configurations and methods, providing a probabilistic classifier on top of a feature extractor and within a model capable of learning end-to-end. This opens many potential research directions that could exploit the well-modeled statistical properties of Gaussians. In addition to that, we show that our class-incremental mixture model, analogously to the centroid-driven algorithms, is characterized by some inherent properties useful in continual learning scenarios. They allow it for much better separation of concepts at the level of the classification module, leading to significant improvements in memory-free scenarios when pre-trained extractors are used. Through an extensive empirical study, we analyze different configurations of our method, provide the reader with some intuition about its parameters and show its competitiveness in the context of other continual learning algorithms.

\section{Related works}
\label{sec:related}

\paragraph{Continual learning:} In continual learning, our focus should be on effective incorporation of the arriving data and retention of the acquired knowledge \cite{Kemker:2018}. The main problem that learning algorithms will encounter here is catastrophic forgetting \cite{French:1999, Parisi:2019}. The most straightforward approaches involve replaying instances of previously seen tasks or classes while learning new ones \cite{Bang:2021, Hayes:2021rep, Rolnick:2019er, Verwimp:2021}. Instead of putting instance-level constraints on the learning directions, we can apply direct adjustments to the loss using dedicated regularization terms. The most commonly used approach involves utilizing the knowledge-distillation loss \cite{Hinton:2015} combined with standard cross-entropy \cite{Dhar:2019, Hou:2018, Li:2018} or maintaining importance weights to distinguish parameters that are crucial for the retention \cite{Aljundi:2018mas, Kirkpatrick:2017, Zenke:2017}. These methods generally cannot be used in more realistic class-incremental or data-incremental scenarios (if they do not use memory buffers), since they cannot learn how to discriminate new instances from the older ones \cite{Hsu:2018, Maltoni:2019sit, Ven:2019three}. Other approaches may employ masking to isolate parameters per task to keep them static when learning new ones \cite{Hung:2019compact, Mallya:2018pack, Mallya:2018piggy}, use dynamic structures to expand the network for new concepts \cite{Veniat:2021modular, Yan:2021der, Yoon:2018den}, utilize ensemble techniques \cite{Aljundi:2017expert, Rios:2020tmap, Wen:2020be} or meta-learning and hypernetworks \cite{Finn:2017maml, Javed:2019meta, Oswald:2019hyper}. Finally, interesting alternative approaches focus on hybridizing the neural networks with different machine learning methods, e.g. decision trees \cite{Korycki:2021lldt} or centroid-driven algorithms \cite{Mai:2021contr, Rebuffi:2017, Yu:2020sem}. The latter group of methods has been found especially useful in one-class-incremental scenarios, since, as mentioned in the introduction, centroids can be stored independently per class, allowing for natural class-incremental learning without additional interference at the level of a classifier. In this work, we follow these approaches and replace basic centroids learned separately from the feature extractor with more complex end-to-end mixture models.

\paragraph{Mixture optimization:} Various techniques can be applied for the task of fitting the mixture model to given data. The most standard approach utilizes the EM algorithm, which can be realized in both offline and online settings \cite{Engel:2010igmm, Melnykov:mix}. While EM provides a stable framework for learning the mixtures -- in terms of mathematical constraints and convergence -- it is critically limited when it comes to working with high-dimensional data and feasible memory consumption \cite{Gepperth2021:grad}. On top of that, this algorithm is intrinsically incapable of being fully integrated with neural networks, preventing it from achieving joint end-to-end deep learning and benefiting from dedicated features. An alternative approach involves gradient-based optimization \cite{Gepperth2021:grad}. This method has been proved to be able to provide more scalable and flexible algorithms capable of working in challenging scenarios with high-dimensional data and in online settings. Most importantly, the gradient-based approach naturally enables combining the model as a classifier with a trainable deep feature extractor \cite{Variani:2015joint}, allowing for extending the optimization process with the input space adjustments. Methods utilizing such a compound learning process showed much evidence of its usability in offline and unsupervised scenarios, while at the same time encouraging researchers to develop further extensions and improvements \cite{Gepperth2021:grad, Pfulb:2021:gmmcf}. Given all of the characteristics, we decided to use this approach in our scenario of continual learning.

\section{Mixture of Gaussians for Class-Incremental Learning}
\label{sec:igmm-cl-intro}

Formally, the general goal of our work is to incrementally learn a classification model defined as $\phi^{(t)}: \mathcal{X} \rightarrow \mathcal{C}$ that can effectively incorporate subsequent class batches $\langle (\bm{X}^{(1)}, c=1), (\bm{X}^{(2)}, c=2), ..., (\bm{X}^{(t)}, c=t)\rangle$, where $\bm{X}^{(t)}$ contains instances $\bm{x}$ only for a given class $c$. After $t$ classes the model $\phi^{(t)}$ should aim at minimizing the loss for the current class $c=t$ and all previously observed ones:
\begin{equation}
\bm{\mathcal{L}}^{(t)} = \sum_{c=1}^{t}\sum_{n=1}^{N_c}\mathcal{L}^{(c)}(\phi^{(t)}(\bm{x}_n^{(c)})),
\end{equation}

\noindent where $\bm{x}_n^{(c)} \in \bm{X}^{(c)}$ and $\mathcal{L}^{(c)}$ can be any supervised loss. 

Additionally, since we are interested in deep learning, we define the whole trainable model as a tuple $\phi^{(t)}=\langle\mathcal{F}^{(t)}, \mathcal{G}^{(t)}\rangle$ consisting of a feature extractor $\mathcal{F}^{(t)}$ and a classifier $\mathcal{G}^{(t)}$ jointly aggregating knowledge from $t$ classes. The model makes prediction by classifying the features provided from the extractor $\phi^{(t)}(\bm{x})=\mathcal{G}^{(t)}(\mathcal{F}^{(t)}(\bm{x}))=\mathcal{G}^{(t)}(\hat{\bm{x}})$. In this work, we aim at employing the mixture of Gaussians as a jointly trained incremental classifier. Although the model learns from dedicated features $\hat{\bm{x}}$, in the next section, we use $\bm{x}$ for the sake of simplicity of notation.

\subsection{Generic supervised mixture model}
\label{sec:igmm-generic}

Formally, in a standard unsupervised setting the density for a given point $\mathbf{x}$ can be expressed using a multivariate normal distribution defined as:
\begin{align}\label{eq:mult-gauss-full}
\mathcal{N}(\mathbf{x}|\bm{\mu}_k, \mathbf{\Sigma}_k) & = \frac{1}{\sqrt{(2\pi)^{D}|\mathbf{\Sigma}_k|}} \nonumber \\ 
& \times exp(-\frac{1}{2}(\mathbf{x}-\bm{\mu}_k)^T\mathbf{\Sigma}_k^{-1}(\mathbf{x}-\bm{\mu}_k)),
\end{align}

\noindent where $\mathbf{\mu}$ and $\mathbf{\Sigma}$ are its mean and covariance, and $D$ is the size of the input (number of dimensions). The Gaussian mixture models (GMM) have been designed to approximate more complex multivariate densities by decomposing them into K components:
\begin{equation}\label{eq:igmm-basic-density}
p(\mathbf{x}) = \sum^K_{k=1}\omega_k\mathcal{N}(\mathbf{x}|\bm{\mu}_k,\mathbf{\Sigma}_k),
\end{equation}

\noindent where each of them is defined using a single Gaussian defined above and $\omega_k$ are their weights. The combined model, equipped with more degrees of freedom, should be capable of providing more accurate expressions of the overall observed distributions than a simpler approach utilizing only a single component. In such a framework, the fitting of the mixture to given data $\bm{X}$ is based on minimizing the loss defined using the log-likelihood function:
\begin{align}\label{eq:igmm-basic-loss}
\bar{\mathcal{L}} & = -\log p(\bm{X}|\omega,\bm{\mu},\bm{\Sigma}) \nonumber \\ 
& = -\frac{1}{N}\sum^N_{n=1}\log\sum^K_{k=1}\omega_k
\mathcal{N}(\mathbf{x}_n|\bm{\mu}_k,\mathbf{\Sigma}_k),
\end{align}

\noindent where we adjust the free parameters of the model -- means $\bm{\mu}$, covariance matrices $\bm{\Sigma}$ and weights $\omega$. To adapt the given framework to supervised scenarios we can simply specify a separate mixture model for each class $c$:
\begin{equation}\label{eq:igmm-basic-cls-density}
p(\mathbf{x} | c) = \sum^K_{k=1}\omega_k^{(c)}\mathcal{N}(\mathbf{x}|\bm{\mu}_k^{(c)},\mathbf{\Sigma}_k^{(c)}),
\end{equation}

\noindent and focus on minimizing the aforementioned loss also per class $\bar{\mathcal{L}}^{(c)}$:
\begin{equation}\label{eq:igmm-full-class}
\hat{\mathcal{L}} = \sum_{c=1}^{C} \bar{\mathcal{L}}^{(c)} = -\log \sum_{c=1}^{C} p(\bm{X}^{(c)}|\omega^{(c)},\bm{\mu}^{(c)},\bm{\Sigma}^{(c)}),
\end{equation}

\noindent where $\bm{X}^{(c)}$ are $N_c$ class-specific observations.

In continual learning we should aim at minimizing the interference of current updates with previously created models to alleviate the detrimental effect of catastrophic forgetting. Therefore, it is worth mentioning here that GMMs create such an opportunity by allowing for maximizing the log-likelihood only for a currently learned class through $\bar{\mathcal{L}}^{(c)}$. It provides a perfect separation at the level of the classification model.

\subsection{Mixture optimization for class-incremental deep learning}

In order to apply gradient-based learning to GMM in class-incremental deep learning scenarios, we have to address several different issues. Some of them are common for all GMM models using gradient-based learning, while others are specific for the class-incremental deep learning settings.

In general, we say that our goal is to optimize the class-incremental joint model $\phi^{(t)}=\langle\mathcal{F}^{(t)}, \mathcal{G}^{(t)}\rangle$, defined at the beginning of Sec. \ref{sec:igmm-cl-intro}, using some supervised loss $\mathcal{L}$. Since we set $\mathcal{G}^{(t)} = \bm{\mathcal{N}}^{(t)}$, where $\bm{\mathcal{N}}^{(t)}$ is a whole GMM model, we have $\phi^{(t)}(\bm{x})=\bm{\mathcal{N}}^{(t)}(\mathcal{F}^{(t)}(\bm{x}))$. The trainable parameters are weights $\partial \mathcal{L} / \partial\bm{W}$ and biases $\partial \mathcal{L} / \partial\bm{b}$ for the extractor, and means $\partial \mathcal{L} / \partial\bm{\mu}$, covariance matrices $\partial \mathcal{L} / \partial\bm{\Sigma}$ and component weights $\partial \mathcal{L} / \partial\omega$ for the classifier. All of the subsequent paragraphs focus on designing optimization in the classifier (mixture) space, as it was introduced in Sec. \ref{sec:igmm-generic}.

\subsubsection{Loss design}
\label{sec:igmm-losses}

\paragraph{Max-component:} It has been shown that optimizing the full loss $\mathcal{\bar{L}}^{(c)}$ given in Eq. \ref{eq:igmm-basic-loss} may lead to some numerical instabilities, especially for high-dimensional data \cite{Gepperth2021:grad}. To address this issue a \textit{max-component approximation} can be used. This approach is very straightforward. Since all $p(\bm{x}|c,k)$ in Eq. \ref{eq:igmm-basic-cls-density} are positive, any component provides a lower bound for the whole sum used in $\mathcal{\bar{L}}^{(c)}$. If for every point $\bm{x}_n$ we find a component providing the highest log-likelihood and sum all of them, we will get the largest (max-component) lower bound \cite{Gepperth2021:grad}:
\begin{equation}\label{eq:igmm-max-bound}
\mathcal{L}^{(c)}_{max} = -\frac{1}{N_c}\sum^{N_c}_{n=1}\max_k\log(\omega_k^{(c)}
\mathcal{N}(\mathbf{x}^{(c)}_n|\bm{\mu}_k^{(c)},\mathbf{\Sigma}_k^{(c)})).
\end{equation}

\noindent Since we can state that $\mathcal{L}^{(c)}_{max} \geq \bar{\mathcal{L}}^{(c)}$, we are able to minimize $\bar{\mathcal{L}}^{(c)}$ by focusing only on $\mathcal{L}^{(c)}_{max}$. It is also worth mentioning that just like the general formula given in Eq. \ref{eq:igmm-full-class} may eliminate the interference with previously learned classes, the max-component approximation can limit the same issue at the level of class components, for example, in data-incremental scenarios \cite{Ven:2019three}, making this approach a natural candidate for continual learning settings.

\paragraph{Inter-contrastive loss:} All of the introduced losses are limited to scenarios either without a feature extractor or with a fixed pre-trained one. Unfortunately, if we operate in a setting where we can modify the input space of the mixture model and we utilize any of the aforementioned metrics relying entirely on maximizing log-likelihood, we will inevitably end up with a local minimum that for a joint model $\phi^{(t)}$ exists for $\forall\bm{x}(\mathcal{G}^{(t)}(\bm{x}) = \bm{0})$. This issue can be solved by incorporating an inter-contrastive loss that will distance representations for different classes. We define the loss as:
\begin{equation}
\mathcal{L}^{(c)}_{ie} = \frac{1}{N_c}\max_{j \neq c}\sum^{N_c}_{n=1}\max_{k}\log(\omega_k^{(j)}
\mathcal{N}(\mathbf{x}^{(c)}_n|\bm{\mu}_k^{(j)},\mathbf{\Sigma}_k^{(j)})),
\end{equation}

\noindent which boils down to finding the closest component in other classes, and then optimizing against the class that on average is the closest to the one currently being considered. We keep the log-likelihood to ensure a similar numerical space of loss values as the one for the positive part given in Eq. \ref{eq:igmm-max-bound}. However, now one should notice that minimizing such a loss may very easily destabilize learning since optimization will gravitate towards $\bar{\mathcal{L}}^{(c)}_{ie} \rightarrow -\infty$ preventing the model from actually fitting to the class examples. To avoid it we introduce a \textit{tightness bound} $\tau$ that clips the contrastive loss value at some pre-defined point $\mathcal{L}^{(c)}_{ie}(\tau) = \max(\tau, \mathcal{L}^{(c)}_{ie})$. This basically means that we stop the decrease of the contrastive loss below the given bound, allowing for a more significant contribution of the actual fitting part $\mathcal{L}^{(c)}_{max}$. We parametrize the $\tau$ value with a simple linear transformation $\tau = \bar{p}_{max}^{(c)} - 1/\tau_p$, where $\bar{p}_{max}^{(c)}$ is the average maximum density value observed across all class components (can be obtained on-the-fly) and $\tau_p$ is a tunable hyperparameter that takes values between $( 0,1 \rangle$. Such a loss can provide effective discrimination between components of different classes, as shown for an example in Appendix A.


\paragraph{Diverse components:} While all of the introduced techniques and modifications ensure reliable discrimination between components of different classes, they do not consider differentiation between components of the same class or their quality. In fact, even in offline gradient-driven settings without dynamic feature extraction it is common to obtain mixtures reduced to a single component per class with all the others practically meaningless, e.g., due to zeroed weights \cite{Gepperth2021:grad}. In scenarios with a trainable extractor, this problem becomes even more significant as it is very easy for the optimizer to focus on maximizing log-likelihood from a single component, as both mixture model and flexible extractor lack constraints to prevent this. While in standard scenarios this problem can be successfully addressed with a good initialization method, e.g., using k-means \cite{Su2007:init}, we observed that it was not enough in our case. As a consequence, we introduced two elements to the learning process.

\smallskip
\noindent\textbf{Regionalization} -- before learning each class, we first divide it into $K$ clusters using the k-means clustering. Then we force each component to fit only to the data from its cluster called a \textit{region} $\mathcal{R}^{(c)}_k$. This replaces the max-component loss $\mathcal{L}^{(c)}_{max}$ defined in Eq. \ref{eq:igmm-max-bound} with:
\begin{equation}
\mathcal{L}^{(c)}_{reg} = -\sum^K_{k=1}\frac{1}{N_k}\sum_{\bm{x} \in \mathcal{R}^{(c)}_k}\log(\omega_k^{(c)}
\mathcal{N}(\mathbf{x}|\bm{\mu}_k^{(c)},\mathbf{\Sigma}_k^{(c)})).
\end{equation}

\smallskip
\noindent\textbf{Intra-contrastive loss} -- the regionalization approach is necessary yet not sufficient to provide sufficient diversification between same-class components. The reason for it is the same as for discrimination between different classes, as described in the previous paragraph. Analogously to the inter-contrastive loss, we add the intra-contrastive loss with the tightness bound $\tau$:
\begin{align}
\mathcal{L}^{(c)}_{ia}(\tau) & = \sum^K_{k=1}\max( \tau, \max_{m \neq k} \frac{1}{N_k} \nonumber \\ 
& \times \sum_{\bm{x} \in \mathcal{R}_k}\log(\omega_m^{(c)}
\mathcal{N}(\mathbf{x}|\bm{\mu}_m^{(c)},\mathbf{\Sigma}_m^{(c)})),
\end{align}

\noindent which for each class region pushes away other same-class components that on average are closest to the currently considered one, based on the regionalization conducted in the previous step. Obviously, one can define separate $\tau$ for the inter- and intra-contrastive loss.
	
	
\noindent Such an approach can effectively increase the diversity of the same-class components, as given for an example in Appendix A. However, this approach imposes a hard constraint on how the representation and mixture may look, which limits the flexibility of the whole model. Regardless of these concerns, this method can still effectively improve the overall performance of a multi-component model over a method without the proposed improvement, as we will show in our extensive experiments.


\paragraph{Final component-based losses:} To summarize, we distinguish two component-based losses. One uses the max-component approach (\textbf{MC}):
\begin{equation}
\mathcal{L}_{mc} = \sum_{c=1}^{t}\mathcal{L}_{max}^{(c)} + \mathcal{L}_{ie}^{(c)}(\tau_{ie}),
\end{equation}

\noindent while the second loss adds the regionalization technique with the intra-contrastive part (\textbf{MCR}):
\begin{equation}
\mathcal{L}_{mcr} = \sum_{c=1}^{t}\mathcal{L}_{reg}^{(c)} + \beta(\mathcal{L}_{ie}^{(c)}(\tau_{ie}) + \mathcal{L}_{ia}^{(c)}(\tau_{ia})).
\end{equation}


\paragraph{Cross-entropy loss:} Last but not least, we can also attempt to directly optimize the whole standard loss $\hat{\mathcal{L}}$ given in Eq. \ref{eq:igmm-basic-loss}, using a high-level supervised wrapper loss, e.g., based on cross-entropy (\textbf{CE}). In such a case, our loss is defined as:
\begin{equation}
\mathcal{L}_{ce} = -\sum_{c=1}^{t}\sum_{n=1}^{N_c}\bm{y}_{n}^{(c)} \log \hat{y}_{n}^{(c)},
\end{equation}

\noindent where $\bm{y}$ is a one-hot target vector and $\hat{y}_{n}^{(c)}$ comes from the softmax function $\hat{y}_{n}^{(c)} = e^{p_{n}^{(c)}}/\sum_{c=1}^{t}e^{p_{n}^{(c)}}$ and $p_{n}^{(c)}=p(\bm{x}_n|c)$ is a density value for a given class produced by the mixture model accordingly to Eq. \ref{eq:igmm-basic-cls-density}.

\subsubsection{Constraints}
Other issues that have to be addressed when using gradient-based mixture training are the mathematical constraints that have to be enforced to preserve a valid mixture model. This is required since gradient-based learning does not constrain the possible values for means, covariance matrices and weights, and the last two have to remain in a specific range of values.

\paragraph{Component weights:} For the GMM model its component weights $\omega_k$ have to sum up to one: $\sum_{k=1}^{K}\omega_k=1$. To ensure that the effective weights satisfy this requirement we simply train auxiliary free parameters $\hat{\omega}_k$ and use the softmax-based normalization $\omega_k = e^{\hat{\omega}_k}/\sum_{j=1}^{K}e^{\hat{\omega_j}}$ to obtain required values \cite{Gepperth2021:grad, Hosseini:2015}.

\paragraph{Covariance matrices:} For a general case, the covariance matrices of the GMM model should be symmetric positive definite $\bm{v}^T\Sigma\bm{v} > 0$ for all nonzero vectors $\bm{v}$. This can be enforced using the Cholesky decomposition \cite{Higham:2009cholesky} $\bm{\Sigma} = \bm{A}\bm{A}^T$, where $\bm{A}$ is a triangular matrix with positive diagonal values $a_{ii} > 0$ and, at the same time, our trainable proxy parameter. To enforce positive diagonal values, after each gradient-based update we clamp them with $a_{ii} = min(a_{ii}, d_{min})$ using some predefined $d_{min}$ value. Finally, we also consider a case of a mixture using only the diagonal of the covariance -- variance $\bm{\sigma}$, which we control using the same clamp-based approach ${\sigma_{i} = min(\sigma_{i}, d_{min})}$.

\subsection{Memory buffer}
\label{sec:igmm-mem}

In our work, we consider the class-incremental scenario with strictly limited access to previously seen observations (classes). Therefore, in all of the introduced losses we use all available data for the currently learned class $t$, while for the others we sample from the memory buffers $\mathcal{M}_c$ that store an equal number of examples per each previously seen class. On the other hand, if the feature extractor is pre-trained and static we could remove the inter-contrastive loss and even get rid of the memory buffer, allowing for memory-free training, as we will show in our experimental study. The memory buffer is needed in a general case when we assume the joint training of the whole model.



\subsection{Classification}
\label{sec:igmm-class}

Finally, in the presented model, the classification of an instance $\bm{x}_n$ can be performed using two approaches, either utilizing the softmax function $\hat{y}_{n}^{(c)} = e^{p_{n}^{(c)}}/\sum_{c=1}^{t}e^{p_{n}^{(c)}}$, where $p_{n}^{(c)} = p(\bm{x}_n|c)$, or by taking the weighted support of the closest component $\hat{y}_{n}^{(c)} = \max_{k}\omega_k^{(c)}\mathcal{N}(\mathbf{x}_n|\bm{\mu}_k^{(c)},\mathbf{\Sigma}_k^{(c)})$. We will empirically show that these methods work best with specific losses designed in the previous sections.

\section{Experimental study}
\label{sec:igmm-exp}

In our experiments, we empirically explore all of the introduced methods and parameters and put our method in the performance context of different state-of-the-art baselines. We show how our model performs in end-to-end scenarios and with a pre-trained extractor, compared with other solutions. For more specific details regarding data, configurations and results, please refer to Appendix A and B, as well as to our repository containing source code for our methods and all experiments: (\textit{please check the source code provided in the supplementary materials, a public URL will be added later}). All of the experiments were conducted using 4 GPUs (Tesla V100) that were part of an internal cluster.

\subsection{Setup}
\label{sec:igmm-setup}

For the purpose of the evaluation we selected commonly utilized image classification datasets that were turned into class-incremental sequences by presenting their classes subsequently to the models \cite{Maltoni:2019sit, Masana:2020}. We used: MNIST, FASHION, SVHN, CIFAR and IMAGENET datasets using various variants (number of classes, pre-trained features). For the analysis of different configurations of our model we used shorter sequences. We extended them with the longer benchmarks for the comparison with baselines.

In the final section of this work, we compared our class-incremental Gaussian mixture model (\textbf{MIX-MCR}, \textbf{MIX-CE}) with other classifiers dedicated for continual learning scenarios. We considered: standard experience replay (\textbf{ER}) \cite{Korycki:2021ercd}, experience replay with subspaces (\textbf{ERSB}) \cite{Korycki:2021ercd}, centroid-based \textbf{iCaRL} \cite{Rebuffi:2017}, two gradient-based sample selection methods (\textbf{GSS} and \textbf{A-GEM}) \cite{Aljundi2019:gss, Chaudhry:2019}, experience replay combined with knowledge distillation and regularization (\textbf{DER}) \cite{Buzzega2020:dark}, and two purely regularization-based approaches -- \textbf{LWF} \cite{Li:2018} and \textbf{SI} \cite{Zenke:2017}. Most of the algorithms were implemented as wrappers of the source code provided in \cite{Buzzega2020:dark, Lomonaco2021:ava} under MIT License. For the last two we used their modifications adjusted for single-task learning \cite{Maltoni:2019sit}. As our lower bound we used a naively learning net (\textbf{NAIVE}), and for the upper bound we present results for the offline model (\textbf{OFFLINE}).

We evaluated the presented methods in a class-incremental setting, where all of the classes were presented to the models subsequently and were not shown again after their initial appearance. We measured the accuracy of a given algorithm after each class batch, utilizing holdout testing sets, and then, based on \cite{Kemker:2018}, used it to calculate the average incremental accuracy over the whole sequence:

\begin{equation}
\Omega_{all} = \frac{1}{T}\sum_{t=1}^{T}\alpha_t,
\end{equation}

\noindent where $\alpha_t$ is the model performance after $t$ classes and $T=C$ is the total number of classes. In addition to the whole aggregation, for the final comparison, we provided these values after each batch to present a more complete perspective of the obtained results.

\subsection{Results}
\label{sec:igmm-results}

In this section, we present and describe all of the results that were obtained for the experiments introduced in the previous paragraphs. The first part consists of the analysis of different configurations of MIX, while the second one focuses on a comparison with other class-incremental algorithms.

\paragraph{Loss and classification:}

\begin{figure*}[tb]
	\centering
	\begin{subfigure}{0.19\linewidth}
		\centering
		\includegraphics[width=\linewidth]{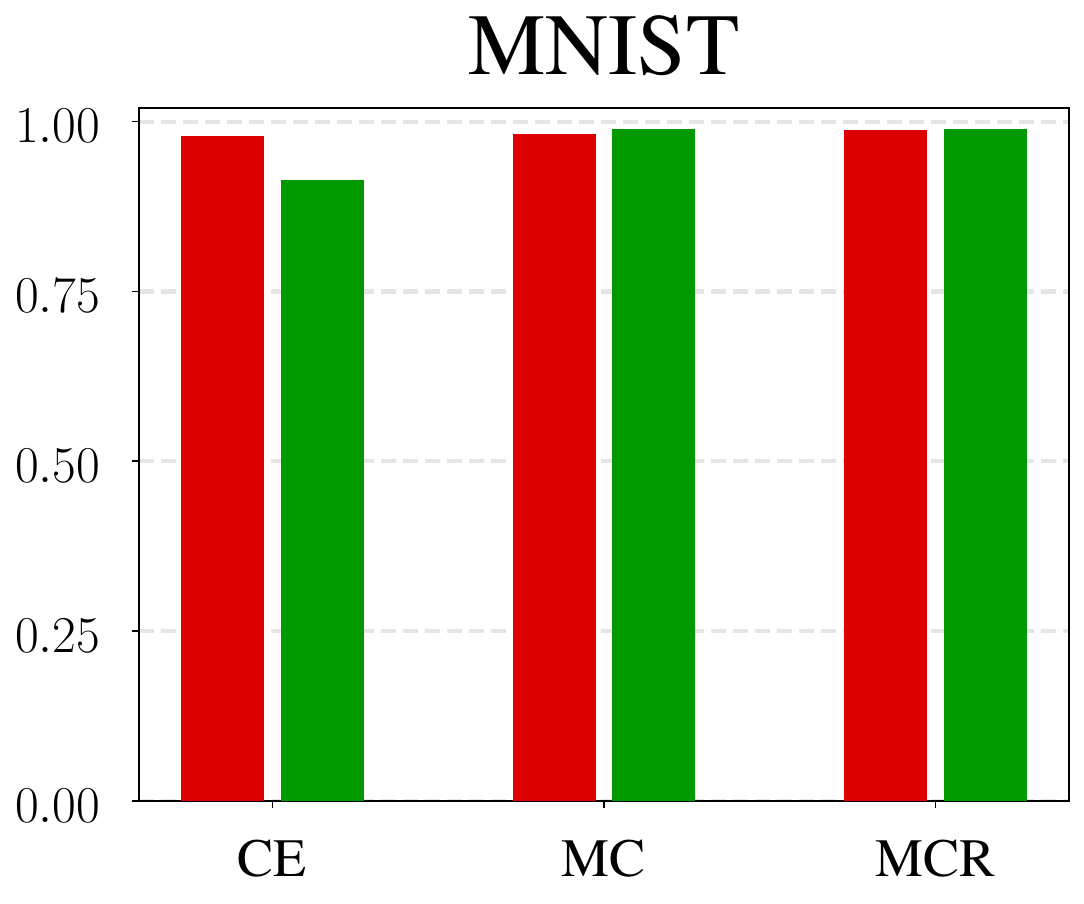}
	\end{subfigure}
	\begin{subfigure}{0.19\linewidth}
		\centering
		\includegraphics[width=\linewidth]{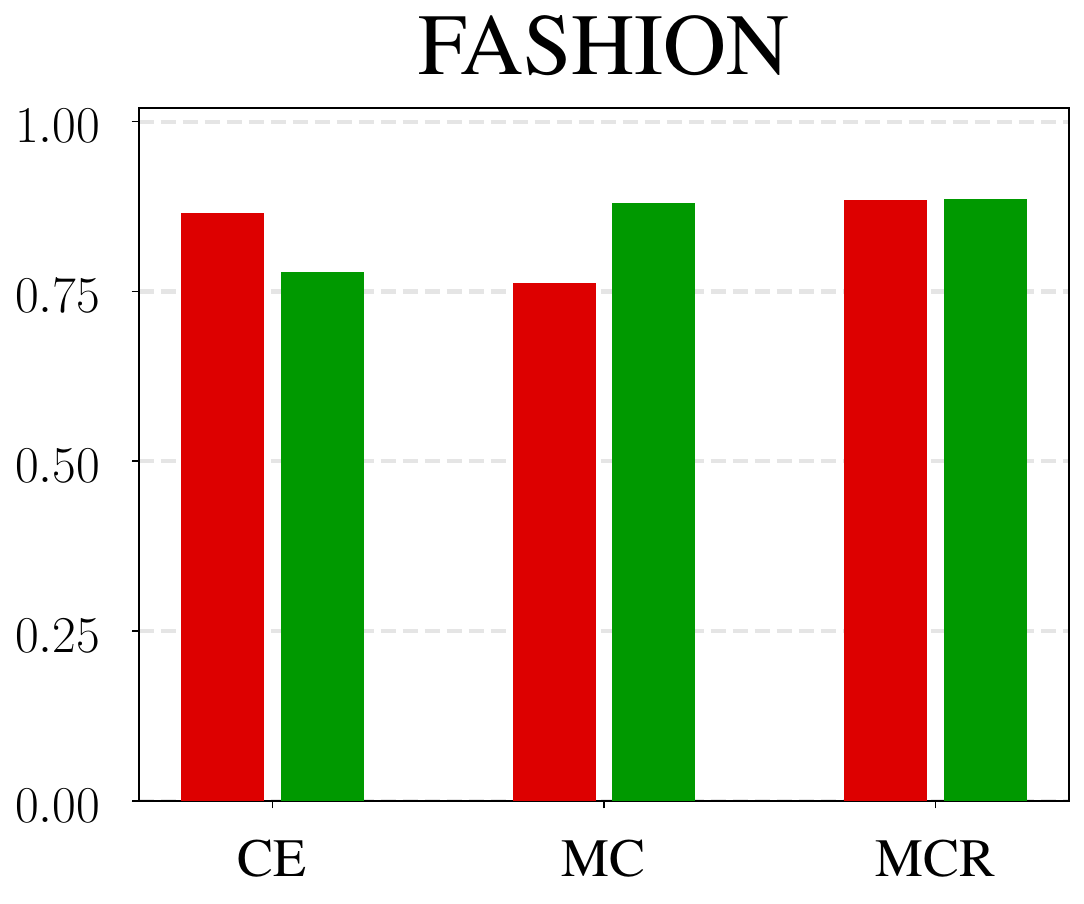}
	\end{subfigure}
	\begin{subfigure}{0.19\linewidth}
		\centering
		\includegraphics[width=\linewidth]{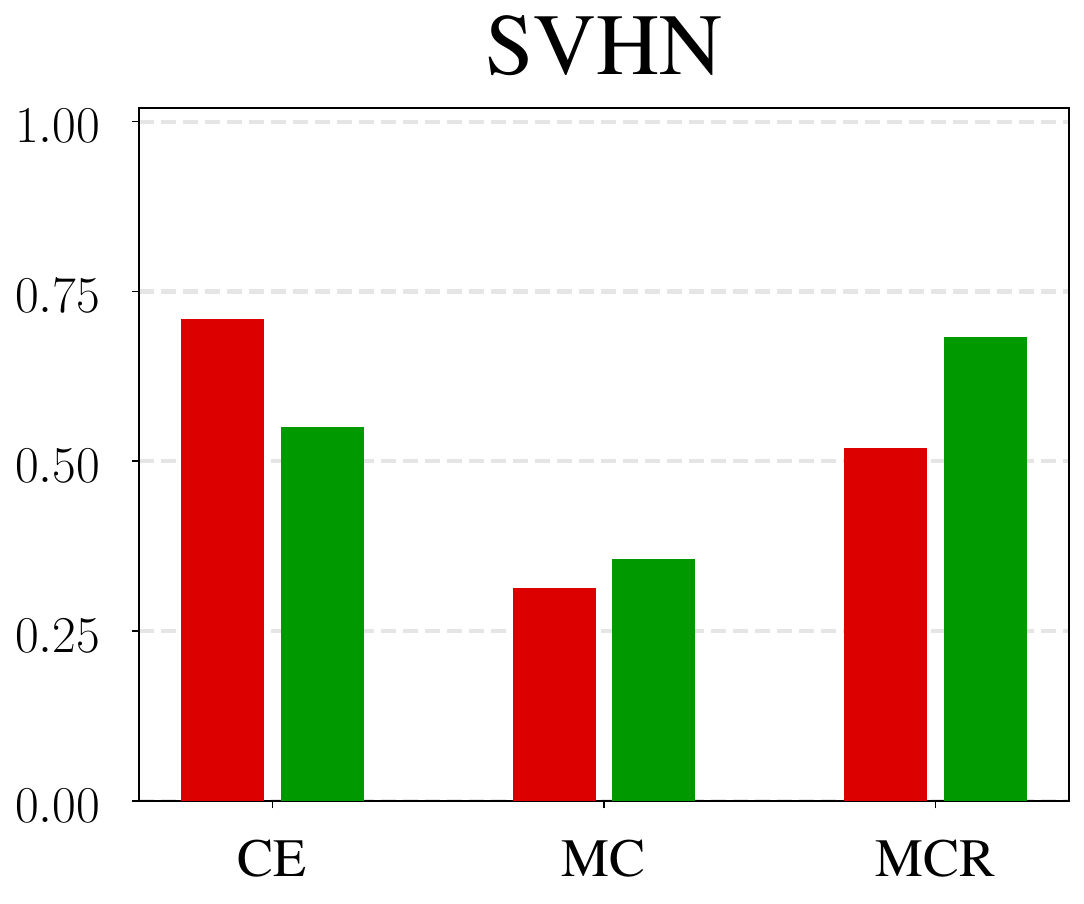}
	\end{subfigure}
	\begin{subfigure}{0.19\linewidth}
		\centering
		\includegraphics[width=\linewidth]{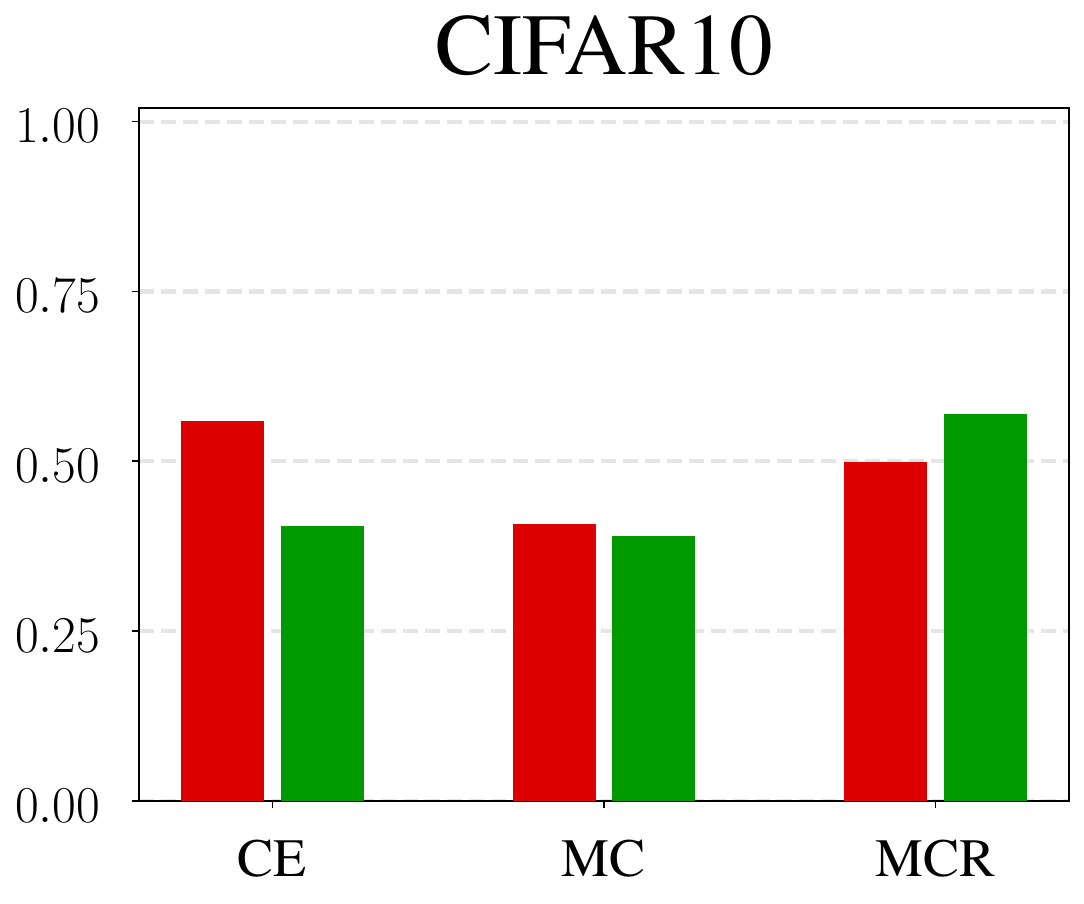}
	\end{subfigure}
	\begin{subfigure}{0.19\linewidth}
		\centering
		\includegraphics[width=\linewidth]{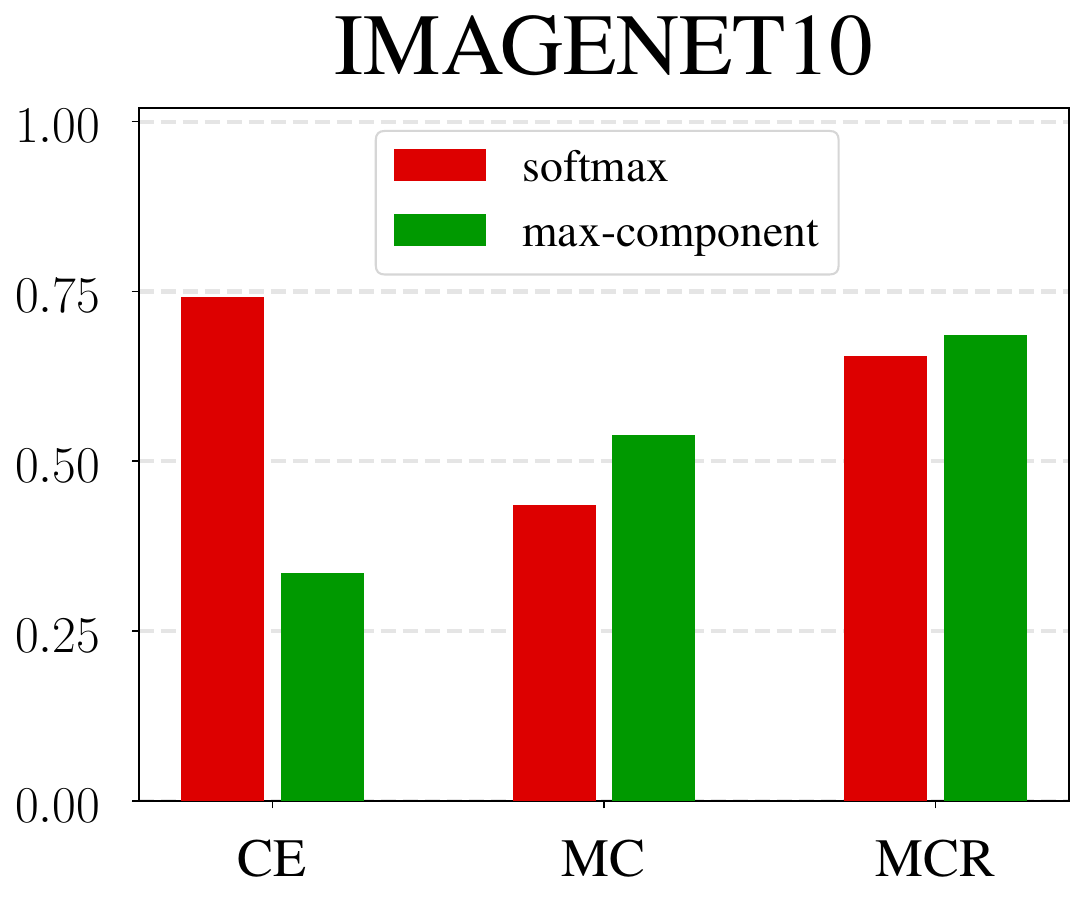}
	\end{subfigure}
	\caption[Average incremental accuracy for different losses combined with different classification methods]{Average incremental accuracy for different losses combined with different classification methods.}
	\label{fig:igmm-loss}
\end{figure*}

\begin{figure*}[tb]
	\centering
	\begin{subfigure}{0.19\linewidth}
		\centering
		\includegraphics[width=\linewidth]{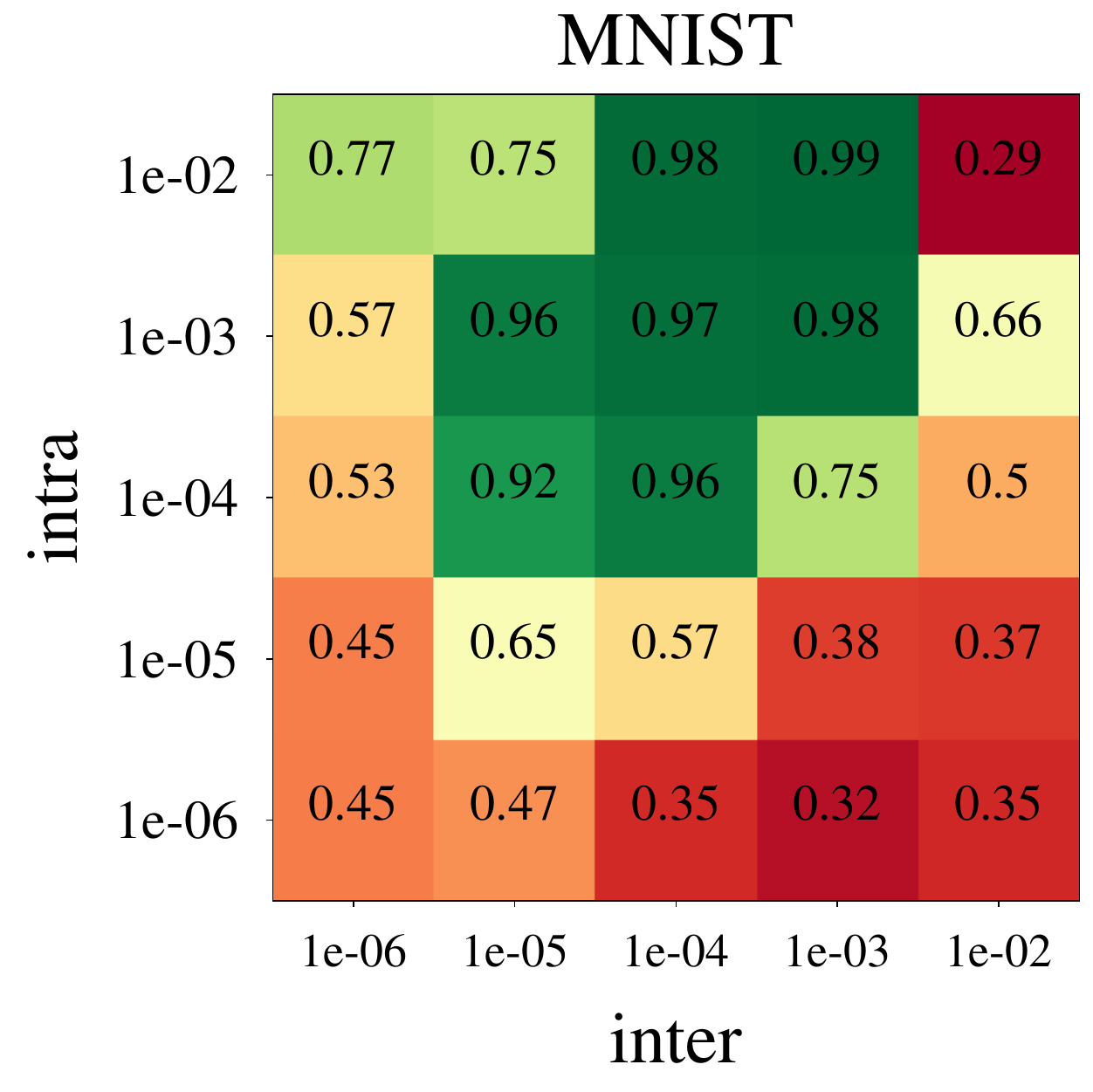}
	\end{subfigure}
	\begin{subfigure}{0.19\linewidth}
		\centering
		\includegraphics[width=\linewidth]{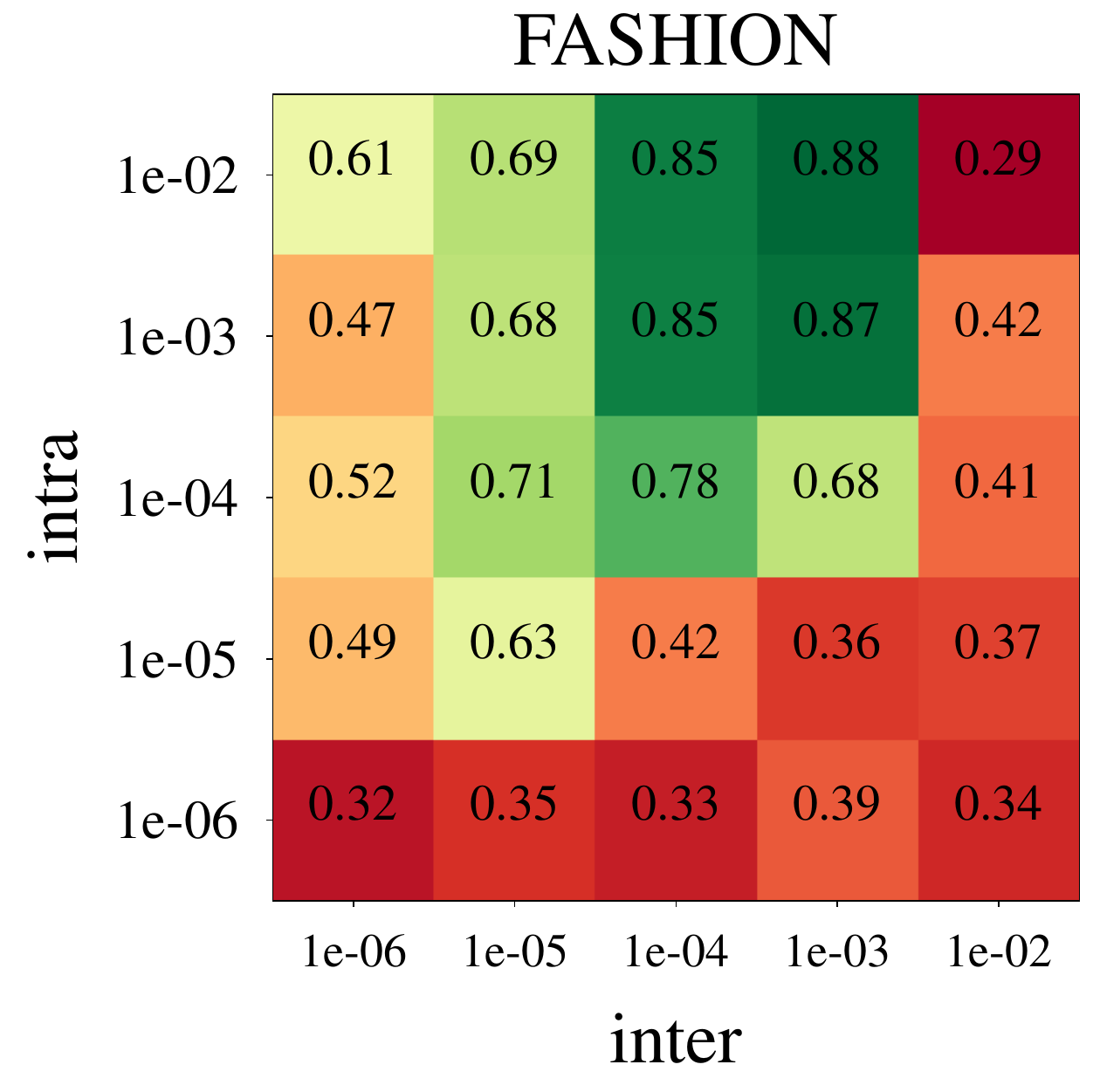}
	\end{subfigure}
	\begin{subfigure}{0.19\linewidth}
		\centering
		\includegraphics[width=\linewidth]{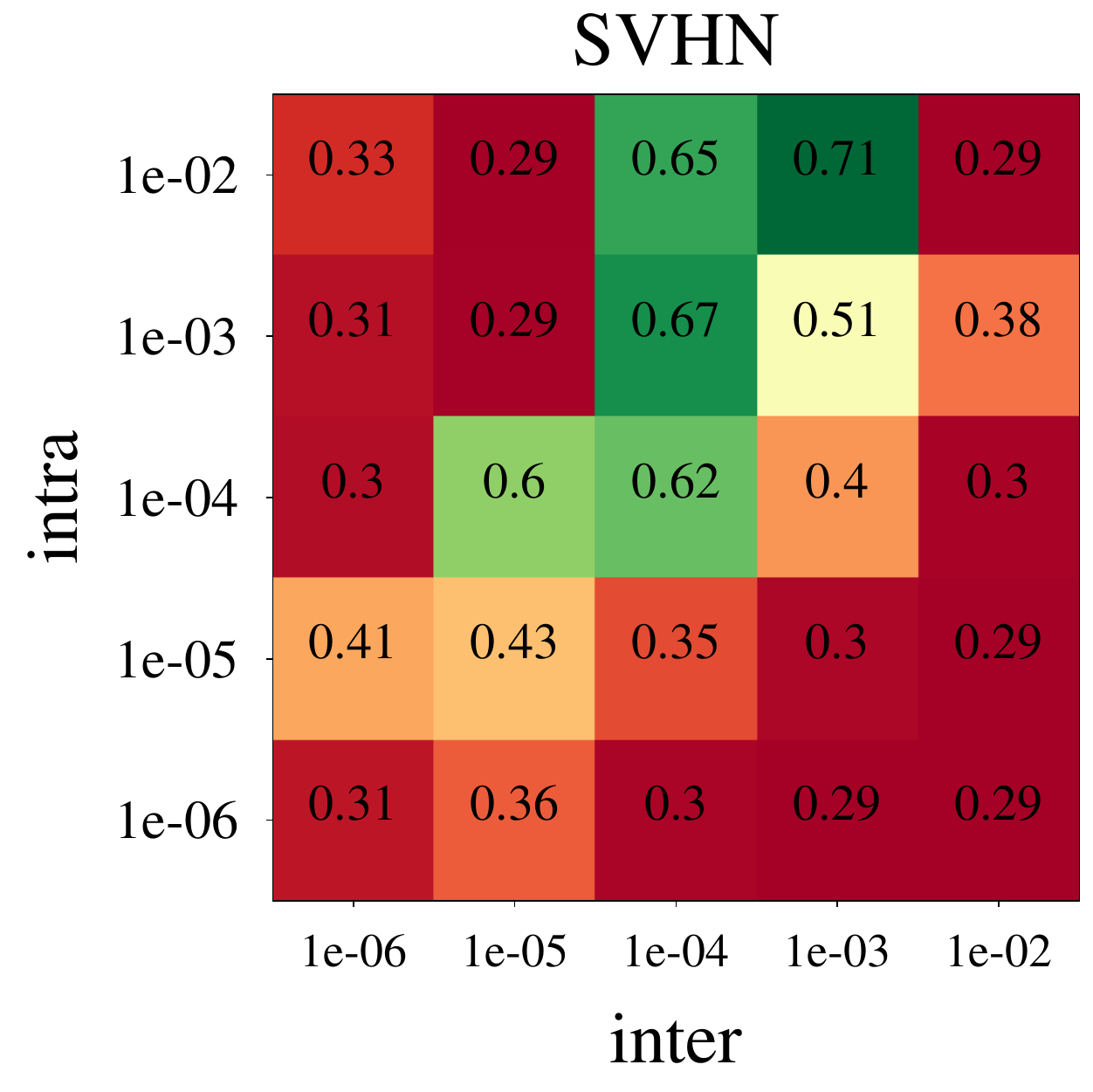}
	\end{subfigure}
	\begin{subfigure}{0.19\linewidth}
		\centering
		\includegraphics[width=\linewidth]{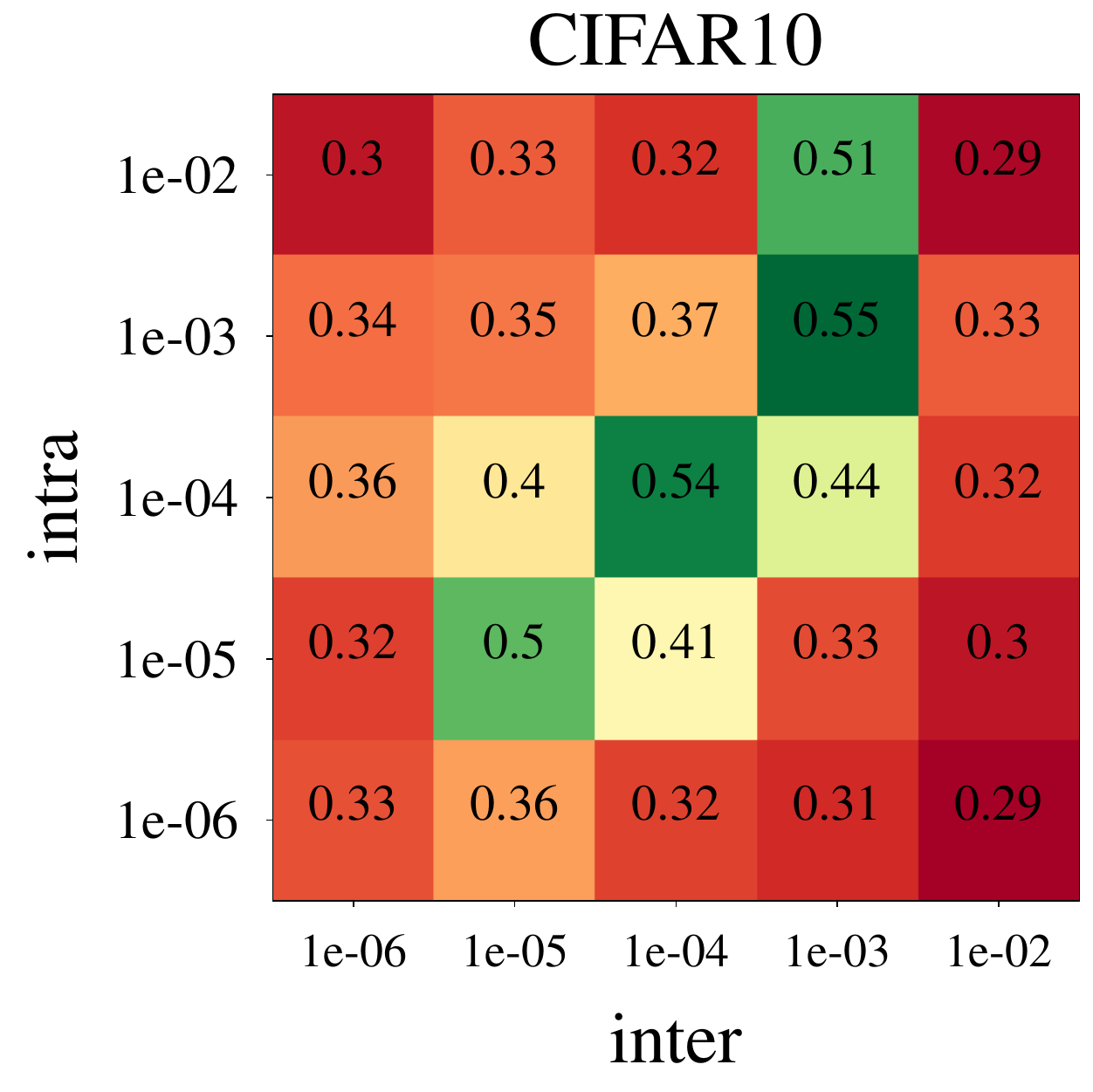}
	\end{subfigure}
	\begin{subfigure}{0.19\linewidth}
		\centering
		\includegraphics[width=\linewidth]{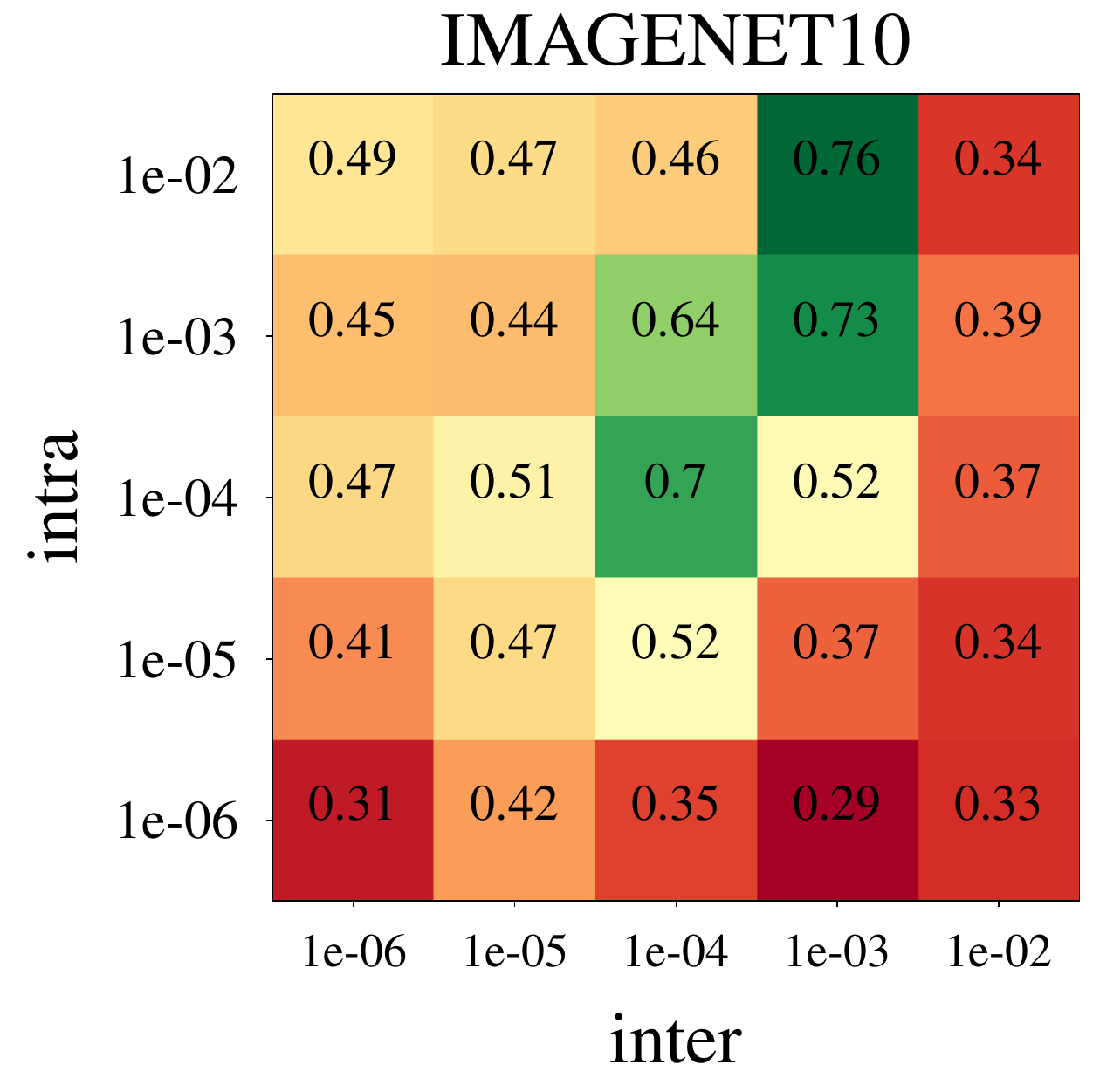}
	\end{subfigure}
	\caption[Average incremental accuracy for different combinations of the tightness parameter values (inter and intra)]{Average incremental accuracy for different combinations of the tightness parameter values (inter and intra).}
	\label{fig:igmm-tight}
\end{figure*}

We analyzed different combinations of the proposed losses and classification methods. Based on Fig. \ref{fig:igmm-loss}, we can make three major observations. Firstly, the softmax classification works significantly better with the CE loss, and max-component can be more efficiently paired with MC and MCR than softmax. It was evident for almost all cases (except for MC on CIFAR10) and resulted in almost 0.15 difference on average between softmax and max-component for CE, and about 0.05 for MC and MCR. 

Secondly, the MCR loss performed better than MC, showing consistent improvements, especially for more complex datasets like SVHN, CIFAR10 or IMAGENET10, which resulted in more than 0.1 for a difference on average. This demonstrate that the regionalization and intra-contrastive loss are capable of providing meaningful improvements over simpler MC loss utilizing only max-component and inter-contrastive elements, and that ensuring higher diversity among class components can be beneficial to the model. 

Finally, we can see that CE with softmax could provide very similar results as MCR with max-component, which means that the general GMM learning formula, wrapped with a high-level supervised loss, can be sometimes as useful as more complex MCR without the need for tuning additional parameters. One drawback of using CE, however, is the fact that it does not model the Gaussian mixtures well (see Appendix B for additional visualizations). The CE loss does not really have to fit the mixtures to the data since it is enough for it to ensure high classification quality. We can also observe a similar behavior for the MC loss. It may be prohibitive if one wants to obtain a reliable description of the latent space. The MCR loss achieves both objectives at the same time: high classification accuracy and high quality of the mixture models for features. This may be important if someone requires interpretable models or would like to extend the proposed algorithm with some Gaussian-oriented techniques that MCR may enable. Furthermore, we believe that analyzing its probabilistic properties in detail could be a part of incremental works built on top of the mixture model. They could utilize its well-defined characteristics, e.g. by proposing new mixture-based losses. 

\paragraph{Tightness:}

\begin{figure*}[tb]
	\centering
	\setlength{\fboxrule}{0.5pt}
	\setlength{\fboxsep}{0pt}
	\begin{subfigure}{0.15\linewidth}
		\centering
		\stackunder[5pt]{\fbox{\includegraphics[trim=76 76 70 70,clip,width=\linewidth]{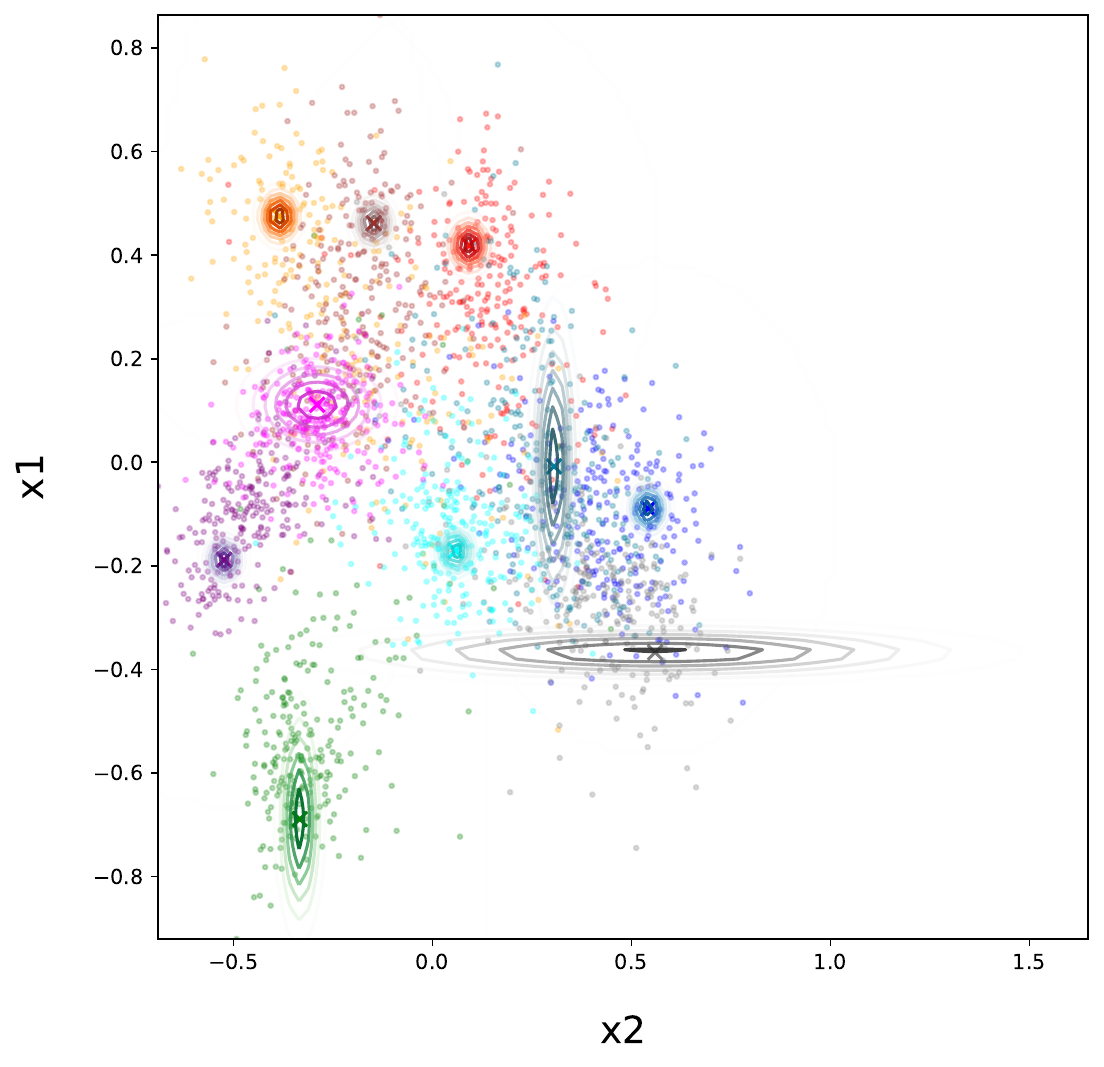}}}{0.05 $\rightarrow$ 0.55}
	\end{subfigure}\hspace{0.05cm}
	\begin{subfigure}{0.15\linewidth}
		\centering
		\stackunder[5pt]{\fbox{\includegraphics[trim=76 78 10 15,clip,width=\linewidth]{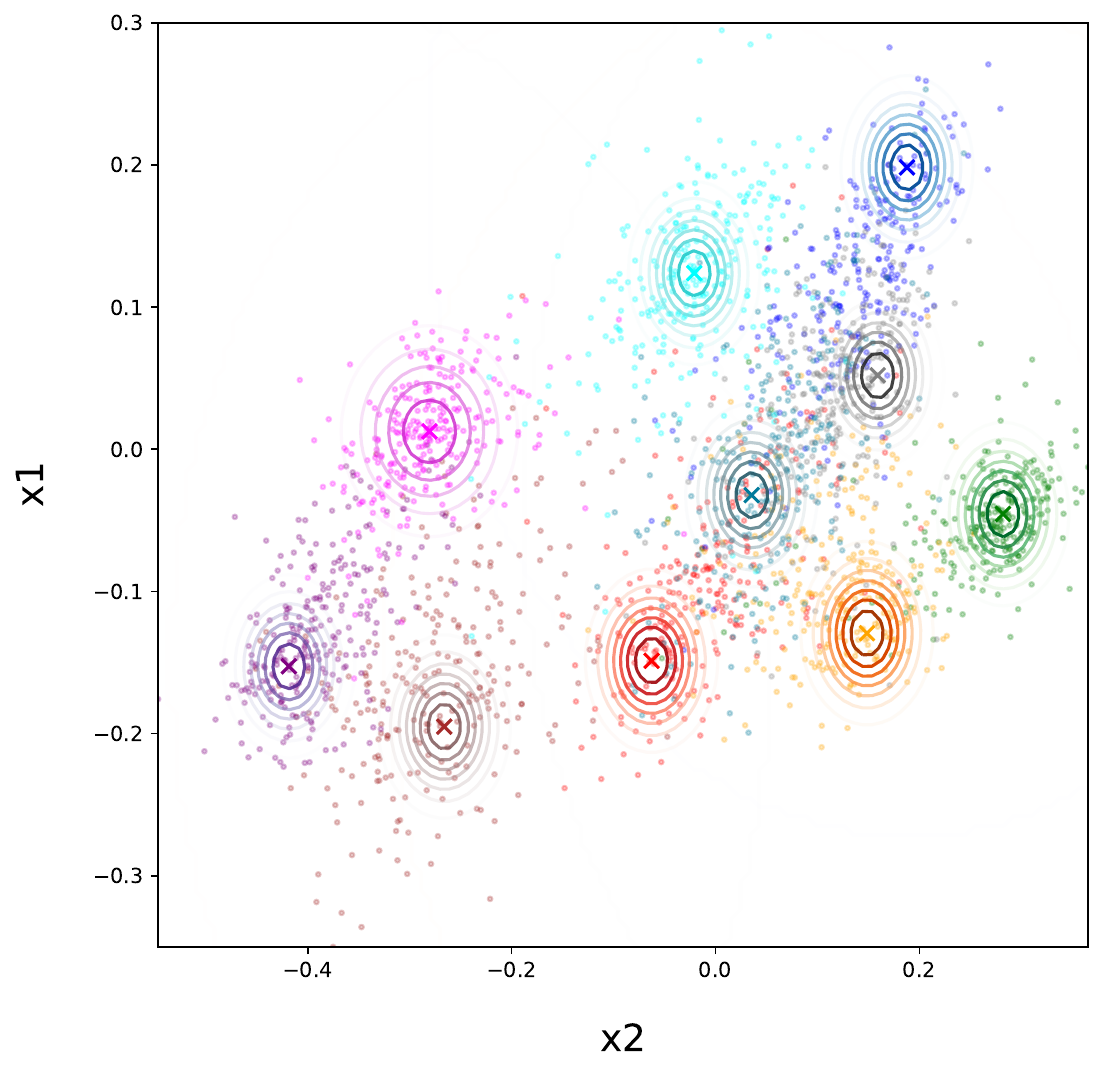}}}{0.1 $\rightarrow$ 0.69}
	\end{subfigure}\hspace{0.05cm}
	\begin{subfigure}{0.15\linewidth}
		\centering
		\stackunder[5pt]{\fbox{\includegraphics[trim=76 78 25 23,clip,width=\linewidth]{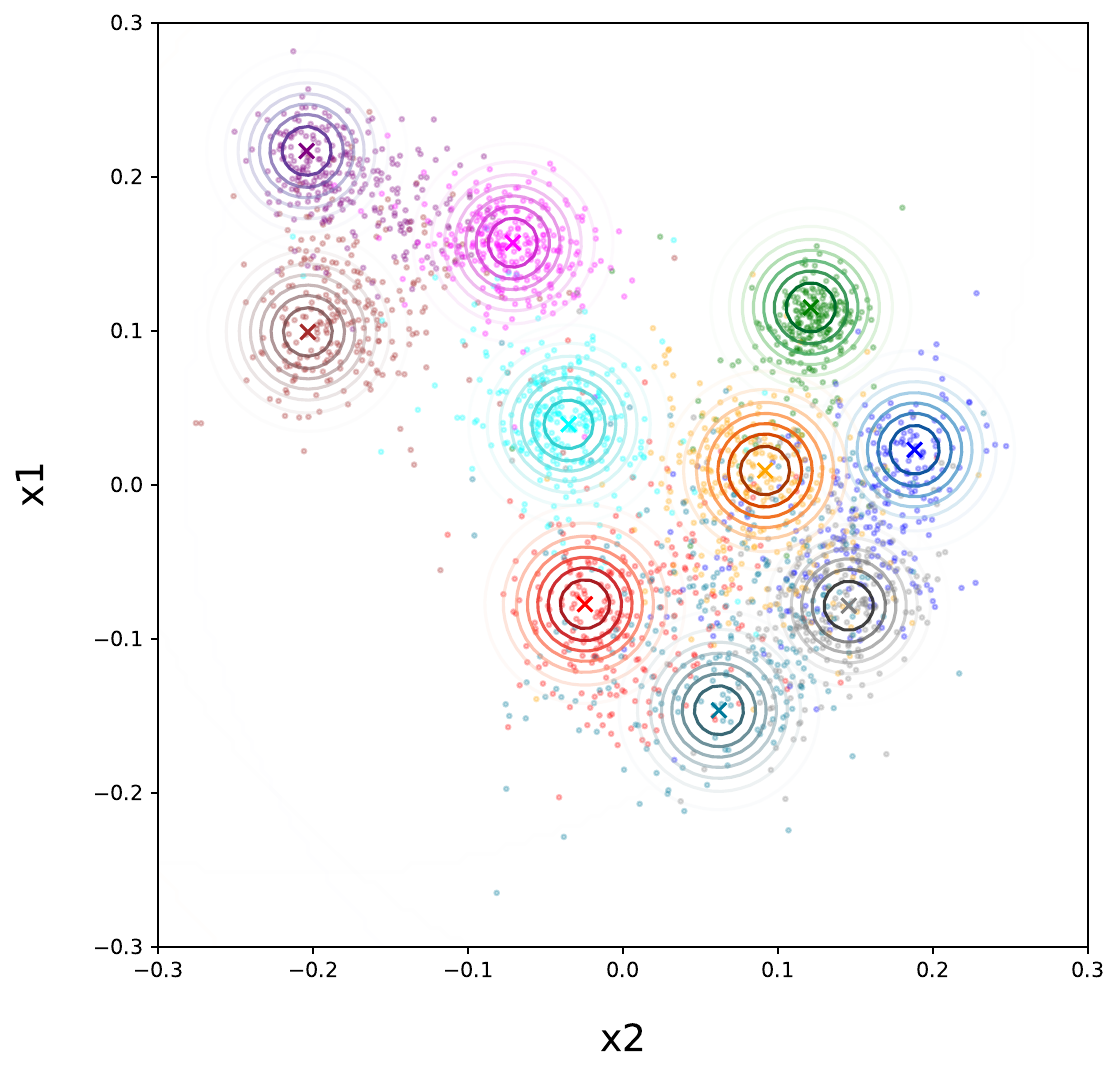}}}{0.2 $\rightarrow$ \textbf{0.72}}
	\end{subfigure}\hspace{0.05cm}
	\begin{subfigure}{0.15\linewidth}
		\centering
		\stackunder[5pt]{\fbox{\includegraphics[trim=140 140 80 75,clip,width=\linewidth]{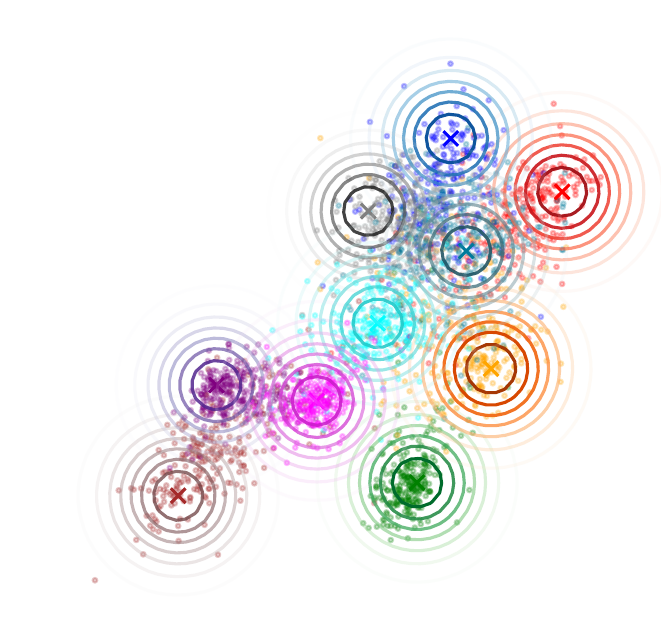}}}{0.5 $\rightarrow$ 0.7}
	\end{subfigure}\hspace{0.05cm}
	\begin{subfigure}{0.15\linewidth}
		\centering
		\stackunder[5pt]{\fbox{\includegraphics[trim=140 140 80 75,clip, width=\linewidth]{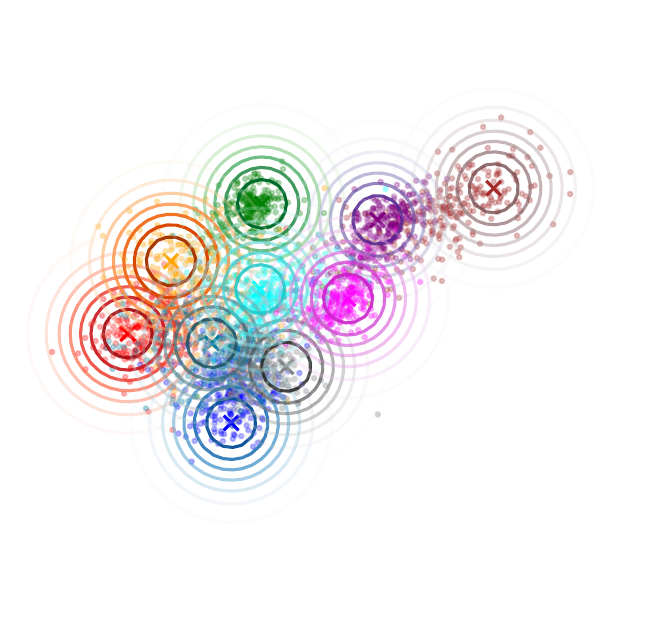}}}{1.0 $\rightarrow$ 0.7}
	\end{subfigure}
	\caption[Visualization of the inter-tightness effect on learned representations]{Visualization of the inter-tightness effect on learned representations ($K$=1) after 10 classes of FASHION.}
	\label{fig:igmm-inter}
\end{figure*}

\begin{figure*}[tb]
	\centering
	\setlength{\fboxrule}{0.5pt}
	\setlength{\fboxsep}{0pt}
	\begin{subfigure}{0.15\linewidth}
		\centering
		\stackunder[5pt]{\fbox{\includegraphics[trim=76 72 25 25,clip,width=\linewidth]{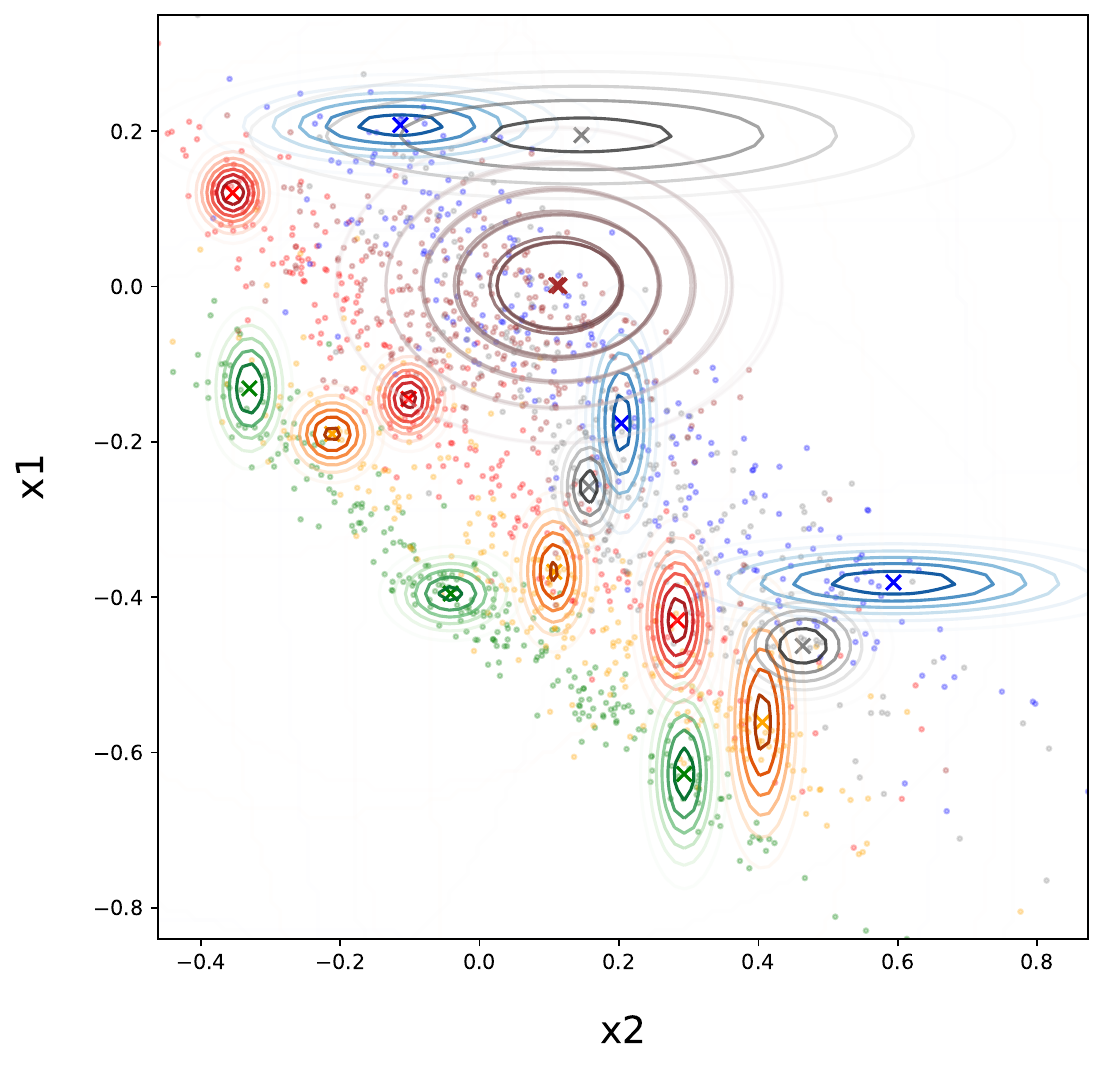}}}{0.05 $\rightarrow$ 0.46}
	\end{subfigure}\hspace{0.05cm}
	\begin{subfigure}{0.15\linewidth}
		\centering
		\stackunder[5pt]{\fbox{\raisebox{0.03\height}{\includegraphics[trim=76 72 15 30,clip,width=\linewidth]{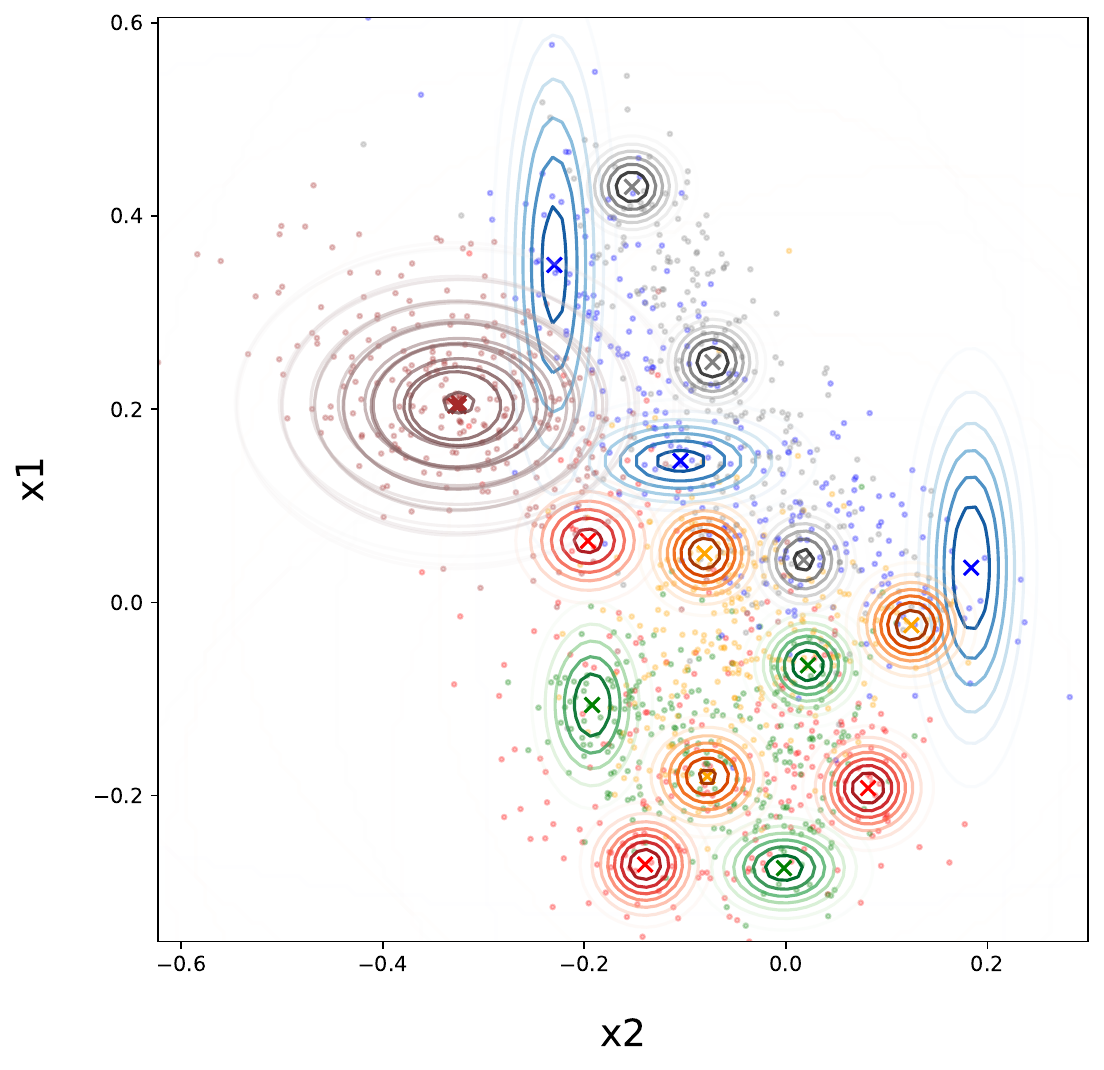}}}}{0.1 $\rightarrow$ 0.51}
	\end{subfigure}\hspace{0.05cm}
	\begin{subfigure}{0.15\linewidth}
		\centering
		\stackunder[5pt]{\fbox{\raisebox{0.2\height}{\includegraphics[trim=76 72 15 88,clip,width=\linewidth]{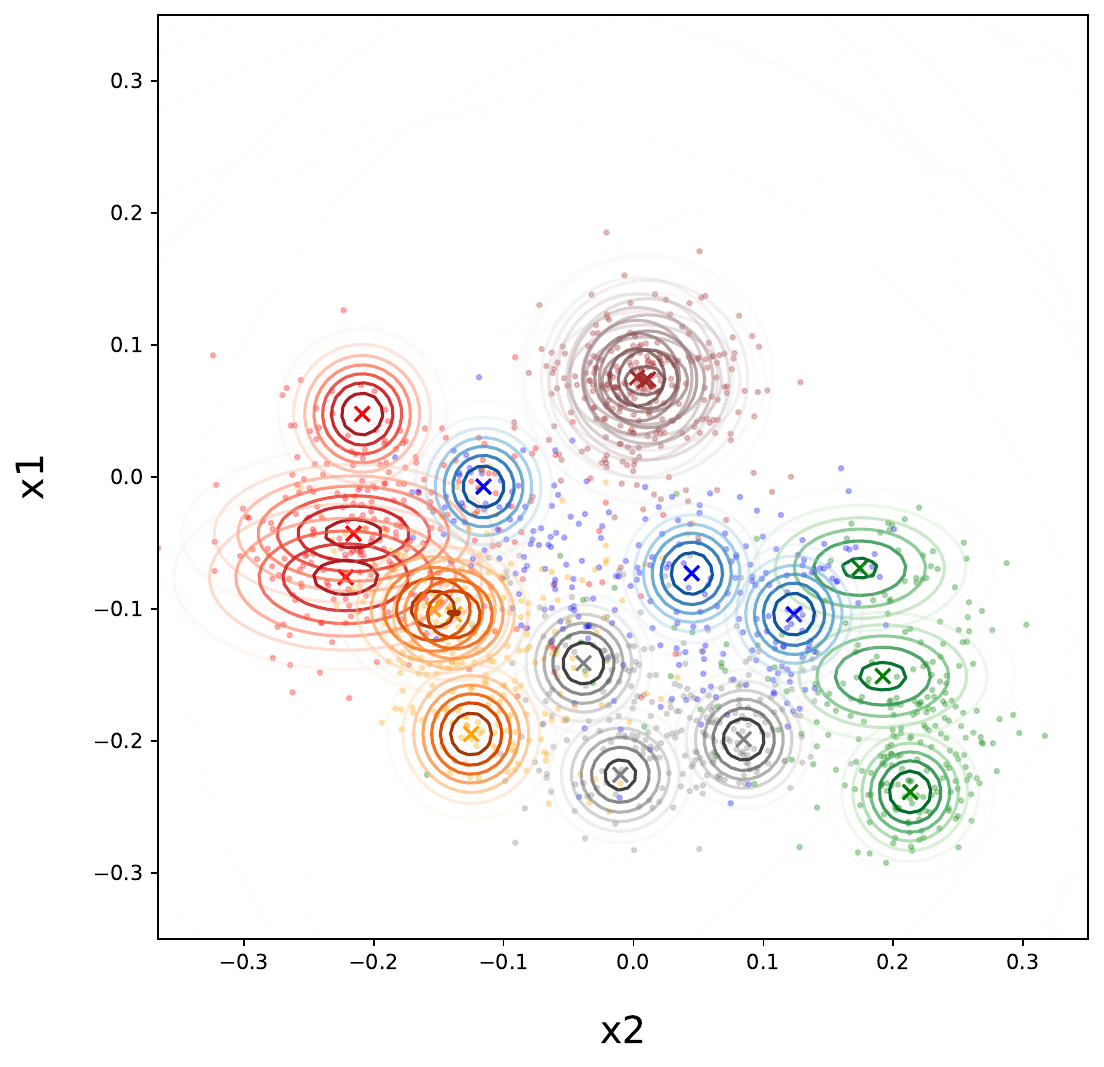}}}}{0.2 $\rightarrow$ 0.74}
	\end{subfigure}\hspace{0.05cm}
	\begin{subfigure}{0.15\linewidth}
		\centering
		\stackunder[5pt]{\fbox{\includegraphics[trim=76 72 15 15,clip,width=\linewidth]{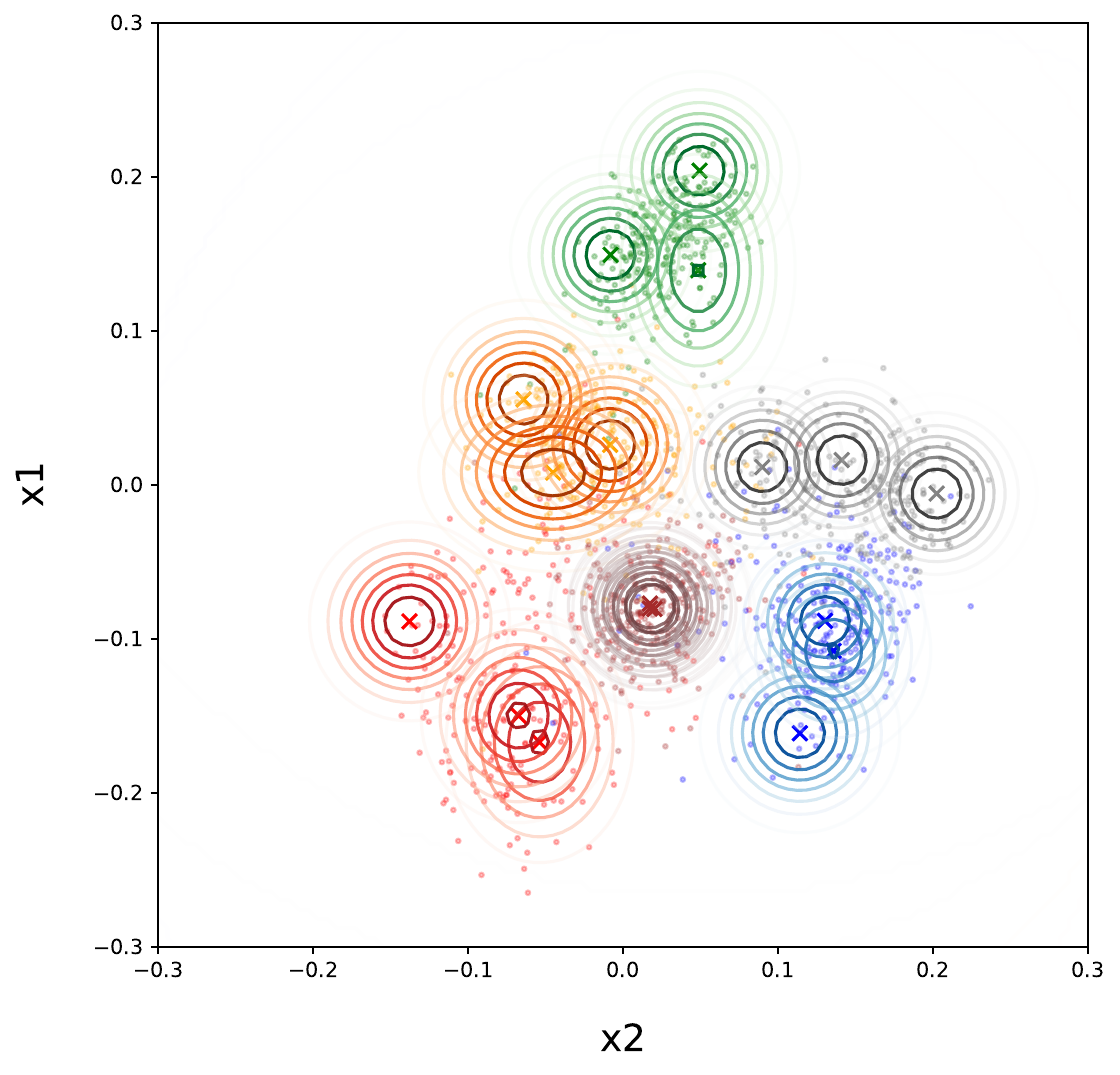}}}{0.25 $\rightarrow$ \textbf{0.82}}
	\end{subfigure}\hspace{0.05cm}
	\begin{subfigure}{0.15\linewidth}
		\centering
		\stackunder[5pt]{\fbox{\includegraphics[trim=76 72 15 15,clip, width=\linewidth]{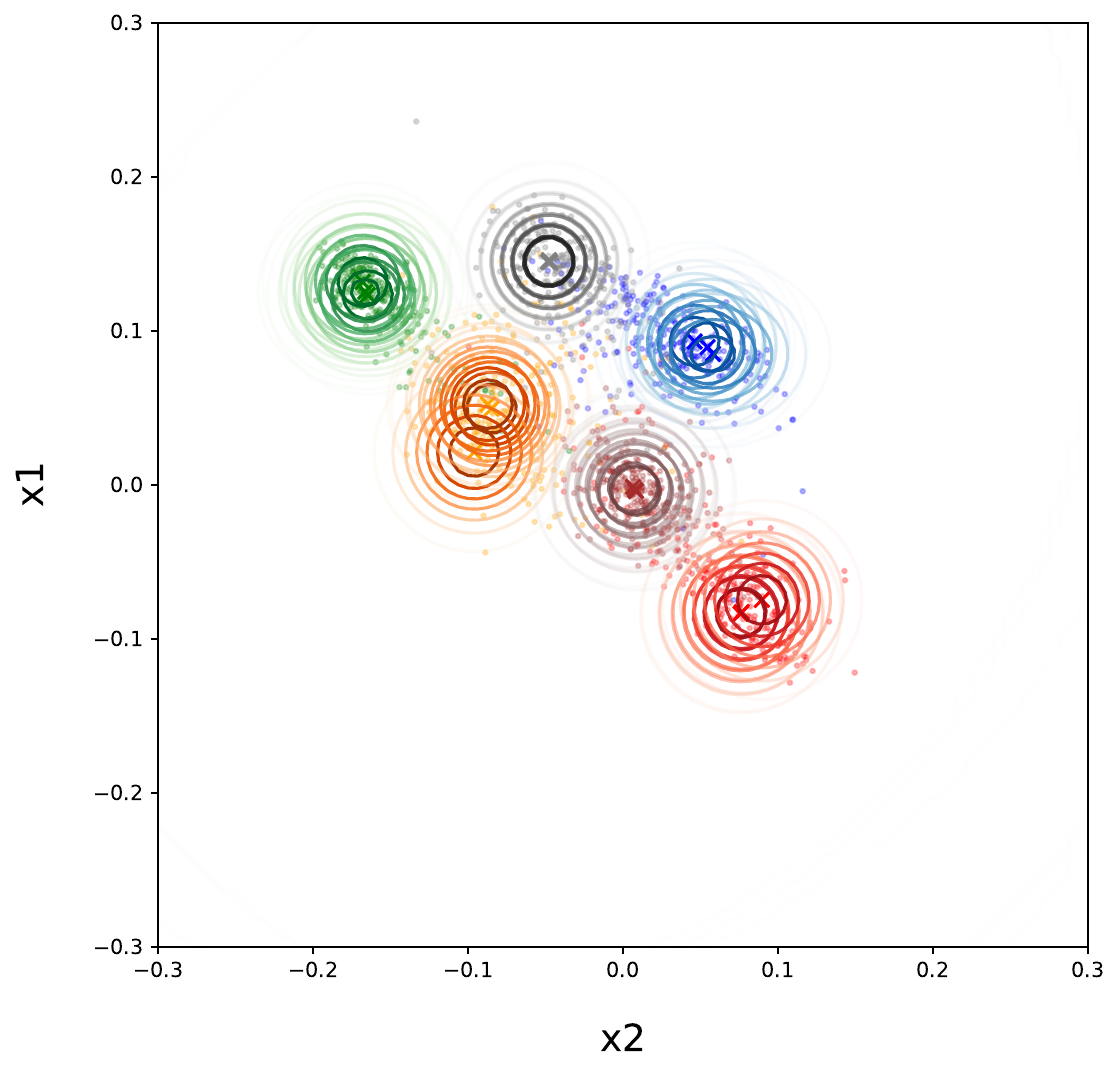}}}{0.3 $\rightarrow$ 0.82}
	\end{subfigure}
	\caption[Visualization of the intra-tightness effect on learned representations]{Visualization of the intra-tightness effect on learned representations ($K$=3) after 6 classes of FASHION.}
	\label{fig:igmm-intra}
\end{figure*}

In Fig. \ref{fig:igmm-tight}, we presented a grid of values for the average incremental accuracy per each pair of inter- and intra-tightness for every dataset. One can clearly see that imposing the constraint (tightness) on the inter- and intra-contrastive loss values is beneficial to the learning process. Most of the benchmarks required $\tau_{p, ie}$ at the level of 0.0001 or 0.001 and slightly higher intra-tightness $\tau_{p, ia}$ around 0.001 or 0.01 to achieve the best results. At the same time, one should notice that imposing too high inter-tightness (0.01) leads to abrupt deterioration of quality, which is a result of blocking the contrastive part of the loss from pushing components of different classes from each other. The influence of setting too high intra-tightness is less important since we may simply end up with a single component that can still be effectively used for classification. 

The examples for FASHION, given in Fig. \ref{fig:igmm-inter} and \ref{fig:igmm-intra}, show how increasing the inter-tightness (the first one) and intra-tightness (the second one) affects learned representations and mixture models. We can observe the positive impact of the constraint and the potential for sweet spots providing a good balance between differentiating components between each other and fitting them to the actual data. It is evident that too low values introduce critical instabilities to the learning process (very high contrastive loss values overwhelming the fitting part), while too high thresholds lead either to the decline of discriminative properties of the model or degenerate solutions. 

\paragraph{Baseline comparison:}

\begin{figure*}[bt]
	\centering
	\begin{subfigure}{0.24\linewidth}
		\centering
		\includegraphics[width=\linewidth]{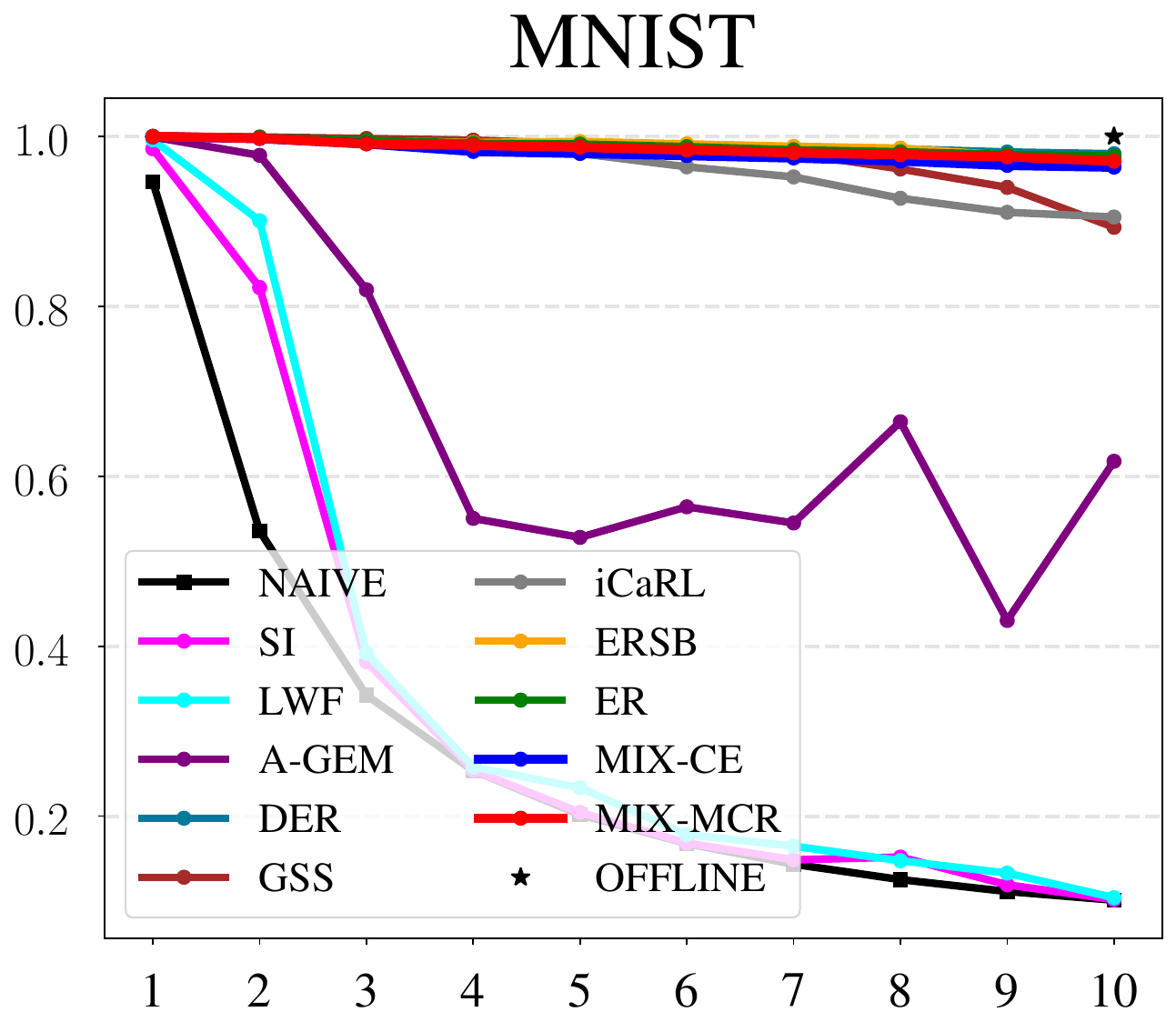}
	\end{subfigure}
	\begin{subfigure}{0.24\linewidth}
		\centering
		\includegraphics[width=\linewidth]{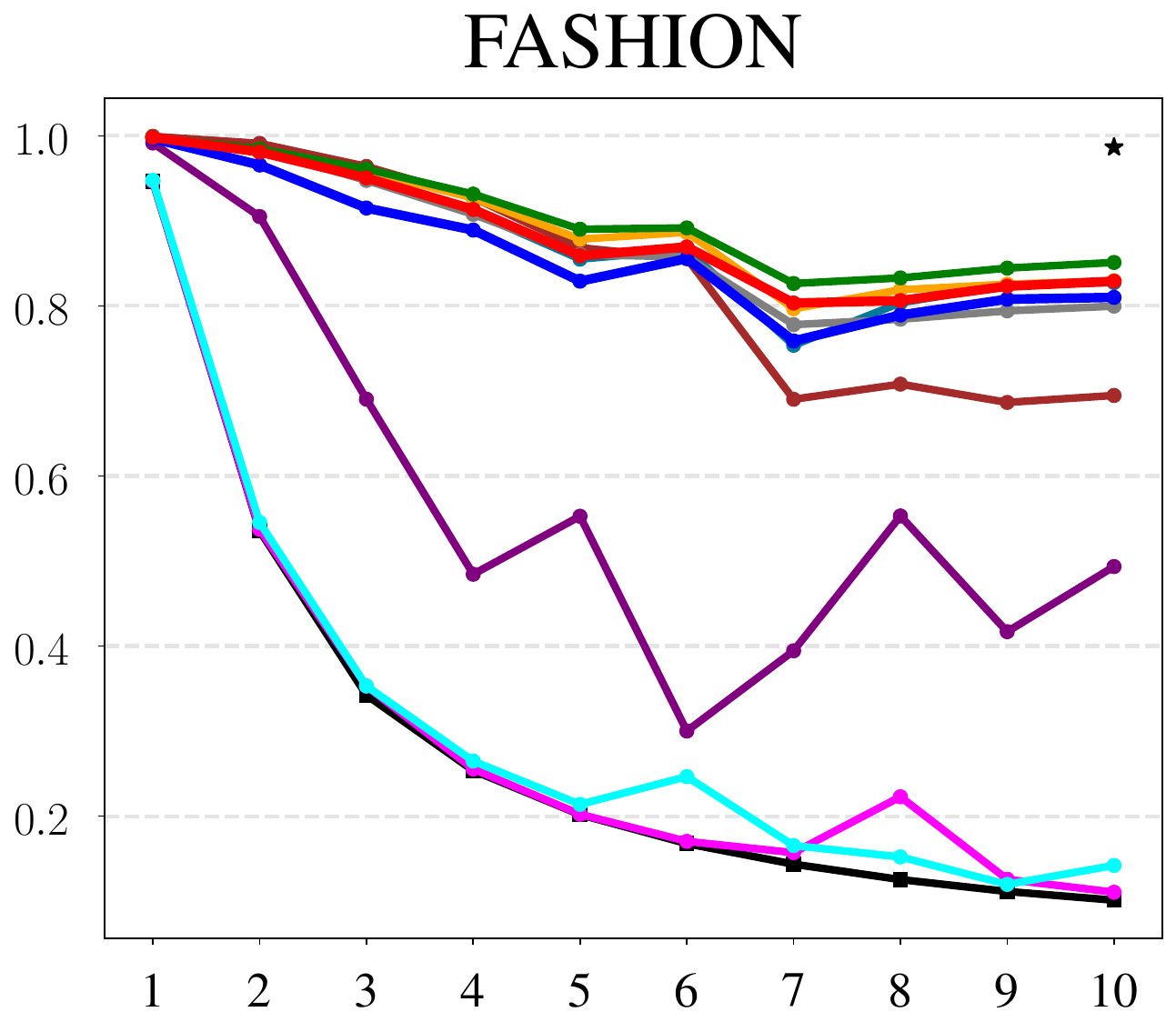}
	\end{subfigure}
	\begin{subfigure}{0.24\linewidth}
		\centering
		\includegraphics[width=\linewidth]{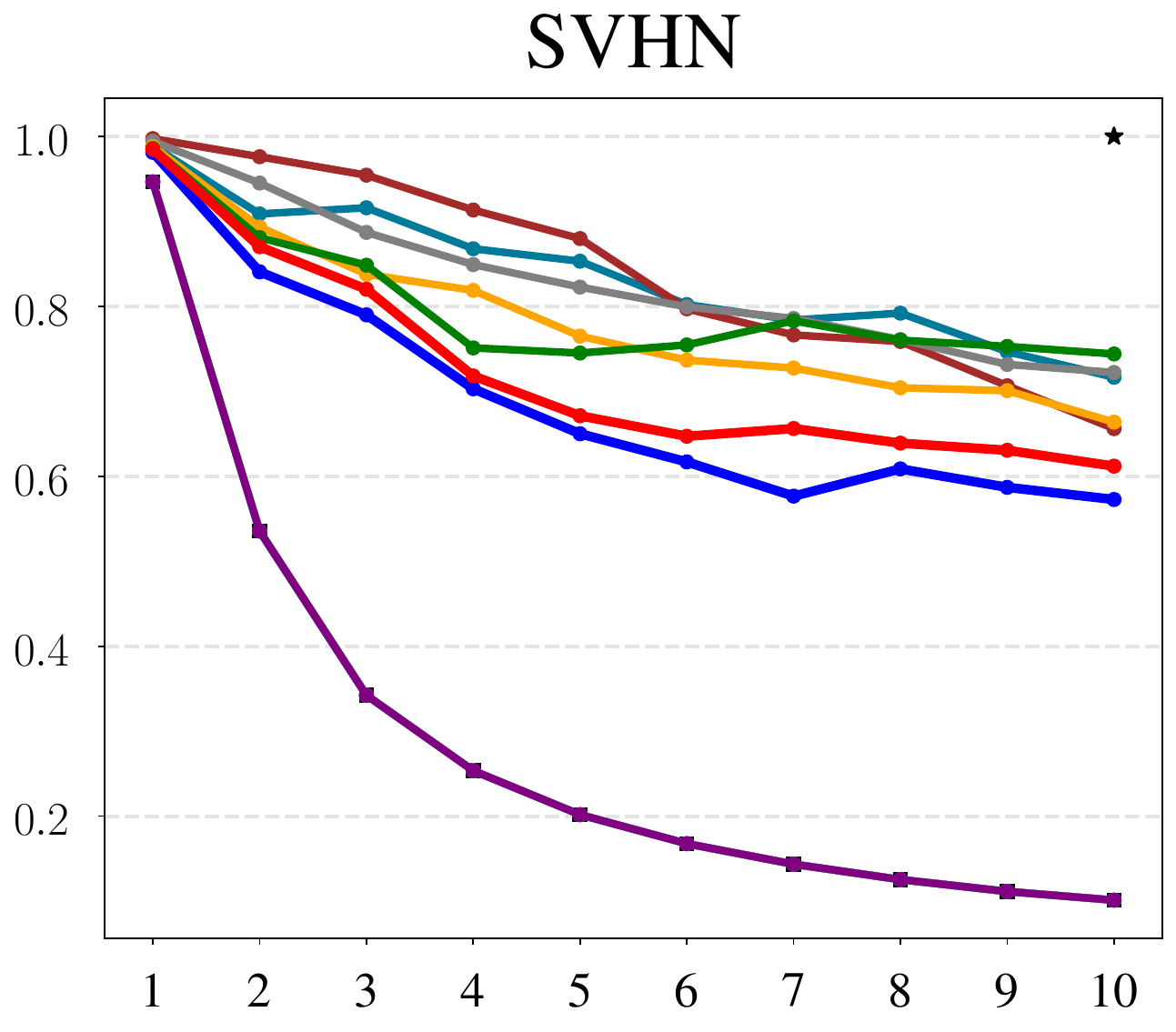}
	\end{subfigure}
	\begin{subfigure}{0.24\linewidth}
		\centering
		\includegraphics[width=\linewidth]{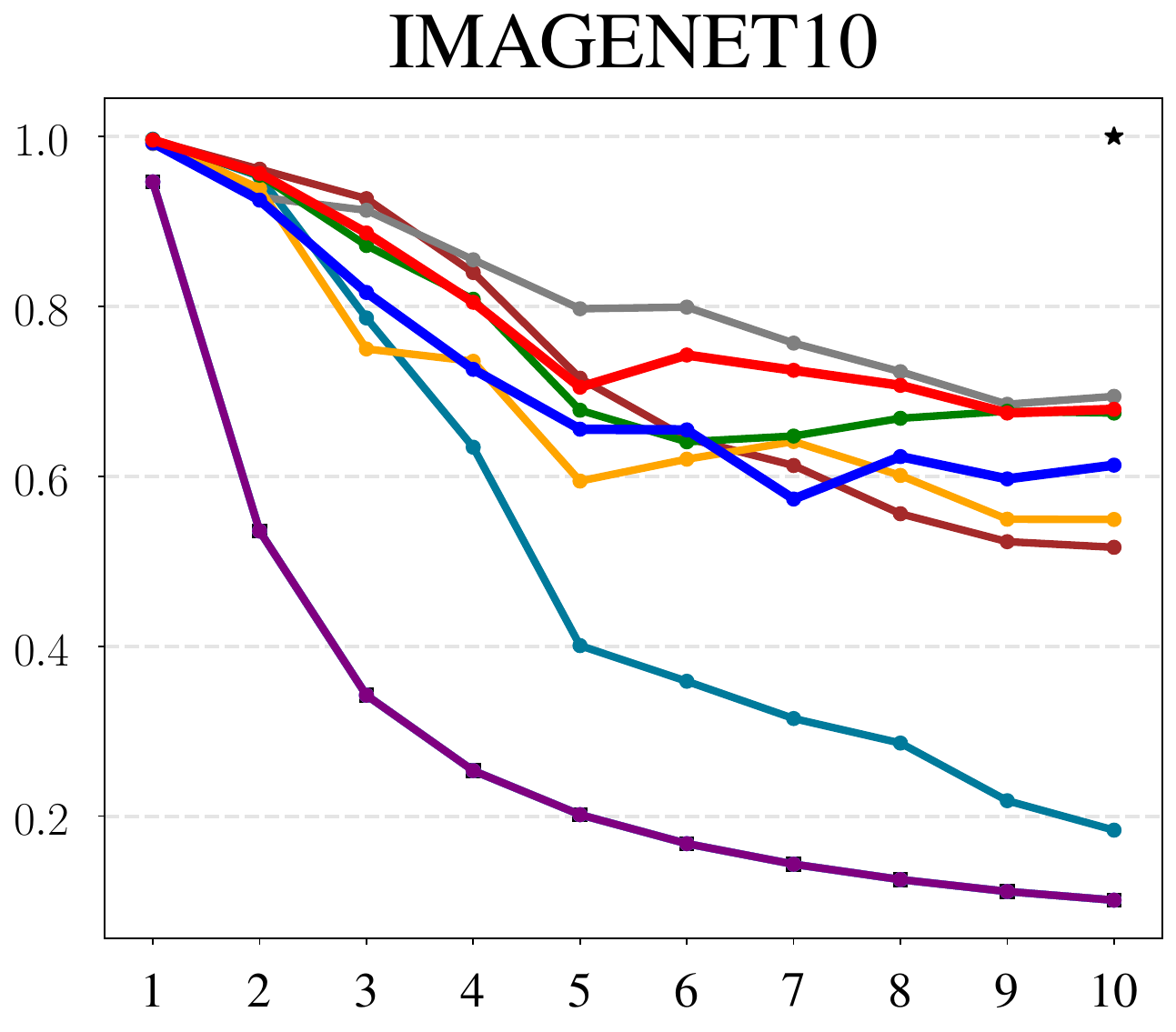}
	\end{subfigure}
	\begin{subfigure}{0.24\linewidth}
		\centering
		\includegraphics[width=\linewidth]{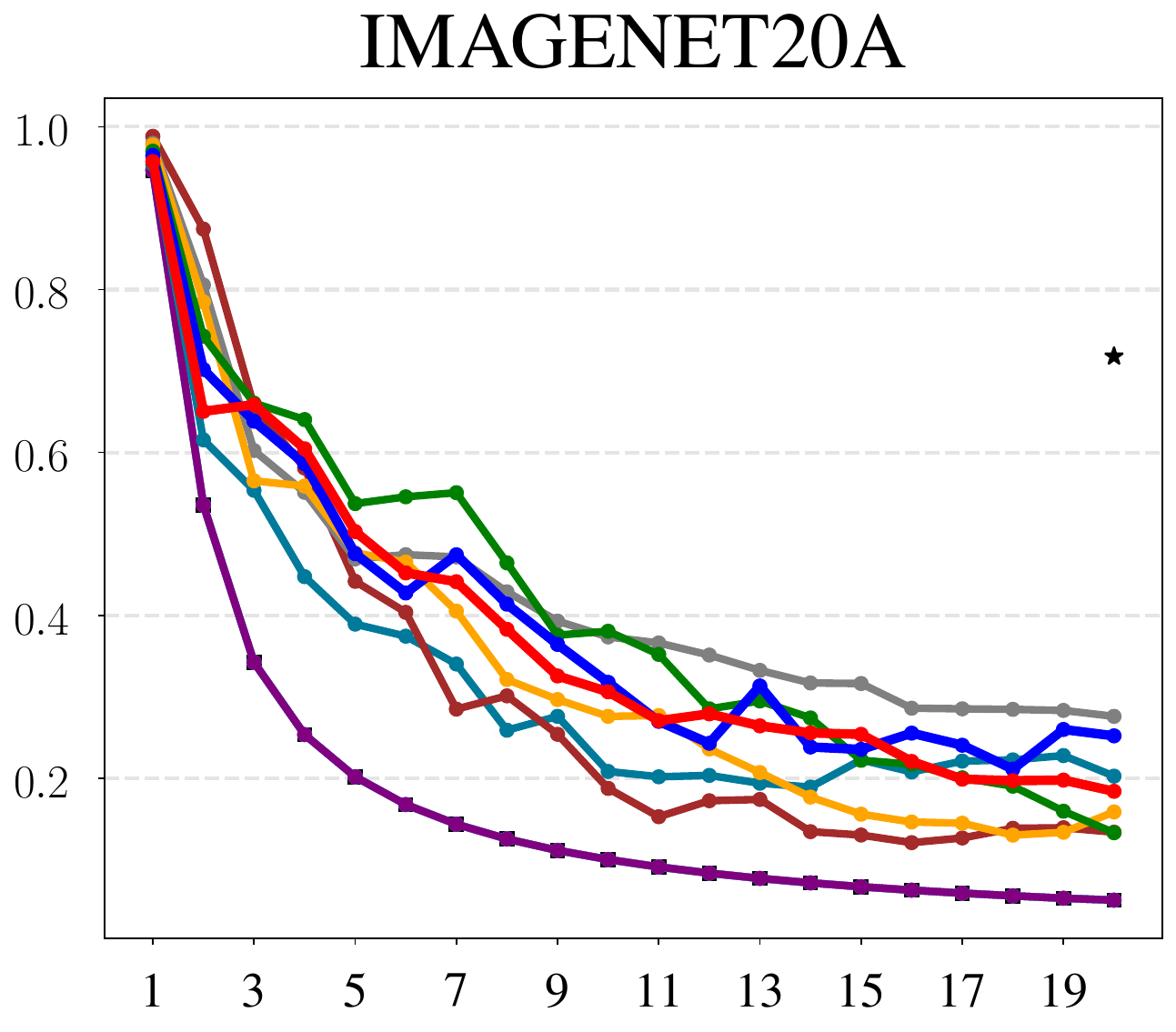}
	\end{subfigure}
	\begin{subfigure}{0.24\linewidth}
		\centering
		\includegraphics[width=\linewidth]{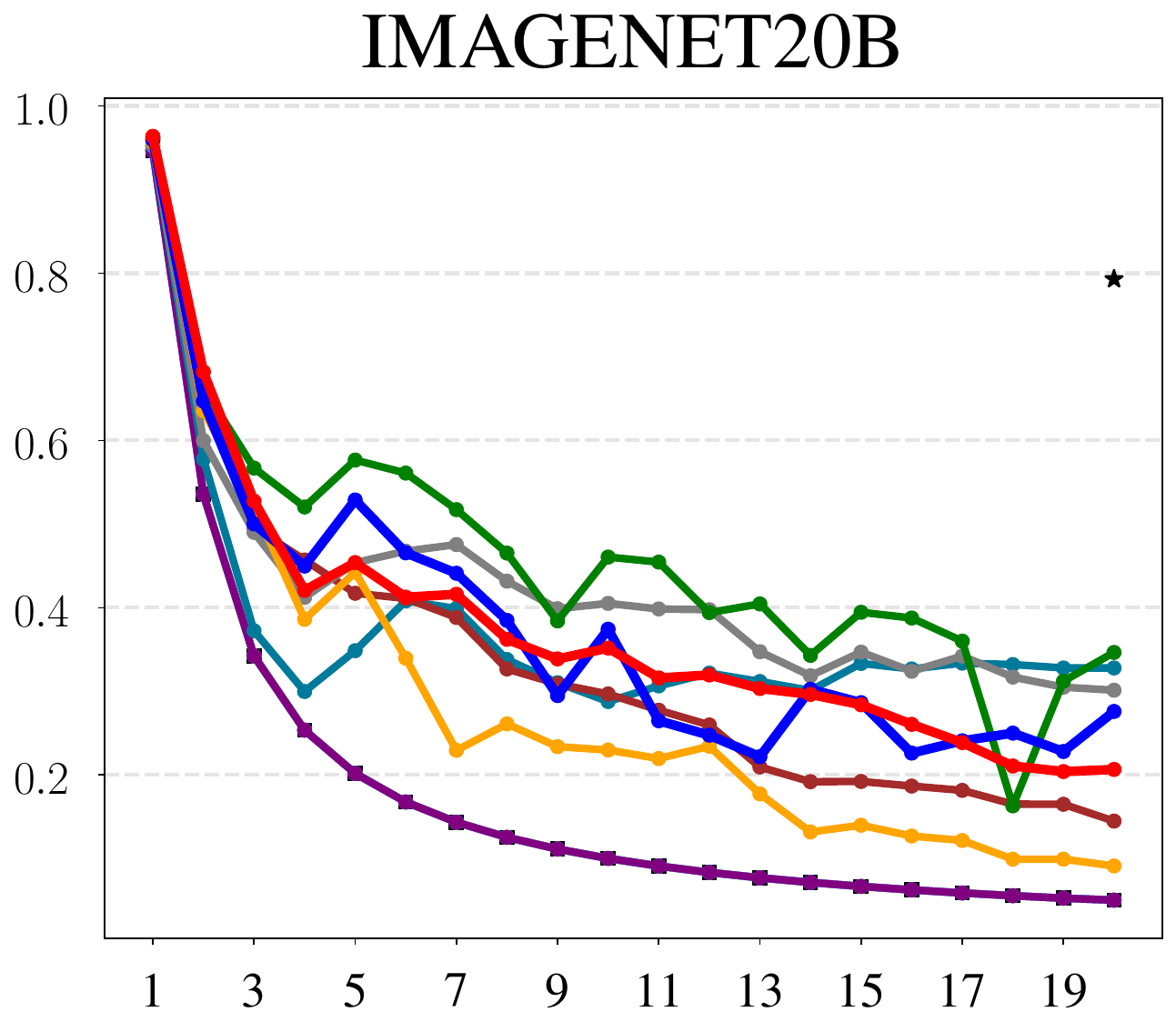}
	\end{subfigure}
	\begin{subfigure}{0.24\linewidth}
		\centering
		\includegraphics[width=\linewidth]{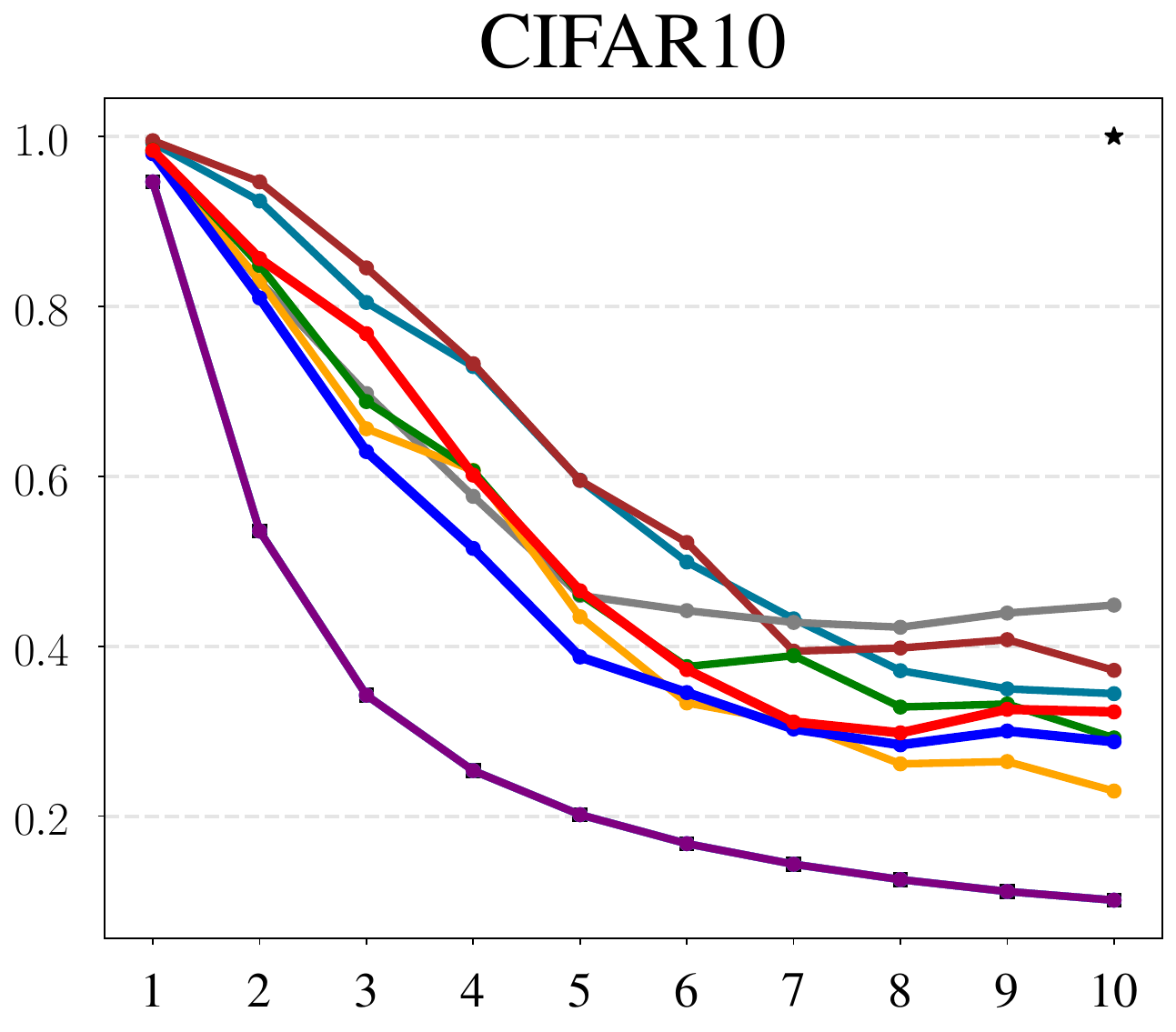}
	\end{subfigure}
	\begin{subfigure}{0.24\linewidth}
		\centering
		\includegraphics[width=\linewidth]{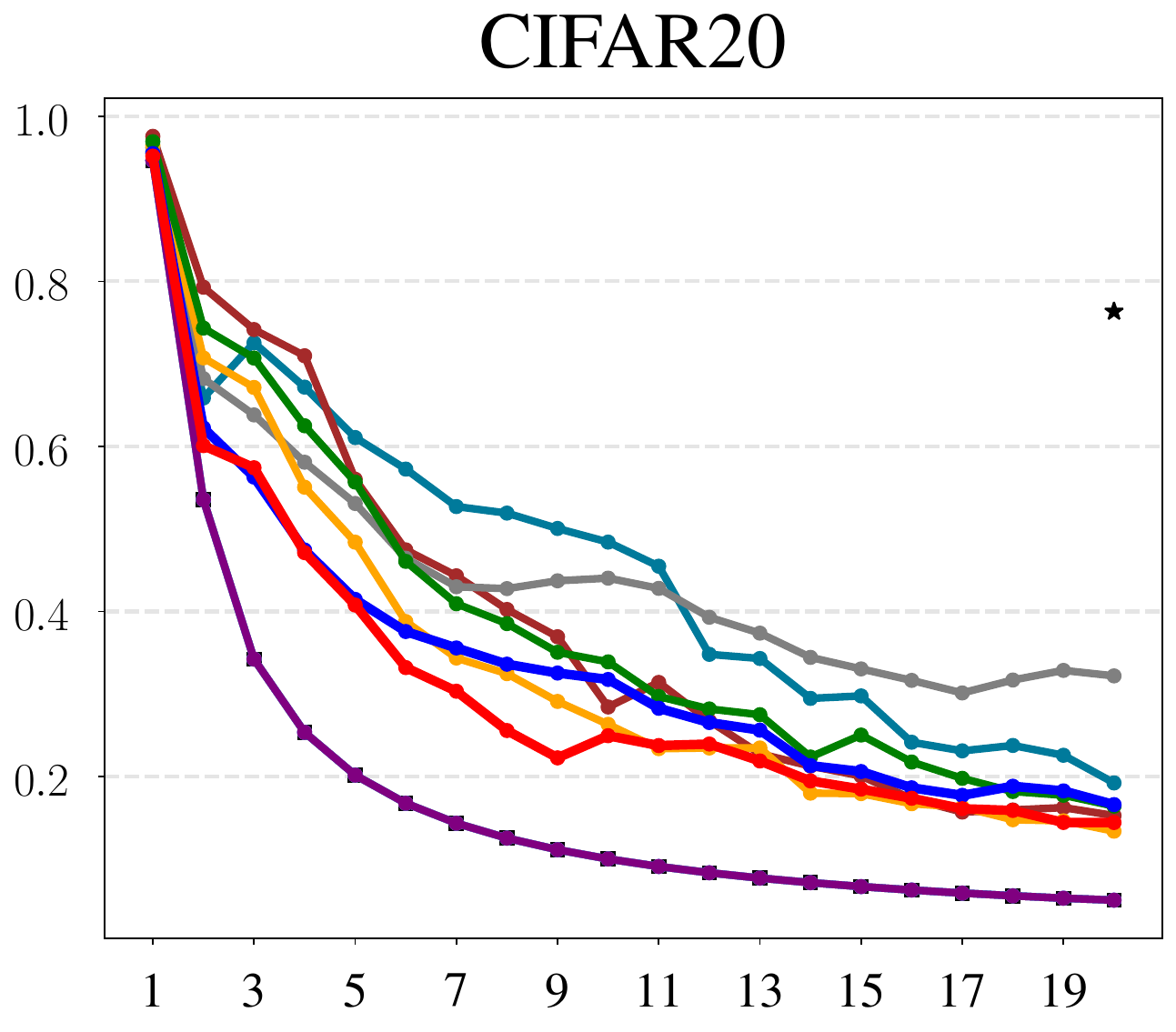}
	\end{subfigure}\\
	\begin{subfigure}{0.24\linewidth}
		\centering
		\includegraphics[width=\linewidth]{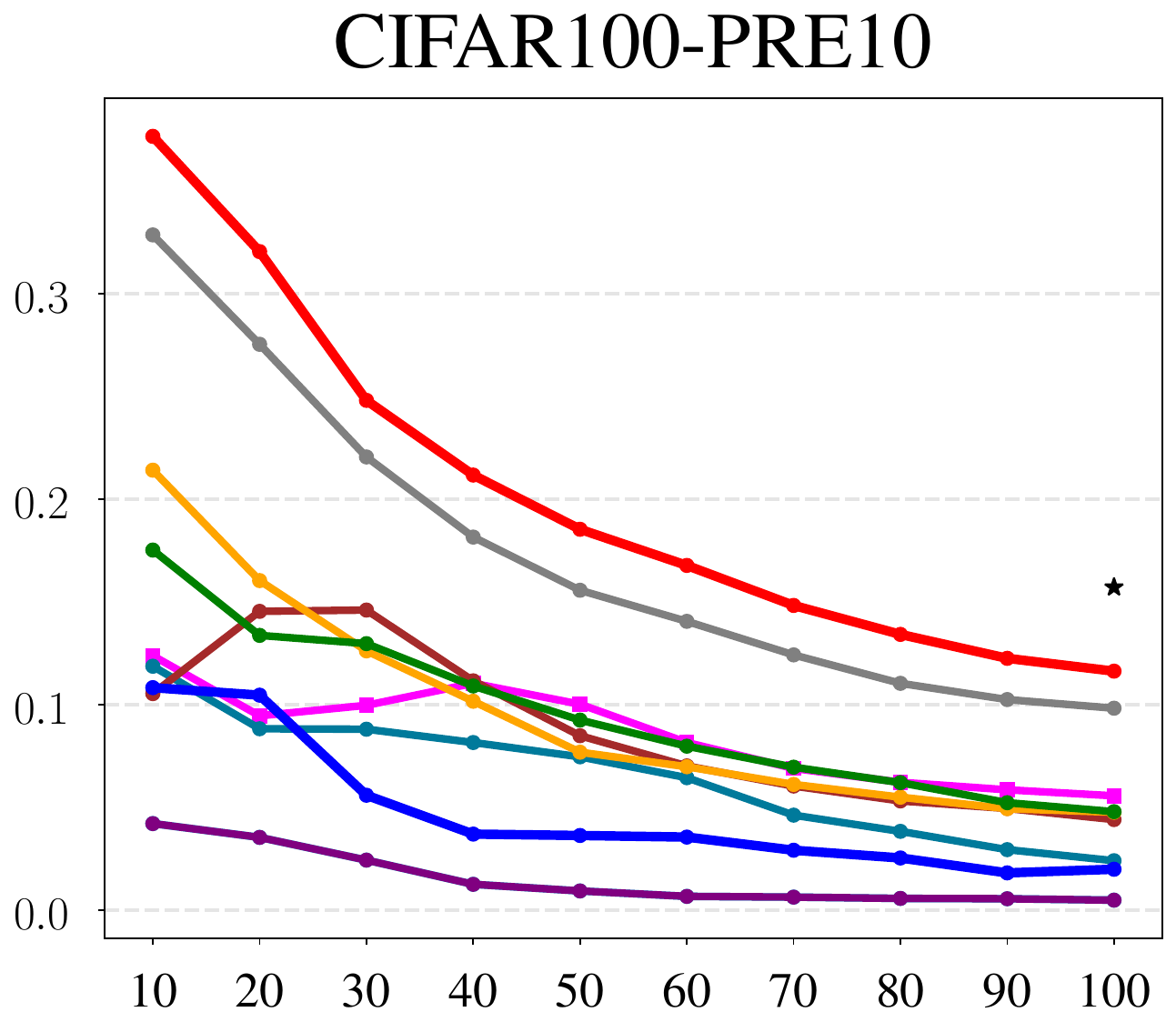}
	\end{subfigure}
	\begin{subfigure}{0.24\linewidth}
		\centering
		\includegraphics[width=\linewidth]{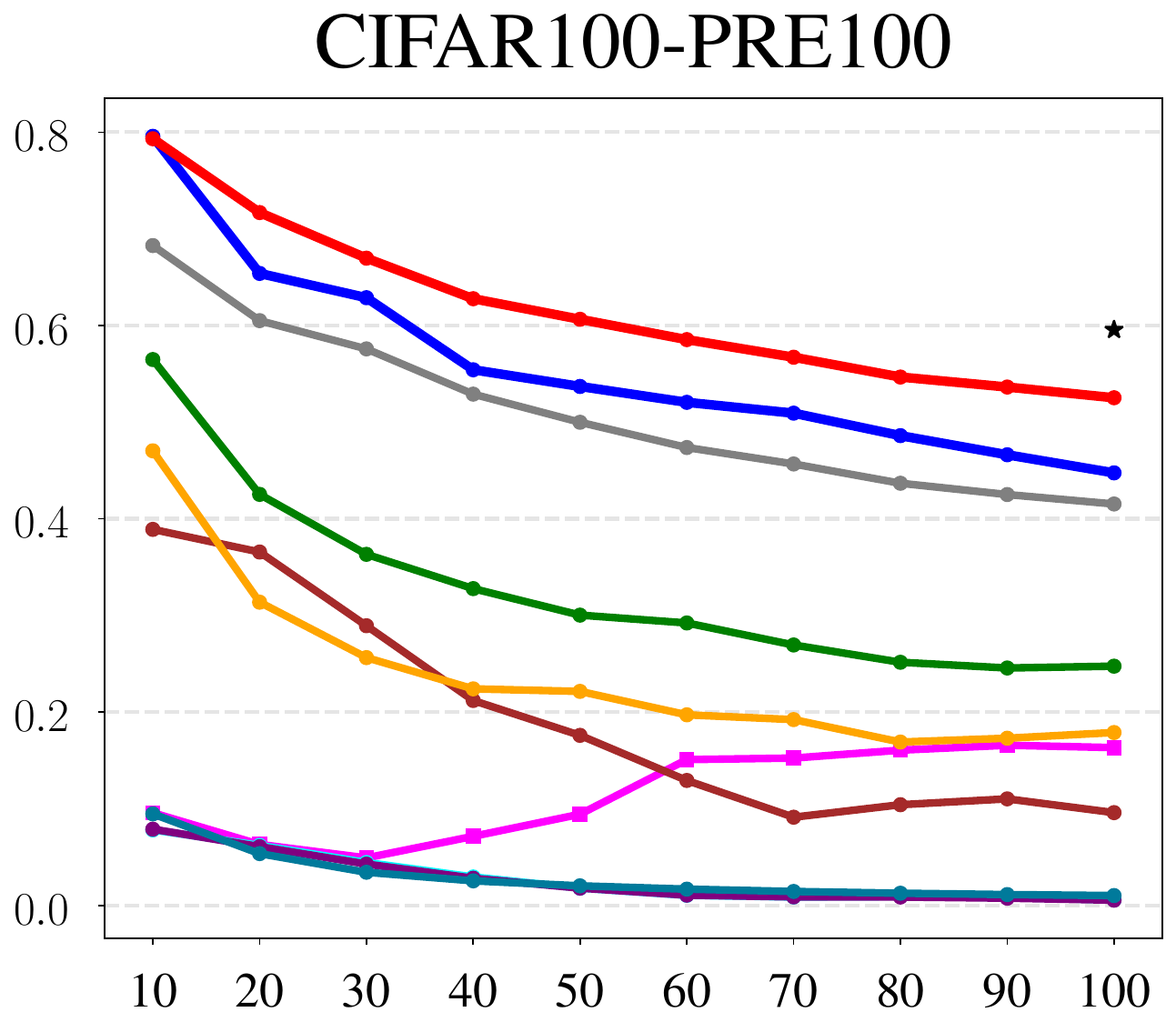}
	\end{subfigure}
	\begin{subfigure}{0.24\linewidth}
		\centering
		\includegraphics[width=\linewidth]{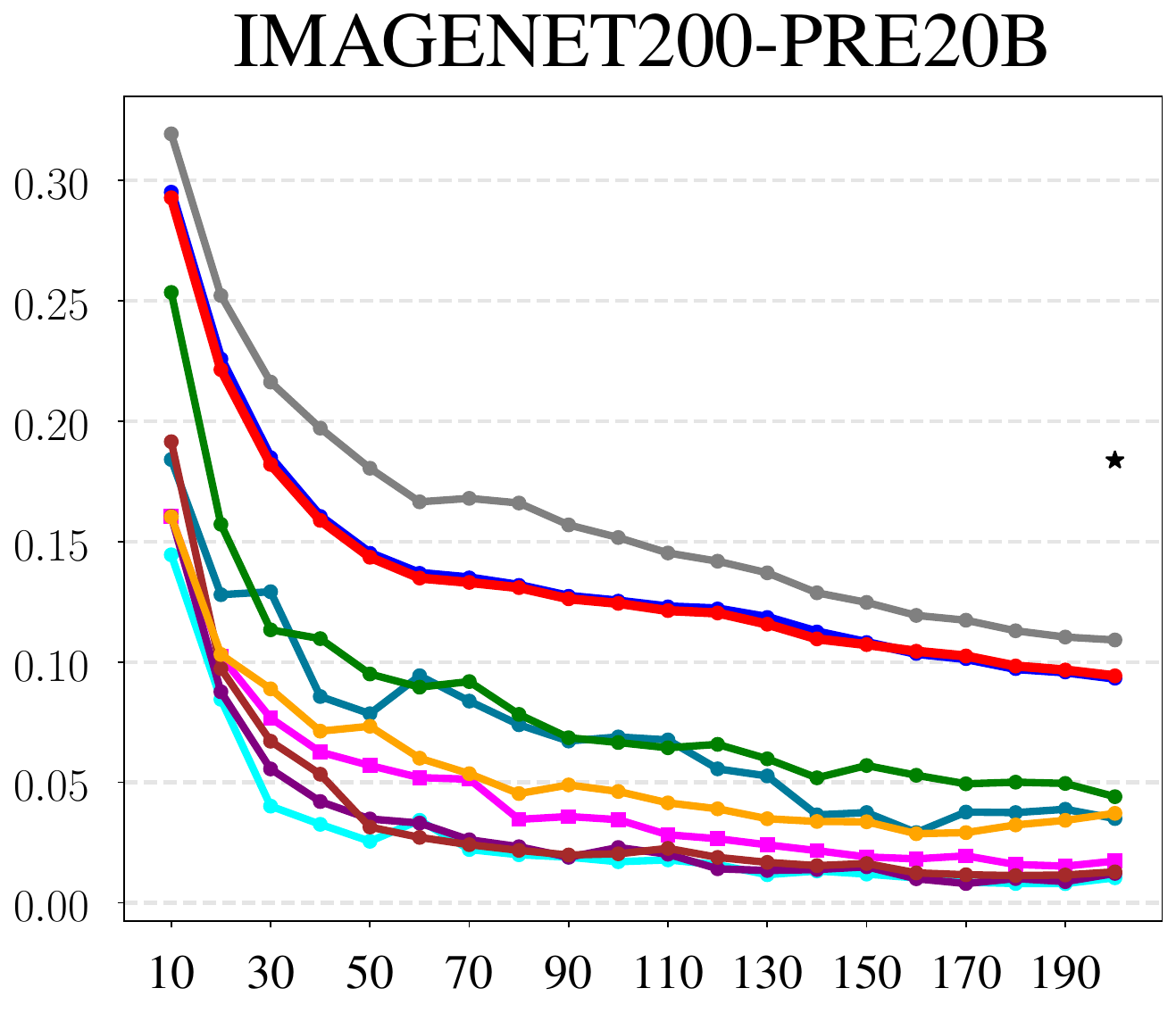}
	\end{subfigure}
	\begin{subfigure}{0.24\linewidth}
		\centering
		\includegraphics[width=\linewidth]{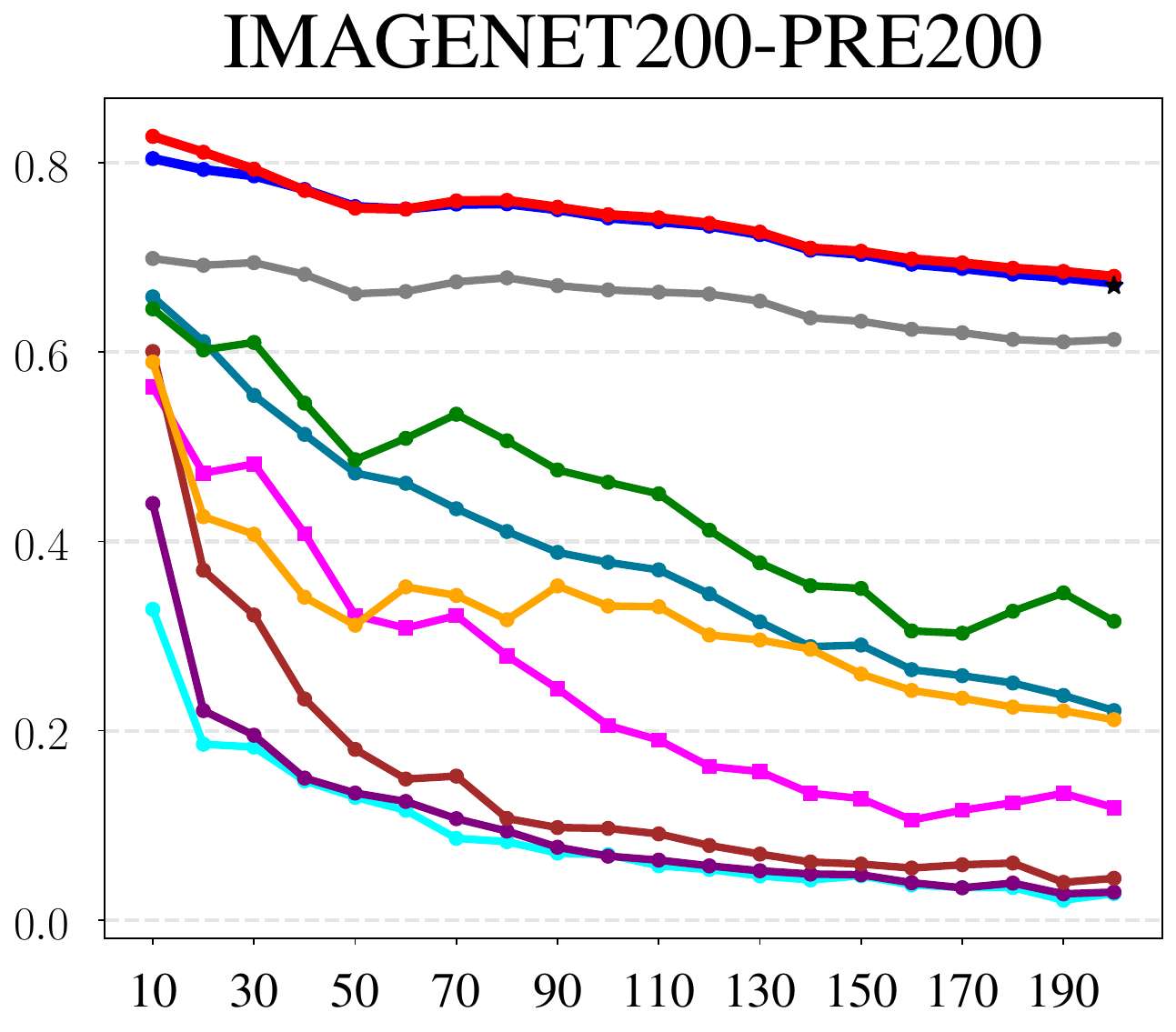}
	\end{subfigure}
	\caption[Incremental accuracy after each class batch for our methods and different baselines (1/2)]{Incremental accuracy after each class batch for our methods and different baselines.}
	\label{fig:igmm-final}
\end{figure*}

In the second section of our experimental study, we placed our algorithm in the class-incremental performance context by comparing it with the introduced baselines (Fig. \ref{fig:igmm-final}). First of all, we can see that the MIX-MCR variant performed better than the MIX-CE for most of the datasets, while being very close to it for the longer sequences (difference between less than 0.01 and 0.03). This proves that MIX-MCR is capable of providing not only a better representation (mixture) model but also that it is more reliable from the accuracy perspective. This also means that it is worth trying to maximize the quality of the produced Gaussian models as an alternative to high-level cross-entropy for classification. Secondly, although our model cannot be distinguished as the best classifier (being worse than iCaRL on average, with a difference equal to about 0.04), it is, at the same time, reliably competitive when compared with the remaining baselines (ER, GSS, DER) with a difference about 0.01 and less than 0.03. Also, it does not fall into the same pitfalls as either the weakest replay method (A-GEM) or the regularization-based ones (LWF, SI), outperforming them by almost 0.4 for accuracy on average. We can see that MIX could be found among the best models for MNIST, FASHION, IMAGENET10, IMAGENET20A and IMAGENET20B, especially at the end of the datasets, providing relatively reliable performance throughout the whole sequences. On the other hand, it struggled with catching up with the best replay methods for SVHN and CIFAR-based datasets showing that there is still a potential for improvements when it comes to predictive accuracy.

The overall very poor performance of LWF and SI (but also A-GEM), which were not much better than the NAIVE approach, confirms the observations made in other publications that the regularization-based methods cannot handle the most challenging 1-class-incremental scenarios without memory buffers \cite{Ven:2019three} even after improvements proposed in \cite{Maltoni:2019sit}.

We can also see that the for the scenarios with end-to-end training the models were much closer (0.01-0.3) to the OFFLINE upper bound for the shorter sequences (MNIST, FASHION, SVHN and IMAGENET10, except for CIFAR10) than for the longer ones (IMAGENET20A, IMAGENET20B, CIFAR20) with differences between 0.4-0.5, which shows that all of the state-of-the-art methods still struggle with bridging the gap between incremental learning and offline optimum.

Finally, the results for the memory-free scenarios with pre-trained models, given in the last row of Fig. \ref{fig:igmm-final}, exhibit the main strength of the MIX algorithm. Since in these scenarios, it does not use the inter-contrastive loss, it can perfectly separate the incremental learning process for each class, preventing catastrophic forgetting at the level of the classifier. As a result, it does not have to rehearse the previous concepts at all ($\mathcal{M}_c$=0) while still being able to conduct very effective learning producing results very close to the OFFLINE upper bound (difference between about 0 and 0.1), regardless of the quality of the extractor (pre-trained on 10 and 20 or 100 and 200 classes). The MIX-MCR method outperforms all of the baselines for all cases except for IMAGENET200-PRE20, for which only iCaRL was able to provide slightly higher accuracy, even though they had a small advantage of having approximately one example per class in the buffer. It is not a coincidence that practically only iCaRL is close to our method on average (worse by about 0.1), since it uses a similar paradigm in the classification layer by storing prototypes/centroids that are used for classification. All of the remaining algorithms cannot handle the memory-free scenario effectively, producing solutions worse by at least 0.2 on average. This can be a crucial property when one has to consider, for example, data privacy issues or mobile and edge computing.

All of the presented observations, conclusions and recommendations can be also found in a condensed form at the end of Appendix B.


\section{Summary}

In this work, we introduced a class-incremental mixture of Gaussians model (MIX) for deep continual learning. We proposed different variants of the algorithm to make it suitable for gradient-based optimization and, through an extensive experimental study, we exhibited its practical configurations and capabilities in the context of other state-of-the-art continual learning models. 

In our future research, we will focus on replacing the regionalization approach with a more flexible method that do not assume any pre-training structure and allows the gradient-based procedure to fully explore potential solutions, e.g. annealing \cite{Gepperth2021:grad}, and on removing the static tightness hyperparameter to increase flexibility even more -- it could be more beneficial to either find a better (parameter-free) distance function or propose an adaptive threshold. It is also an open question whether we can effectively train a gradient-based mixture using a full covariance matrix. Finally, we could consider some kind of hybridization of the mixture models with the feature extractor to benefit from the capabilities of the former to limit interference with previously learned concepts by utilizing max-component losses. All of these potential improvements combined could provide significant performance gains in the class-incremental continual learning scenarios.

{\small
\bibliographystyle{ieee_fullname}
\bibliography{main}

\begin{thebibliography}{10}\itemsep=-1pt

\bibitem{Aljundi:2018mas}
Rahaf Aljundi, Francesca Babiloni, Mohamed Elhoseiny, Marcus Rohrbach, and
  Tinne Tuytelaars.
\newblock {Memory Aware Synapses: Learning what (not) to forget}.
\newblock In {\em European Conference on Computer Vision (ECCV)}, 2018.

\bibitem{Aljundi:2017expert}
Rahaf Aljundi, Punarjay Chakravarty, and Tinne Tuytelaars.
\newblock {Expert Gate: Lifelong Learning with a Network of Experts}.
\newblock {\em 2017 IEEE Conference on Computer Vision and Pattern Recognition
  (CVPR)}, pages 7120--7129, 2017.

\bibitem{Aljundi2019:gss}
Rahaf Aljundi, Min Lin, Baptiste Goujaud, and Yoshua Bengio.
\newblock {Gradient based sample selection for online continual learning}.
\newblock In {\em Advances in Neural Information Processing Systems (NeurIPS)},
  2019.

\bibitem{Bang:2021}
Jihwan Bang, Heesu Kim, Young~Joon Yoo, Jung-Woo Ha, and Jonghyun Choi.
\newblock {Rainbow Memory: Continual Learning with a Memory of Diverse
  Samples}.
\newblock In {\em 2021 IEEE/CVF Conference on Computer Vision and Pattern
  Recognition (CVPR)}, pages 8214--8223, 2021.

\bibitem{Buzzega2020:dark}
Pietro Buzzega, Matteo Boschini, Angelo Porrello, Davide Abati, and Simone
  Calderara.
\newblock {Dark Experience for General Continual Learning: a Strong, Simple
  Baseline}.
\newblock In {\em Advances in Neural Information Processing Systems (NeurIPS)},
  2020.

\bibitem{Chaudhry:2019}
Arslan Chaudhry, Marc’Aurelio Ranzato, Marcus Rohrbach, and Mohamed
  Elhoseiny.
\newblock {Efficient Lifelong Learning with A-{GEM}}.
\newblock In {\em International Conference on Learning Representations (ICLR)},
  2019.

\bibitem{Dhar:2019}
Prithviraj Dhar, Rajat~Vikram Singh, Kuan-Chuan Peng, Ziyan Wu, and Rama
  Chellappa.
\newblock {Learning Without Memorizing}.
\newblock In {\em 2019 IEEE/CVF Conference on Computer Vision and Pattern
  Recognition (CVPR)}, pages 5133--5141, 2019.

\bibitem{Engel:2010igmm}
Paulo~Martins Engel and Milton~Roberto Heinen.
\newblock {Incremental Learning of Multivariate Gaussian Mixture Models}.
\newblock In {\em Advances in Artificial Intelligence -- SBIA 2010}, pages
  82--91, 2010.

\bibitem{Finn:2017maml}
Chelsea Finn, Pieter Abbeel, and Sergey Levine.
\newblock {Model-Agnostic Meta-Learning for Fast Adaptation of Deep Networks}.
\newblock In {\em Proceedings of the 34th International Conference on Machine
  Learning (ICML)}, 2017.

\bibitem{French:1999}
Robert~M. French.
\newblock {Catastrophic forgetting in connectionist networks}.
\newblock {\em Trends in Cognitive Sciences}, 3(4):128--135, 1999.

\bibitem{Gepperth2021:grad}
Alexander Rainer~Tassilo Gepperth and Benedikt Pf{\"u}lb.
\newblock {Gradient-Based Training of Gaussian Mixture Models for
  High-Dimensional Streaming Data}.
\newblock {\em Neural Process. Lett.}, 53:4331--4348, 2021.

\bibitem{Hayes:2021rep}
Tyler~L. Hayes, Giri~P. Krishnan, Maxim Bazhenov, Hava~T. Siegelmann,
  Terrence~J. Sejnowski, and Christopher Kanan.
\newblock {Replay in Deep Learning: Current Approaches and Missing Biological
  Elements}.
\newblock {\em Neural Computation}, 33(11):2908--2950, 10 2021.

\bibitem{Higham:2009cholesky}
{Nicholas J.} Higham.
\newblock {Cholesky factorization}.
\newblock {\em Wiley Interdisciplinary Reviews: Computational Statistics},
  1(2):251--254, 2009.

\bibitem{Hinton:2015}
Geoffrey~E. Hinton, Oriol Vinyals, and Jeffrey Dean.
\newblock {Distilling the Knowledge in a Neural Network}.
\newblock {\em arXiv}, abs/1503.02531, 2015.

\bibitem{Hosseini:2015}
Reshad Hosseini and Suvrit Sra.
\newblock {Matrix Manifold Optimization for Gaussian Mixtures}.
\newblock In {\em Advances in Neural Information Processing Systems (NeurIPS)},
  2015.

\bibitem{Hou:2018}
Saihui Hou, Xinyu Pan, Chen~Change Loy, Zilei Wang, and Dahua Lin.
\newblock {Lifelong Learning via Progressive Distillation and Retrospection}.
\newblock In {\em European Conference on Computer Vision (ECCV)}, 2018.

\bibitem{Hsu:2018}
Yen-Chang Hsu, Yen-Cheng Liu, and Zsolt Kira.
\newblock {Re-evaluating Continual Learning Scenarios: A Categorization and
  Case for Strong Baselines}.
\newblock {\em arXiv}, abs/1810.12488, 2018.

\bibitem{Hung:2019compact}
Steven C.~Y. Hung, Cheng-Hao Tu, Cheng-En Wu, Chien-Hung Chen, Yi-Ming Chan,
  and Chu-Song Chen.
\newblock {Compacting, Picking and Growing for Unforgetting Continual
  Learning}.
\newblock In {\em Proceedings of the 33rd International Conference on Neural
  Information Processing Systems}, 2019.

\bibitem{Javed:2019meta}
Khurram Javed and Martha White.
\newblock {Meta-Learning Representations for Continual Learning}.
\newblock In {\em Advances in Neural Information Processing Systems (NeurIPS)},
  2019.

\bibitem{Kemker:2018}
Ronald Kemker, Angelina Abitino, Marc McClure, and Christopher Kanan.
\newblock {Measuring Catastrophic Forgetting in Neural Networks}.
\newblock In {\em AAAI Conference on Artificial Intelligence (AAAI)}, 2018.

\bibitem{Kirkpatrick:2017}
James Kirkpatrick, Razvan Pascanu, Neil Rabinowitz, Joel Veness, Guillaume
  Desjardins, Andrei~A. Rusu, Kieran Milan, John Quan, Tiago Ramalho, Agnieszka
  Grabska-Barwinska, Demis Hassabis, Claudia Clopath, Dharshan Kumaran, and
  Raia Hadsell.
\newblock {Overcoming catastrophic forgetting in neural networks}.
\newblock {\em Proceedings of the National Academy of Sciences},
  114(13):3521--3526, 2017.

\bibitem{Korycki:2021ercd}
{\L}ukasz Korycki and Bartosz Krawczyk.
\newblock {Class-Incremental Experience Replay for Continual Learning under
  Concept Drift}.
\newblock {\em 2021 IEEE/CVF Conference on Computer Vision and Pattern
  Recognition Workshops (CVPRW)}, pages 3644--3653, 2021.

\bibitem{Korycki:2021lldt}
{\L}ukasz Korycki and Bartosz Krawczyk.
\newblock {Streaming Decision Trees for Lifelong Learning}.
\newblock In {\em Machine Learning and Knowledge Discovery in Databases.
  Research Track - European Conference (ECML/PKDD)}, pages 502--518, 2021.

\bibitem{Lange:2019}
Matthias~De Lange, Rahaf Aljundi, Marc Masana, Sarah Parisot, Xu Jia, Ales
  Leonardis, Gregory~G. Slabaugh, and Tinne Tuytelaars.
\newblock {Continual learning: {A} comparative study on how to defy forgetting
  in classification tasks}.
\newblock {\em CoRR}, abs/1909.08383, 2019.

\bibitem{Li:2018}
Zhizhong Li and Derek Hoiem.
\newblock {Learning without Forgetting}.
\newblock {\em IEEE Transactions on Pattern Analysis and Machine Intelligence},
  40(12):2935--2947, 2018.

\bibitem{Lomonaco2021:ava}
Vincenzo Lomonaco, Lorenzo Pellegrini, Andrea Cossu, Antonio Carta, Gabriele
  Graffieti, Tyler~L. Hayes, Matthias~De Lange, Marc Masana, Jary Pomponi, Gido
  van~de Ven, Martin Mundt, Qi She, Keiland Cooper, Jeremy Forest, Eden
  Belouadah, Simone Calderara, German~I. Parisi, Fabio Cuzzolin, Andreas
  Tolias, Simone Scardapane, Luca Antiga, Subutai Amhad, Adrian Popescu,
  Christopher Kanan, Joost van~de Weijer, Tinne Tuytelaars, Davide Bacciu, and
  Davide Maltoni.
\newblock Avalanche: an end-to-end library for continual learning.
\newblock In {\em IEEE/CVF Conference on Computer Vision and Pattern
  Recognition Workshops (CVPRW)}, 2021.

\bibitem{Mai:2021contr}
Zheda Mai, Ruiwen Li, Hyunwoo Kim, and Scott Sanner.
\newblock {Supervised Contrastive Replay: Revisiting the Nearest Class Mean
  Classifier in Online Class-Incremental Continual Learning}.
\newblock In {\em 2021 IEEE/CVF Conference on Computer Vision and Pattern
  Recognition Workshops (CVPRW)}, pages 3584--3594, 2021.

\bibitem{Mallya:2018pack}
Arun Mallya and Svetlana Lazebnik.
\newblock {PackNet: Adding Multiple Tasks to a Single Network by Iterative
  Pruning}.
\newblock {\em 2018 IEEE/CVF Conference on Computer Vision and Pattern
  Recognition (CVPR)}, pages 7765--7773, 2018.

\bibitem{Mallya:2018piggy}
Arun Mallya and Svetlana Lazebnik.
\newblock {Piggyback: Adding Multiple Tasks to a Single, Fixed Network by
  Learning to Mask}.
\newblock {\em arXiv}, abs/1801.06519, 2018.

\bibitem{Maltoni:2019sit}
Davide Maltoni and Vincenzo Lomonaco.
\newblock {Continuous learning in single-incremental-task scenarios}.
\newblock {\em Neural Networks}, 116:56--73, 2019.

\bibitem{Masana:2020}
Marc Masana, Xialei Liu, Bartlomiej Twardowski, Mikel Menta, Andrew~D.
  Bagdanov, and Joost van~de Weijer.
\newblock {Class-incremental learning: survey and performance evaluation}.
\newblock {\em CoRR}, abs/2010.15277, 2020.

\bibitem{Melnykov:mix}
Volodymyr Melnykov and Ranjan Maitra.
\newblock {Finite mixture models and model-based clustering}.
\newblock {\em Statistics Surveys}, 4:80 -- 116, 2010.

\bibitem{Parisi:2019}
German~I. Parisi, Ronald Kemker, Jose~L. Part, Christopher Kanan, and Stefan
  Wermter.
\newblock {Continual lifelong learning with neural networks: A review}.
\newblock {\em Neural Networks}, 113:54--71, 2019.

\bibitem{Patacchiola:2020coarse}
Massimiliano Patacchiola and Amos~J. Storkey.
\newblock Self-supervised relational reasoning for representation learning.
\newblock In {\em Advances in Neural Information Processing Systems (NeurIPS)},
  2020.

\bibitem{Pfulb:2021:gmmcf}
Benedikt Pf{\"u}lb and Alexander Rainer~Tassilo Gepperth.
\newblock {Overcoming Catastrophic Forgetting with Gaussian Mixture Replay}.
\newblock {\em 2021 International Joint Conference on Neural Networks (IJCNN)},
  pages 1--9, 2021.

\bibitem{Pham2022:bn}
Quang~Hong Pham, Chenghao Liu, and Steven C.~H. Hoi.
\newblock {Continual Normalization: Rethinking Batch Normalization for Online
  Continual Learning}.
\newblock {\em arXiv}, abs/2203.16102, 2022.

\bibitem{Rebuffi:2017}
Sylvestre-Alvise Rebuffi, Alexander Kolesnikov, G. Sperl, and Christoph~H.
  Lampert.
\newblock {iCaRL: Incremental Classifier and Representation Learning}.
\newblock {\em 2017 IEEE Conference on Computer Vision and Pattern Recognition
  (CVPR)}, pages 5533--5542, 2017.

\bibitem{Rios:2020tmap}
Amanda Rios and Laurent Itti.
\newblock {Lifelong Learning Without a Task Oracle}.
\newblock {\em 2020 IEEE 32nd International Conference on Tools with Artificial
  Intelligence (ICTAI)}, pages 255--263, 2020.

\bibitem{Rolnick:2019er}
David Rolnick, Arun Ahuja, Jonathan Schwarz, Timothy~P. Lillicrap, and Gregory
  Wayne.
\newblock {Experience Replay for Continual Learning}.
\newblock In {\em Advances in Neural Information Processing Systems (NeurIPS)},
  2019.

\bibitem{Su2007:init}
Ting Su and Jennifer~G. Dy.
\newblock {In search of deterministic methods for initializing K-means and
  Gaussian mixture clustering}.
\newblock {\em Intell. Data Anal.}, 11:319--338, 2007.

\bibitem{Ven:2019three}
Gido~M. van~de Ven and Andreas~Savas Tolias.
\newblock {Three scenarios for continual learning}.
\newblock {\em arXiv}, abs/1904.07734, 2019.

\bibitem{Variani:2015joint}
Ehsan Variani, Erik McDermott, and Georg Heigold.
\newblock {A Gaussian Mixture Model layer jointly optimized with discriminative
  features within a Deep Neural Network architecture}.
\newblock In {\em 2015 IEEE International Conference on Acoustics, Speech and
  Signal Processing (ICASSP)}, pages 4270--4274, 2015.

\bibitem{Veniat:2021modular}
Tom Veniat, Ludovic Denoyer, and MarcAurelio Ranzato.
\newblock Efficient continual learning with modular networks and task-driven
  priors.
\newblock In {\em International Conference on Learning Representations (ICLR)},
  2021.

\bibitem{Verwimp:2021}
Eli Verwimp, Matthias~De Lange, and Tinne Tuytelaars.
\newblock {Rehearsal revealed: The limits and merits of revisiting samples in
  continual learning}.
\newblock In {\em 2021 {IEEE/CVF} International Conference on Computer Vision
  (ICCV)}, 2021.

\bibitem{Oswald:2019hyper}
Johannes von Oswald, Christian Henning, Jo{\~{a}}o Sacramento, and Benjamin~F.
  Grewe.
\newblock Continual learning with hypernetworks.
\newblock In {\em International Conference on Learning Representations,
  {ICLR}}, 2020.

\bibitem{Wen:2020be}
Yeming Wen, Dustin Tran, and Jimmy Ba.
\newblock {BatchEnsemble: an Alternative Approach to Efficient Ensemble and
  Lifelong Learning}.
\newblock In {\em International Conference on Learning Representations (ICLR)},
  2020.

\bibitem{Xie2017:resnext}
Saining Xie, Ross~B. Girshick, Piotr Doll{\'a}r, Zhuowen Tu, and Kaiming He.
\newblock {Aggregated Residual Transformations for Deep Neural Networks}.
\newblock {\em 2017 IEEE Conference on Computer Vision and Pattern Recognition
  (CVPR)}, pages 5987--5995, 2017.

\bibitem{Yan:2021der}
Shipeng Yan, Jiangwei Xie, and Xuming He.
\newblock {DER: Dynamically Expandable Representation for Class Incremental
  Learning}.
\newblock In {\em Proceedings of the IEEE Conference on Computer Vision and
  Pattern Recognition (CVPR)}, 2021.

\bibitem{Yoon:2018den}
Jaehong Yoon, Eunho Yang, Jeongtae Lee, and Sung~Ju Hwang.
\newblock {Lifelong Learning with Dynamically Expandable Networks}.
\newblock In {\em International Conference on Learning Representations}, 2018.

\bibitem{Yu:2020sem}
Lu Yu, Bartlomiej Twardowski, Xialei Liu, Luis Herranz, Kai Wang, Yong mei
  Cheng, Shangling Jui, and Joost van~de Weijer.
\newblock {Semantic Drift Compensation for Class-Incremental Learning}.
\newblock {\em 2020 IEEE/CVF Conference on Computer Vision and Pattern
  Recognition (CVPR)}, pages 6980--6989, 2020.

\bibitem{Zenke:2017}
Friedemann Zenke, Ben Poole, and Surya Ganguli.
\newblock {Continual Learning through Synaptic Intelligence}.
\newblock In {\em Proceedings of the 34th International Conference on Machine
  Learning (ICML)}, 2017.

\bibitem{Zhou:2022bn}
Minghao Zhou, Quanziang Wang, Jun Shu, Qian Zhao, and Deyu Meng.
\newblock {Diagnosing Batch Normalization in Class Incremental Learning}.
\newblock {\em arXiv}, abs/2202.08025, 2022.

\end{thebibliography}
}

\newpage
\appendix

\section{Appendix}
\label{sec:igmm-appx-a}

\subsection{Data}
\label{sec:igmm-data}

We used: MNIST, FASHION, SVHN, CIFAR10 and IMAGENET10 -- a subset of the tiny IMAGENET200, to gain deeper insights into our method while conducting experiments with hundreds of different configurations. Then, we extended this set with CIFAR20 -- the coarse-grained version of CIFAR100, IMAGENET20A and IMAGENET20B -- larger subsets of IMAGENET200 -- to benchmark our method against other algorithms. 

For the experiments involving fixed extractors, we used pre-trained features to construct four additional sequences -- CIFAR100-PRE10, CIFAR100-PRE100, IMAGENET200-PRE20 and IMAGENET200-PRE200, which consisted of features extracted for CIAFR100 and IMAGENET200, using extractors trained on 10, 20, 100 and 200 classes of the original datasets. The summary of the used benchmarks is given in Tab. \ref{tab:igmm-data}. Details of the feature extractors can be found in the next section.

\begin{table}[htb!]
	\caption{Summary of used datasets.}
	\vspace{0.1cm}
	\centering
	\scalebox{0.8}{
		\begin{tabular}[H]{l >{\centering\arraybackslash} m{1.4cm} >{\centering\arraybackslash} m{1.4cm} >{\centering\arraybackslash} m{0.8cm} >{\centering\arraybackslash} m{0.8cm}}
			\toprule	
			Dataset & Train & Test & Cls & Feats\\
			\midrule
			MNIST & 50 000 & 10 000 & 10 & No\\
			FASHION & 60 000 & 10 000 & 10 & No\\
			SVHN & 73 257 & 26 032 & 10 & No\\
			IMAGENET10 & 5000 & 500 & 10 & No\\
			CIFAR10 & 50 000 & 10 000 & 10 & No\\
			IMAGENET20A & 10 000 & 1000 & 20 & No\\
			IMAGENET20B & 10 000 & 1000 & 20 & No\\
			CIFAR20 & 50 000 & 10 000 & 20 & No\\
			\midrule
			CIFAR100-PRE10 & 50 000 & 10 000 & 100 & 128\\
			CIFAR100-PRE100 & 50 000 & 10 000 & 100 & 512\\
			IMAGENET200-PRE20 & 100 000 & 10 000 & 200 & 256\\
			IMAGENET200-PRE200 & 100 000 & 10 000 & 200 & 256\\
			\bottomrule
		\end{tabular}
	}
	\label{tab:igmm-data}
\end{table}

\subsection{Model configurations}

In the first section of our experiments, we explored different configurations of our algorithm, which can be mostly seen as an ablation study. Firstly, we evaluated different \textbf{losses} (CE, MC and MCR) combined with different \textbf{classification methods} (softmax, max-component). Secondly, we checked different settings for the \textbf{tightness bound} parameter $\tau_p$ by evaluating a grid of values for inter-tightness and intra-tightness --  we considered $\tau_p \in \langle$1e-06, 1e-05, 0.0001, 0.001, 0.01$\rangle$ for both. Thirdly, we analyzed how assuming different \textbf{numbers of components} affects the classification performance on different datasets. We used $K \in \langle$1, 3, 5, 10, 20$\rangle$. Then we checked if it is better to maintain a whole \textbf{covariance matrix} or only its variance (FULL, VAR).  Finally, we evaluated different \textbf{learning rates} for the extractor and GMM part, using $\alpha_{\mathcal{F}} \in \langle$1e-07, 1e-06, 1e-05, 0.0001, 0.001$\rangle$ and $\alpha_{\mathcal{G}} \in \langle$1e-05, 0.0001, 0.001, 0.01, 0.1$\rangle$, to check whether it may be beneficial to configure them separately, and different \textbf{memory sizes} $\mathcal{M}_c \in \langle$8, 64, 128, 256, 512$\rangle$ to analyze how our method exploits limited access to class examples.

While evaluating specific parameters we kept others fixed. For our base configuration we chose a setup that was capable of providing performance comparable with a standard experience replay. We used the MCR with max-component as our loss and classification method, $K=3$, $\tau_{p,ie}=0.002$, $\tau_{p,ia}=0.01$, $\beta=0.5$, $\alpha_{\mathcal{F}}=$0.0001, $\alpha_{\mathcal{G}}=$0.001 and $d_{min}=0.001$ with only variance stored per each component. We assumed a modest memory buffer per class $\mathcal{M}_c=256$ and matched the size of a memory sample per class with the training batch size. The model was trained for 10 (MNIST, FASHION) or 25 epochs per class, with 32 (IMAGENET) or 64 instances in a mini-batch.

\subsection{Algorithms}

\begin{figure*}[tb]
	\centering
	\setlength{\fboxrule}{0.5pt}
	\setlength{\fboxsep}{0pt}
	\begin{subfigure}{0.15\linewidth}
		\centering
		\stackunder[5pt]{\fbox{\includegraphics[trim=85 120 40 12,clip,width=\linewidth]{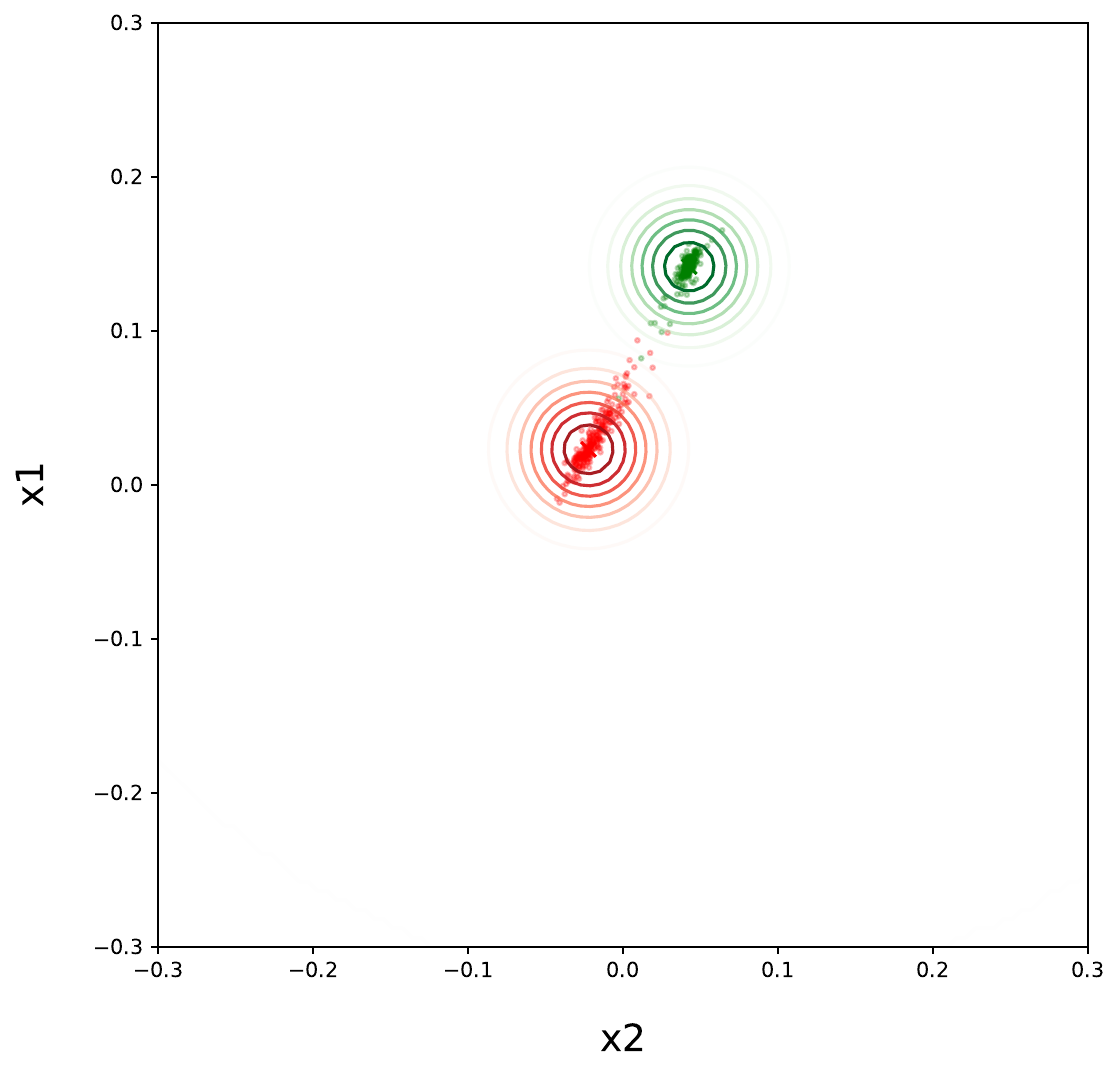}}}{$t$=2}
	\end{subfigure}\hspace{0.05cm}
	\begin{subfigure}{0.15\linewidth}
		\centering
		\stackunder[5pt]{\fbox{\includegraphics[trim=76 100 50 35,clip,width=\linewidth]{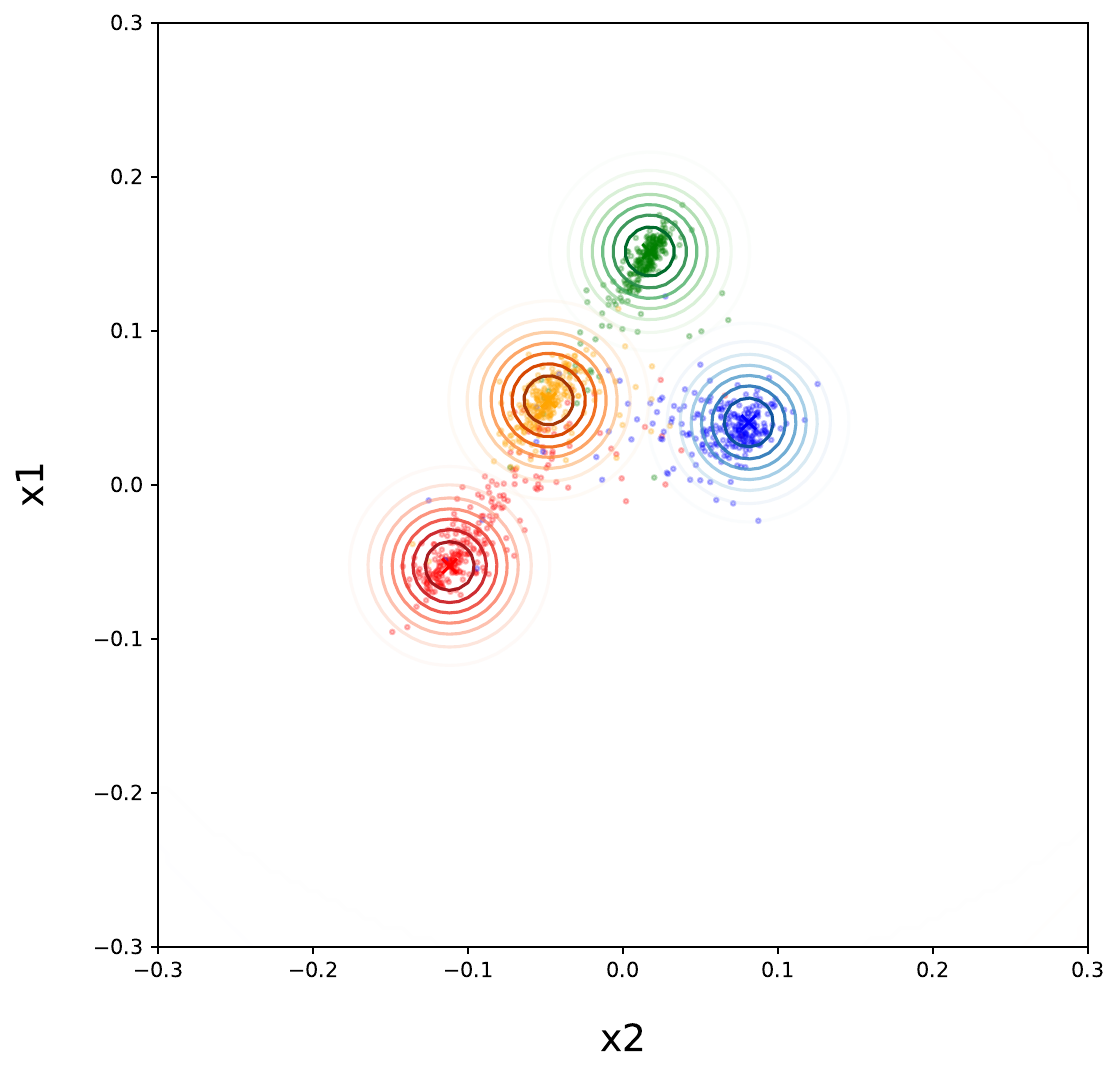}}}{$t$=4}
	\end{subfigure}\hspace{0.05cm}
	\begin{subfigure}{0.15\linewidth}
		\centering
		\stackunder[5pt]{\fbox{\includegraphics[trim=100 100 60 65,clip,width=\linewidth]{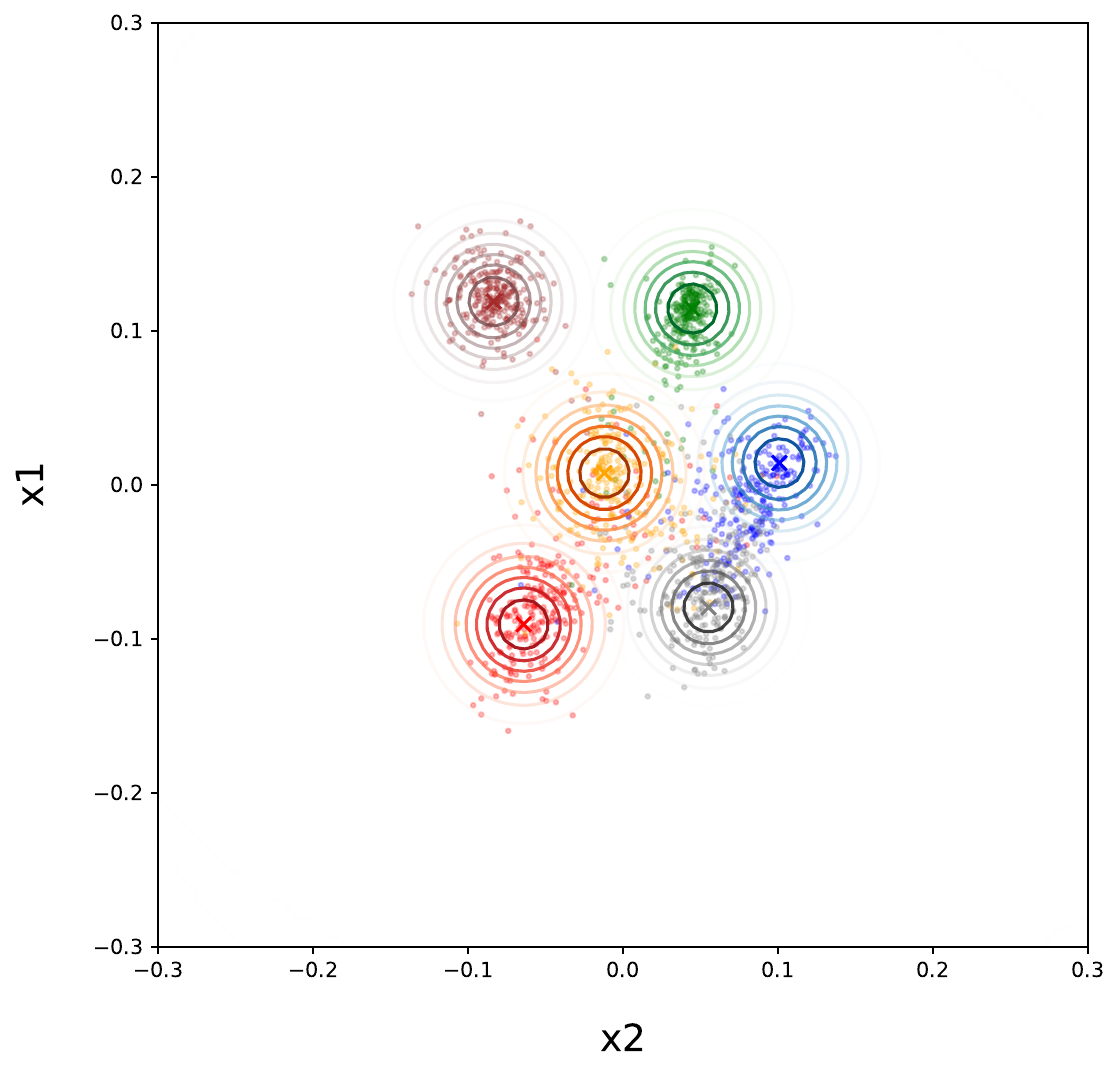}}}{$t$=6}
	\end{subfigure}\hspace{0.05cm}
	\begin{subfigure}{0.15\linewidth}
		\centering
		\stackunder[5pt]{\fbox{\includegraphics[trim=76 90 39 30,clip,width=\linewidth]{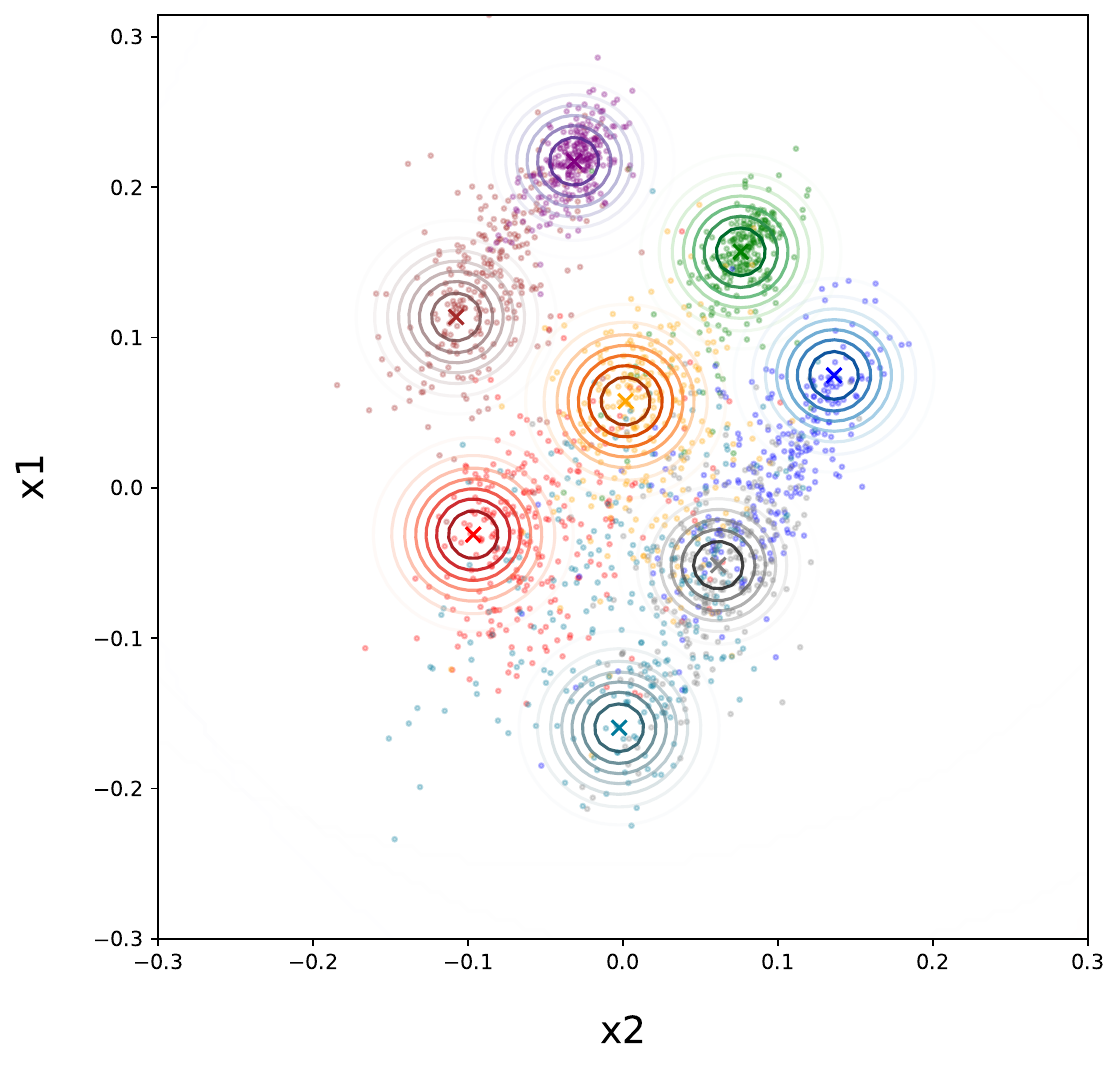}}}{$t$=8}
	\end{subfigure}\hspace{0.05cm}
	\begin{subfigure}{0.15\linewidth}
		\centering
		\stackunder[5pt]{\fbox{\includegraphics[trim=76 85 25 25,clip, width=\linewidth]{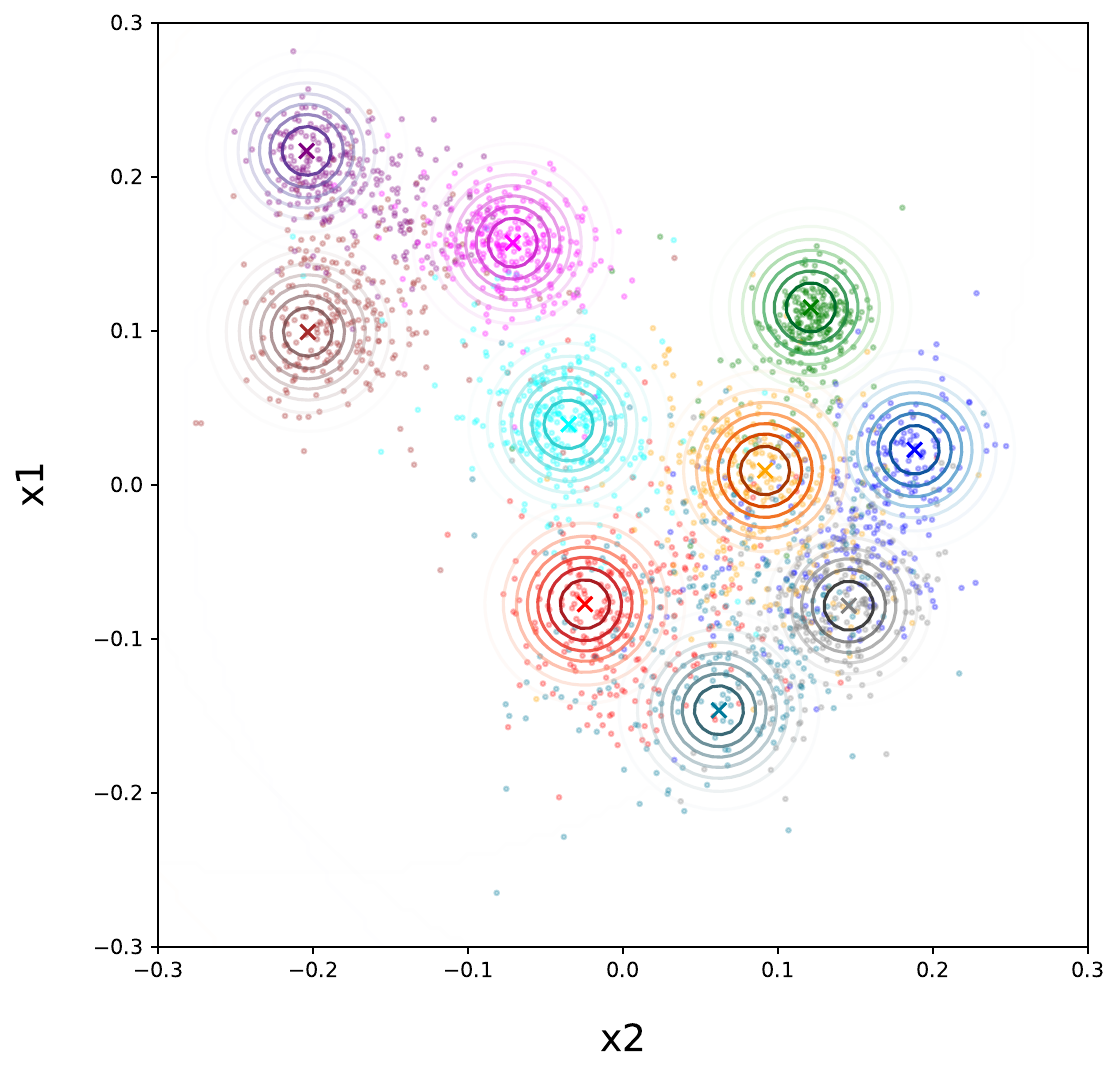}}}{$t$=10}
	\end{subfigure}
	\caption[Learning subsequent classes incrementally with the inter-contrastive loss utilizing the tightness bound]{Learning subsequent classes of FASHION incrementally ($K$=1) with the inter-contrastive loss utilizing the tightness bound ($\tau_{p, ie}$=0.2).}
	\label{fig:igmm-mp}
\end{figure*}

\begin{figure*}[tb]
	\centering
	\setlength{\fboxrule}{0.5pt}
	\setlength{\fboxsep}{0pt}
	\begin{subfigure}{0.15\linewidth}
		\centering
		\stackunder[5pt]{\fbox{\includegraphics[trim=95 100 40 39,clip,width=\linewidth]{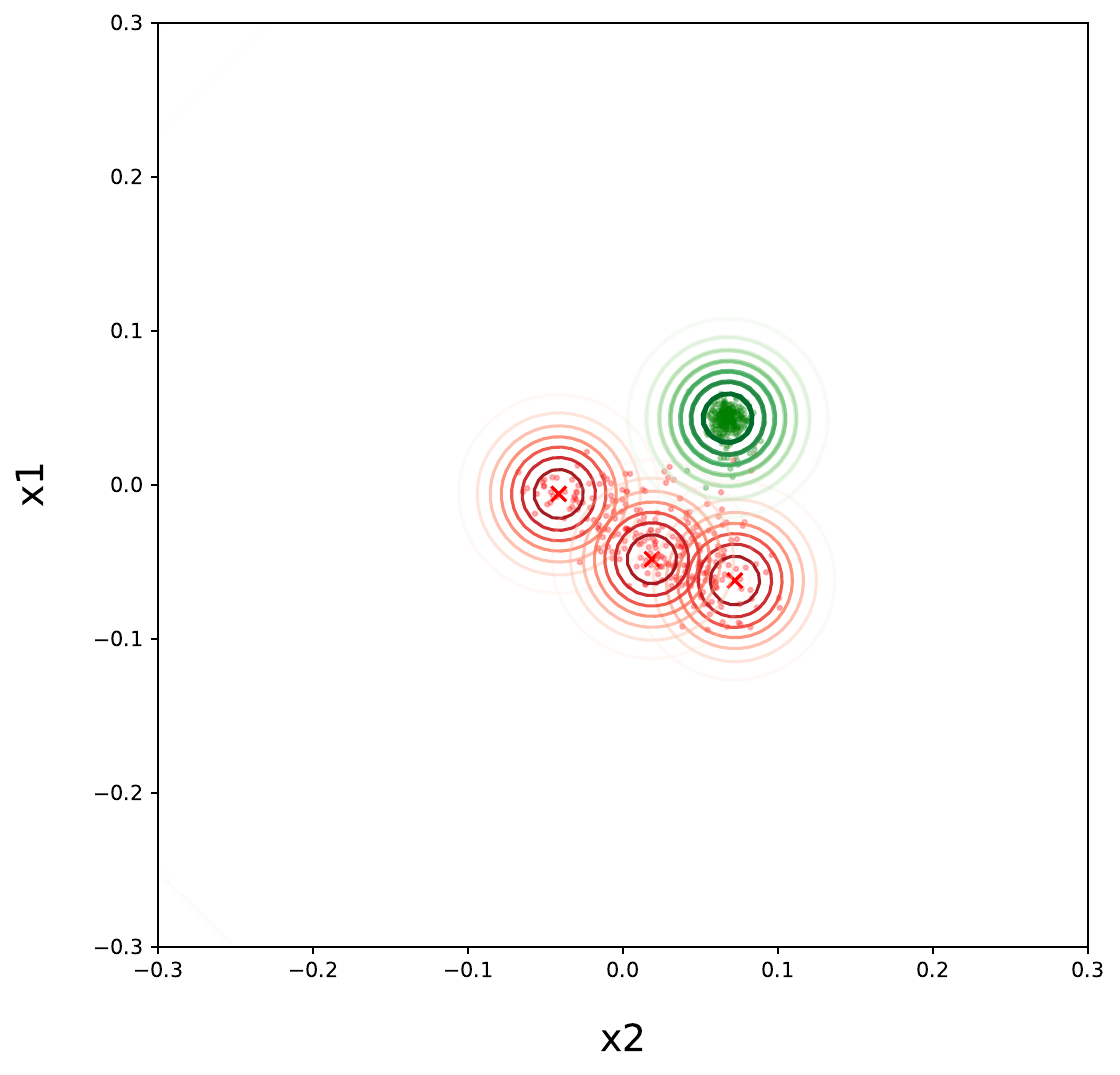}}}{$t$=2}
	\end{subfigure}\hspace{0.05cm}
	\begin{subfigure}{0.15\linewidth}
		\centering
		\stackunder[5pt]{\fbox{\includegraphics[trim=100 100 35 40,clip,width=\linewidth]{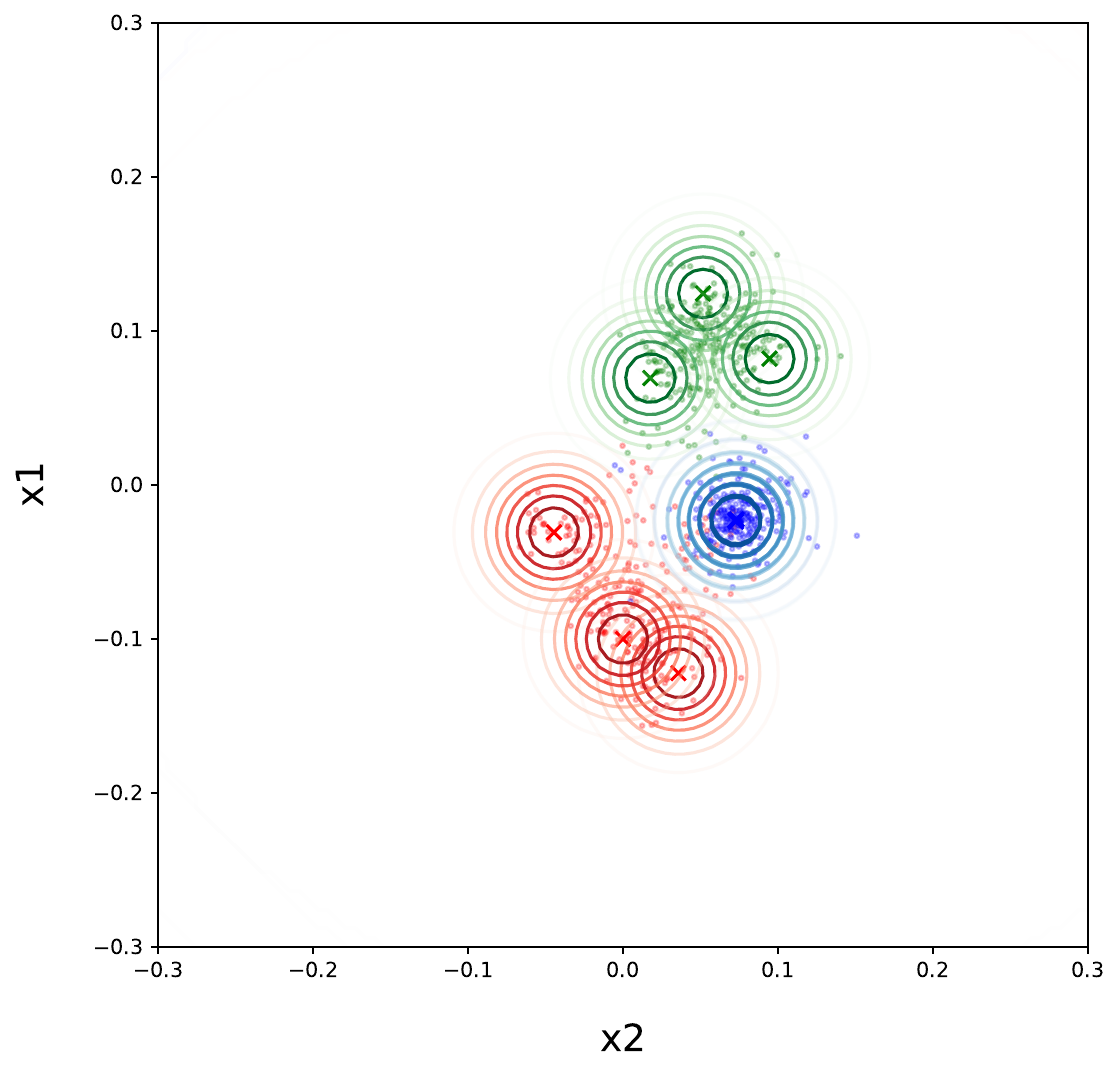}}}{$t$=3}
	\end{subfigure}\hspace{0.05cm}
	\begin{subfigure}{0.15\linewidth}
		\centering
		\stackunder[5pt]{\fbox{\includegraphics[trim=100 100 35 38,clip,width=\linewidth]{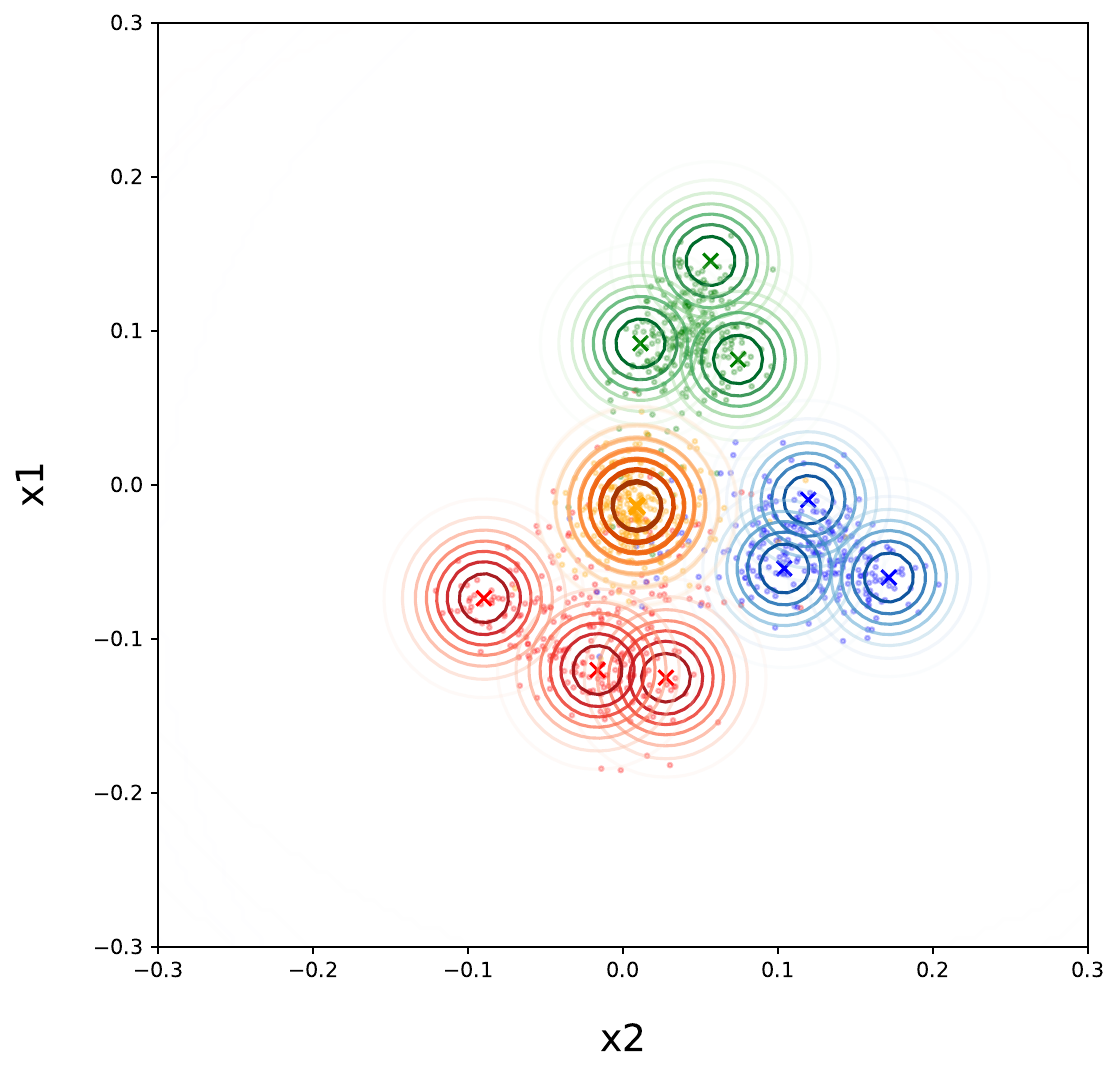}}}{$t$=4}
	\end{subfigure}\hspace{0.05cm}
	\begin{subfigure}{0.15\linewidth}
		\centering
		\stackunder[5pt]{\fbox{\includegraphics[trim=76 85 30 25,clip,width=\linewidth]{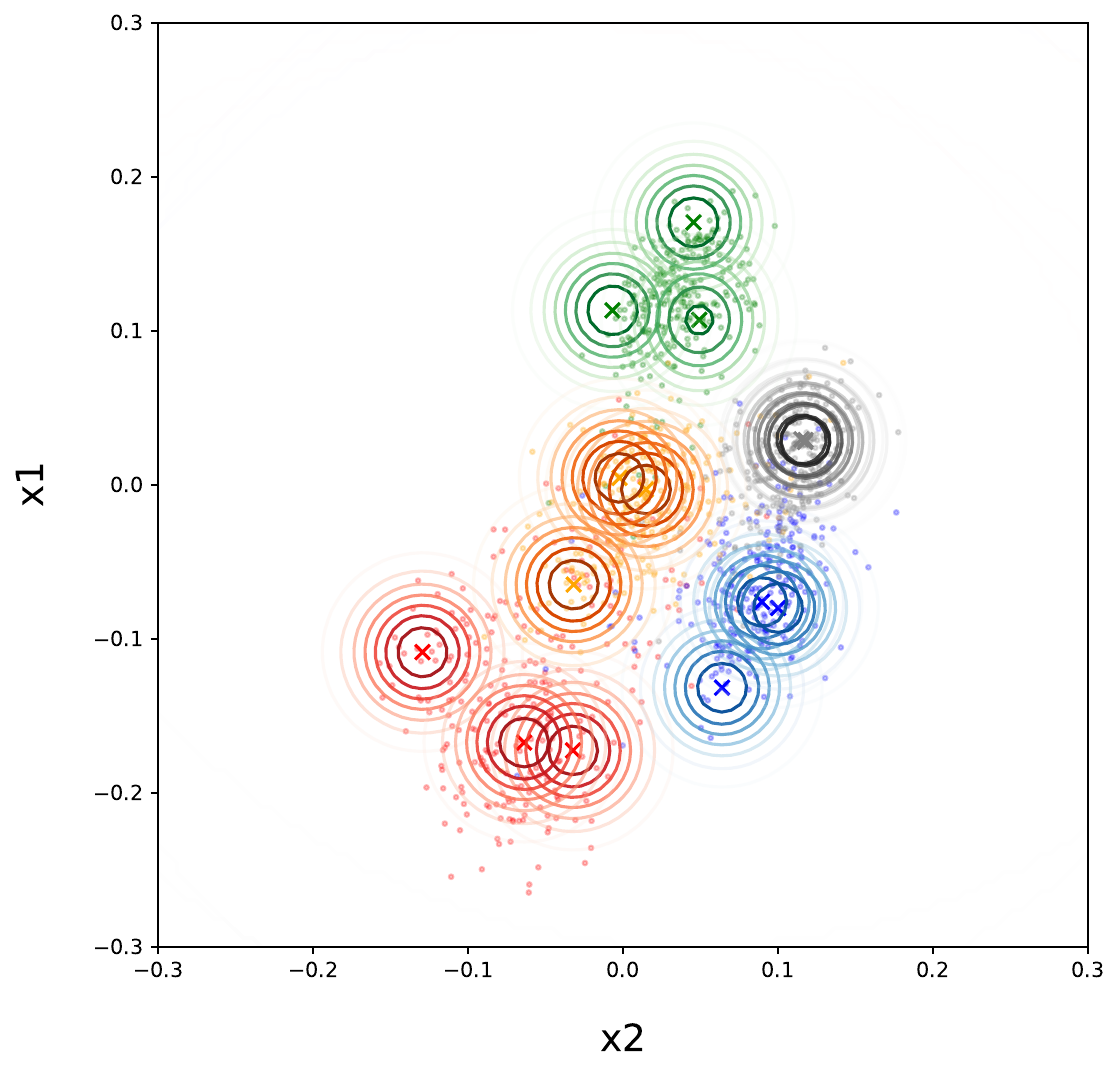}}}{$t$=5}
	\end{subfigure}\hspace{0.05cm}
	\begin{subfigure}{0.15\linewidth}
		\centering
		\stackunder[5pt]{\fbox{\includegraphics[trim=76 76 15 20,clip, width=\linewidth]{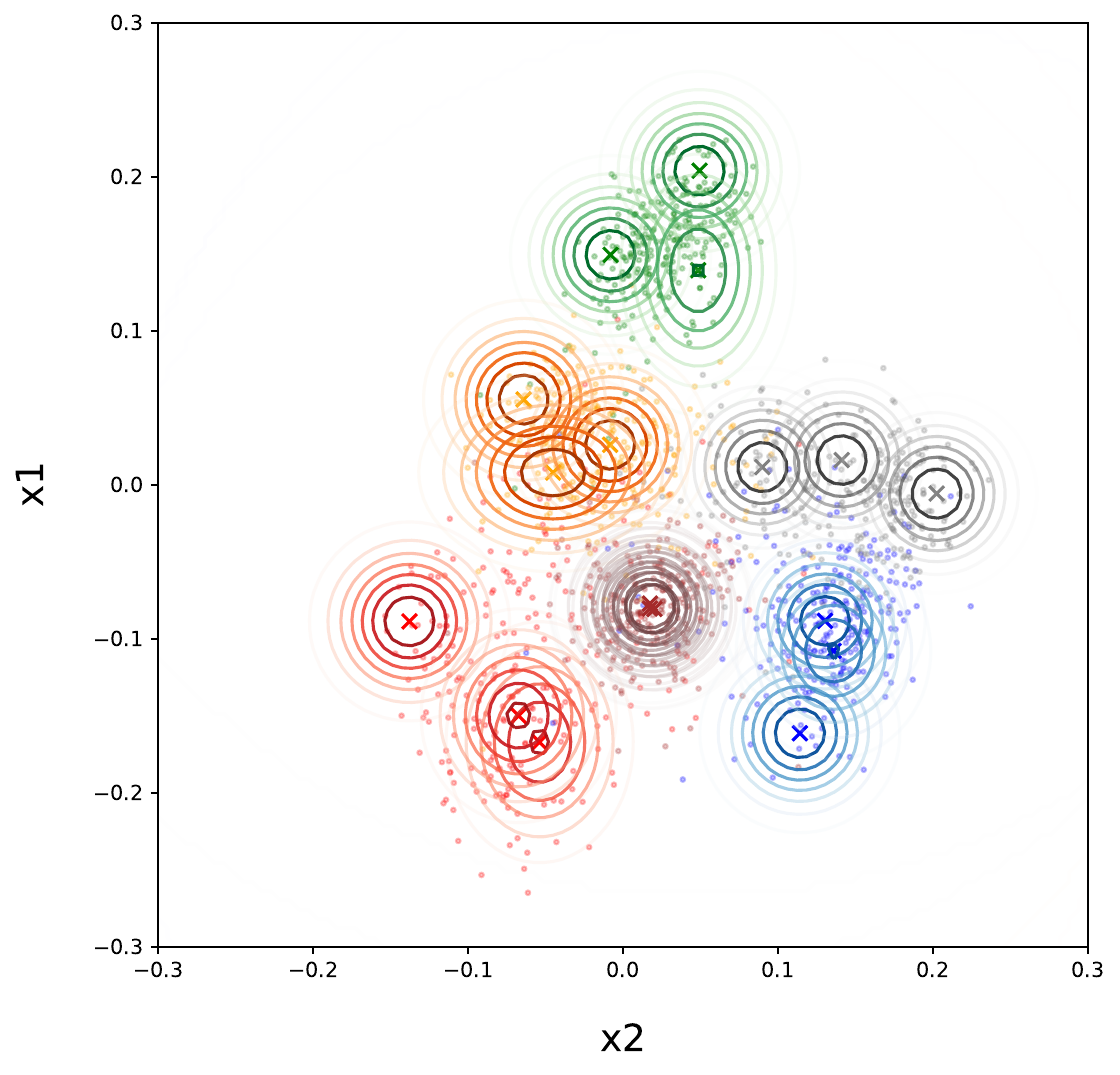}}}{$t$=6}
	\end{subfigure}
	\caption[Learning subsequent classes incrementally with regionalization and the intra-contrastive loss utilizing the tightness bound]{Learning subsequent classes of FASHION incrementally ($K$=3) with regionalization and the intra-contrastive loss utilizing the tightness bound ($\tau_{p, ia}$=0.25).}
	\label{fig:igmm-mpr}
\end{figure*}

\begin{figure*}[tb]
	\centering
	\setlength{\fboxrule}{0.5pt}
	\setlength{\fboxsep}{0pt}
	\begin{subfigure}{0.15\linewidth}
		\centering
		\stackunder[5pt]{\fbox{\includegraphics[trim=100 70 25 55,clip,width=\linewidth]{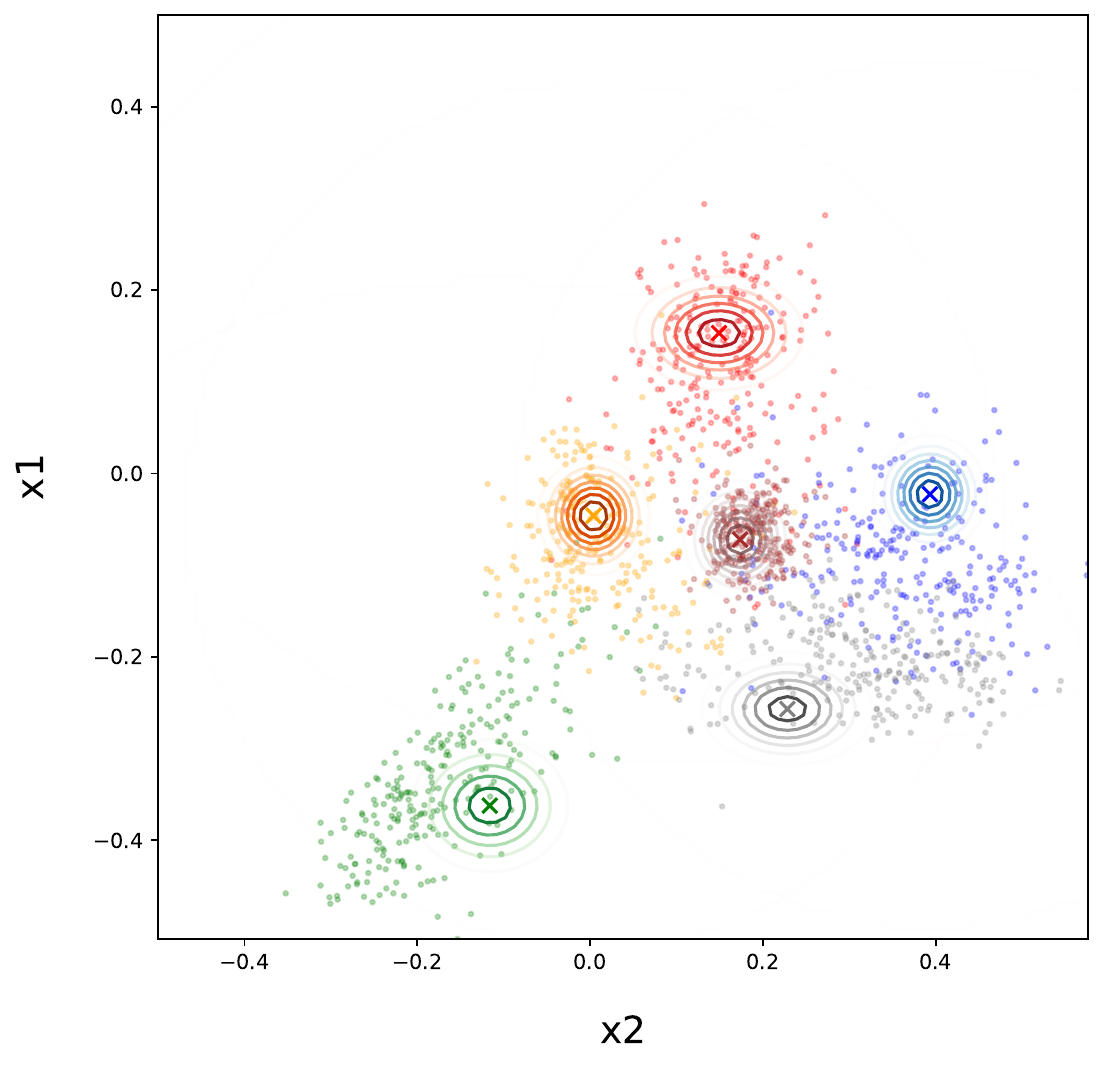}}}{CE $\rightarrow$ 0.84}
	\end{subfigure}\hspace{0.05cm}
	\begin{subfigure}{0.15\linewidth}
		\centering
		\stackunder[5pt]{\fbox{\includegraphics[trim=100 70 25 55,clip,width=\linewidth]{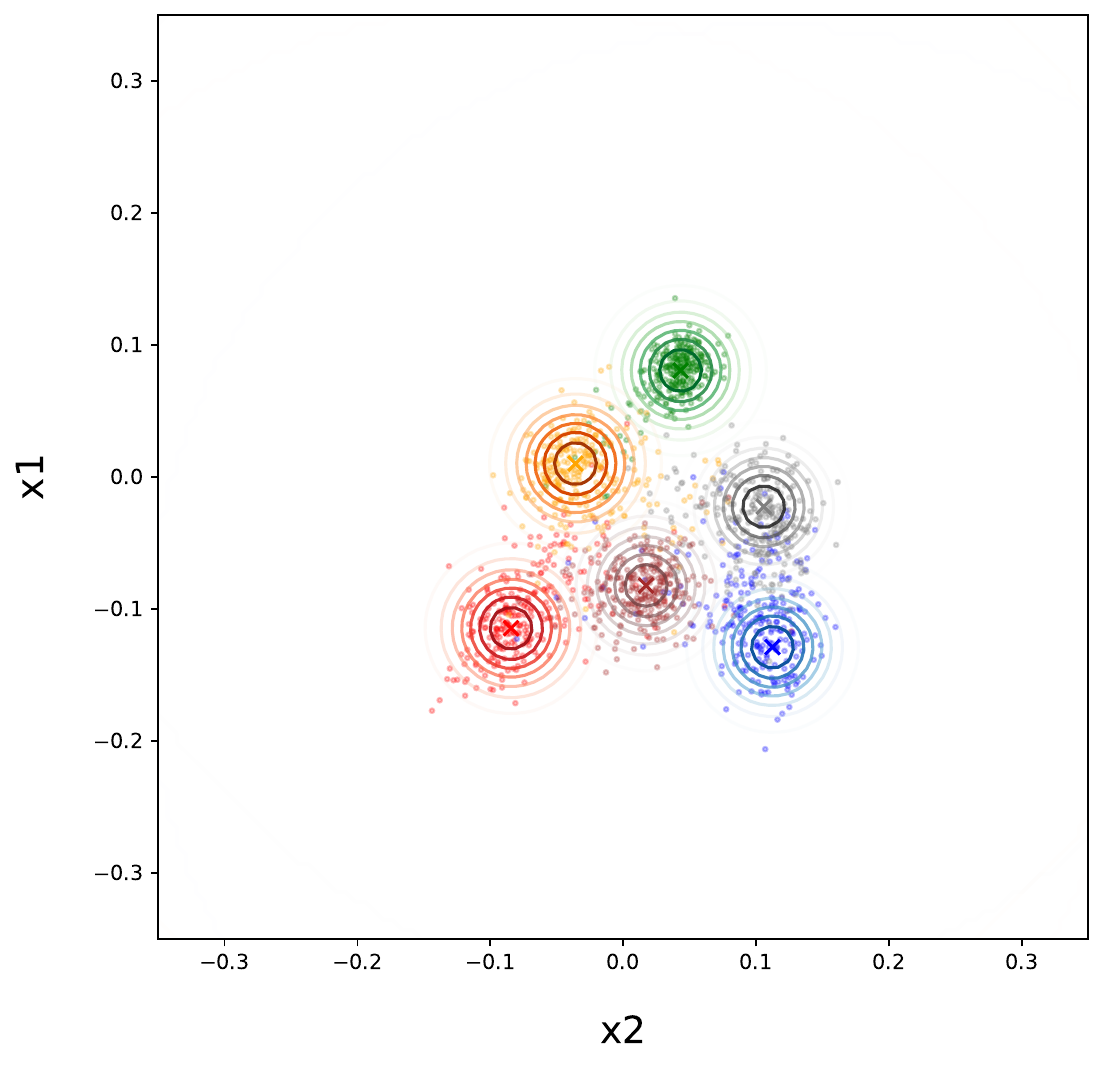}}}{MC $\rightarrow$ 0.85}
	\end{subfigure}\hspace{0.05cm}
	\begin{subfigure}{0.15\linewidth}
		\centering
		\stackunder[5pt]{\fbox{\includegraphics[trim=76 72 15 22,clip,width=\linewidth]{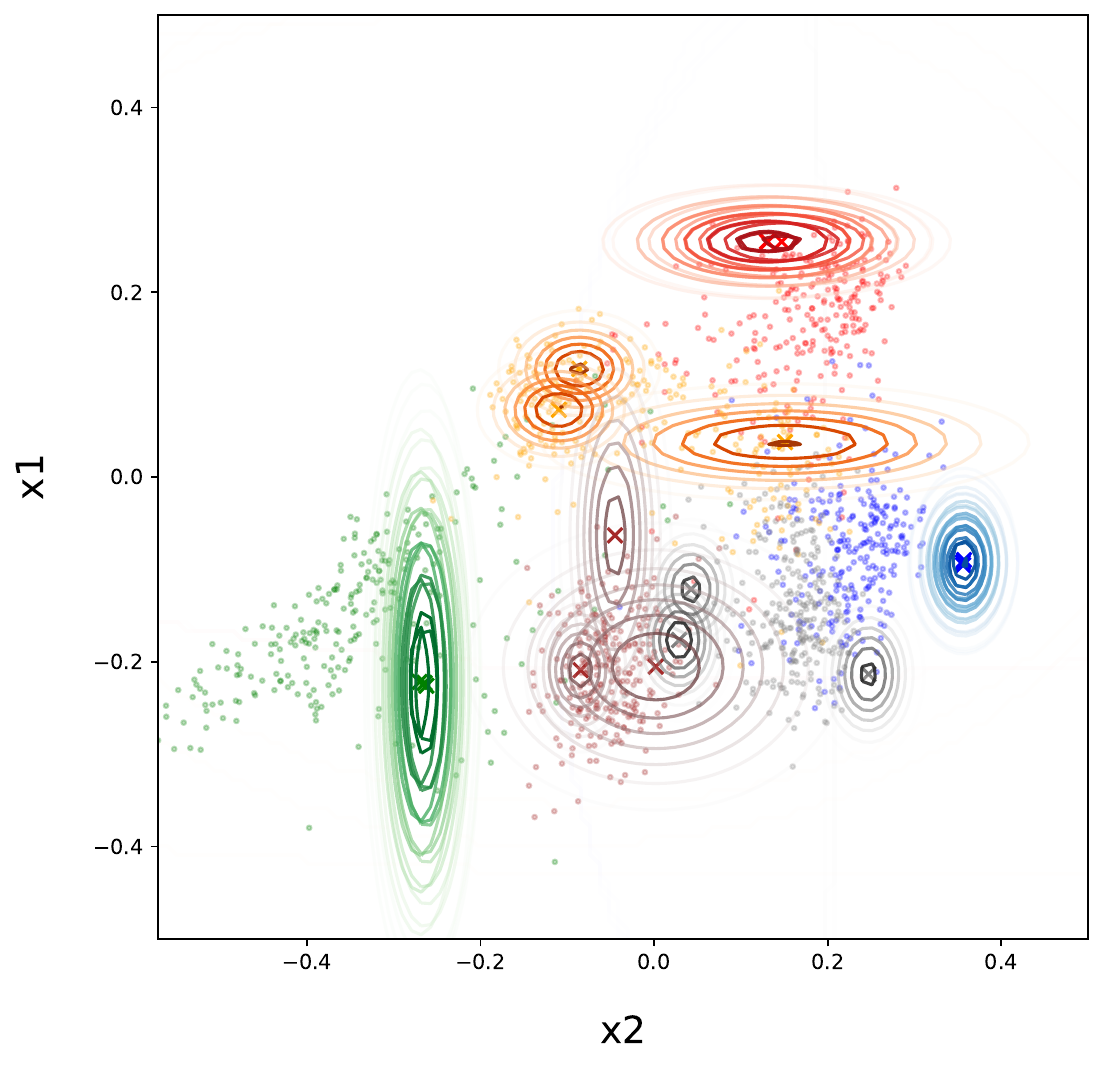}}}{CE $\rightarrow$ 0.84}
	\end{subfigure}\hspace{0.05cm}
	\begin{subfigure}{0.15\linewidth}
		\centering
		\stackunder[5pt]{\fbox{\includegraphics[trim=75 75 75 75,clip,width=\linewidth]{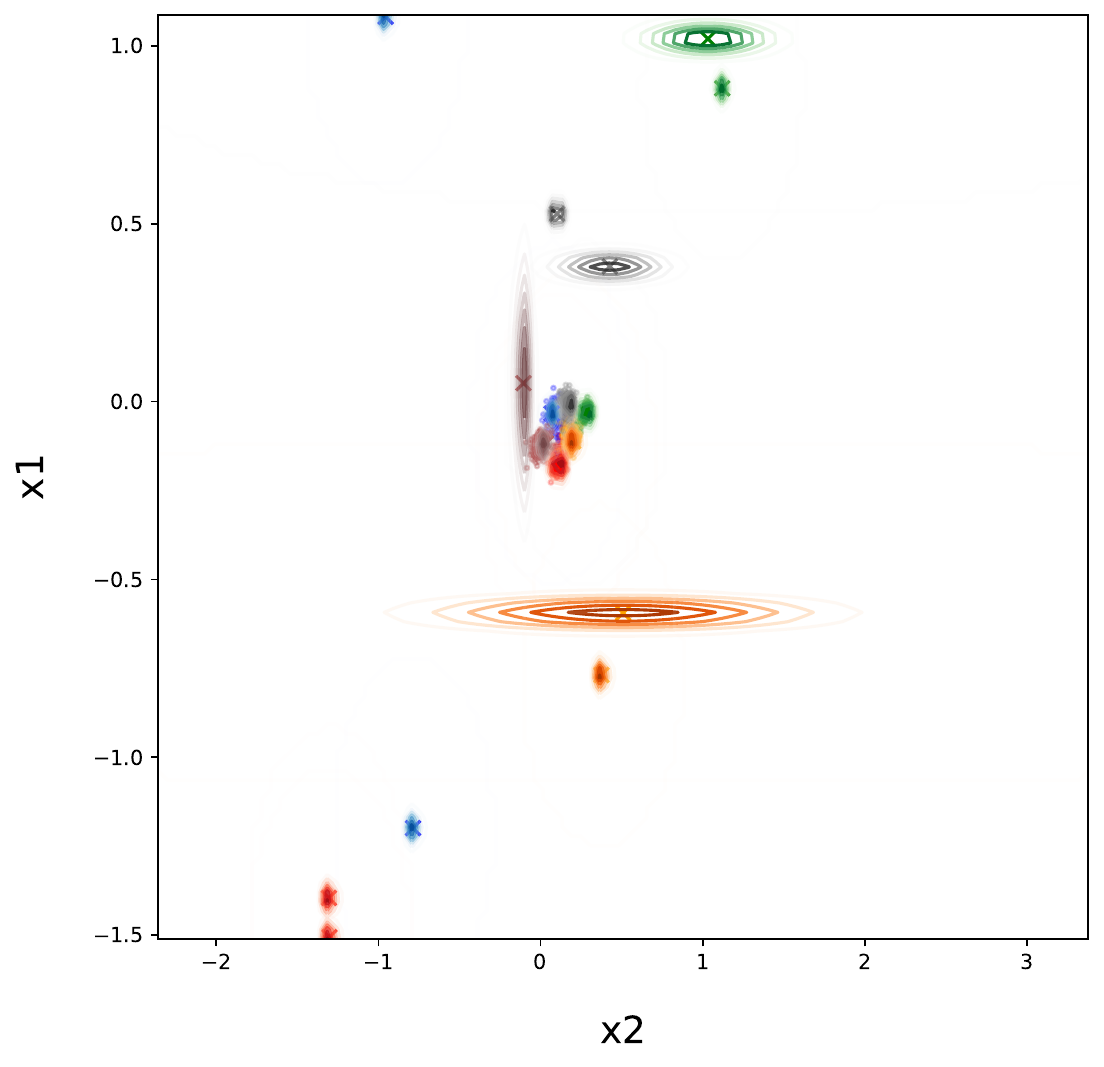}}}{MC $\rightarrow$ \textbf{0.86}}
	\end{subfigure}
	\caption[Mixtures learned with cross-entropy and simple max-component strategy]{Mixtures learned with cross-entropy and simple max-component strategy ($K$=1 and $K$=3) after 6 classes of FASHION.}
	\label{fig:igmm-cemp}
\end{figure*}

Based on the observations made in the first section of the experiments, in the final evaluation we used two variants of our algorithm: \textbf{MIX-CE} and \textbf{MIX-MCR} with $\tau_{p,ie}=$0.0001, $\tau_{p,ia}=$0.001, $\alpha_{\mathcal{F}}=$0.0001, $\alpha_{\mathcal{G}}=$1e-05 and, once again, $d_{min}=0.001$ with only variance maintained per each component. The only parameter that we tuned per each dataset was the number of components $K$. We used Adam as the optimizer. For the memory-free scenarios with pre-trained extractors, we turned off the inter-contrastive loss to minimize interference with previously learned classes.

The main parameters of the baselines methods were set based on the original papers and other literature, including empirical surveys or works containing vast empirical studies \cite{Buzzega2020:dark, Chaudhry:2019, Korycki:2021ercd, Maltoni:2019sit, Masana:2020, Ven:2019three}. For all memory sampling methods we matched the memory sampling size with the training batch size. For ERSB we used 10 centroids per class each containing up to either 25 or 15 instances to match the total memory size. DER used $\alpha_d$=0.5, for LWF we set the softmax temperature $T=2$ and progressively increased its distillation coefficient as suggested in \cite{Maltoni:2019sit}, and SI used $\lambda =$0.0001. All of the methods utilized the Adam optimizer with a learning rate $\alpha$=0.0001 as we did not observe any significant differences when changing this parameter.

Analogously to the configuration section, all of the algorithms, including ours, were trained for 10 (MNIST, FASHION) or 25 epochs per class, using 32 (IMAGENET) or 64 instances per mini-batch. The offline models were trained for either 50 or 100 epochs, until they achieved a saturation level. The memory buffer was set to $\mathcal{M}_c=128$ (IMAGENET) or $\mathcal{M}_c=256$ for methods supporting memory per class (ER, ERSB, iCaRL), and $\mathcal{M}=C\cdot128$ or $\mathcal{M}=C\cdot256$ for the remaining ones (GSS, A-GEM, DER), where $C$ was the total number of classes. The latter group was equipped with reservoir buffers \cite{Buzzega2020:dark}. For the experiments with pre-trained extractors we wanted to check the \textbf{memory-free scenario}, therefore we set $\mathcal{M}_c=0$ for our methods and $\mathcal{M}_c=1$ or $\mathcal{M}=C$ for others, since most of them could not be run without storing any examples.

All of the algorithms, including different configurations of our method, were combined with feature extractors. For MNIST and FASHION we used a simple CNN with two convolutional layers consisting of 32 (5x5) and 64 (3x3) filters, interleaved with ReLU, batch normalization and max pooling (2x2). For SVHN and IMAGENET we utilized ResNet18, its modified version for CIFAR10 and CIFAR20, and ResNeXt29 for CIFAR100 \cite{Xie2017:resnext}. The classification layers consisted of the default configurations. 

Finally, for our method, ER, ERSB, A-GEM and DER we disabled batch normalization, since, consistently with \cite{Pham2022:bn, Zhou:2022bn}, we observed a significant difference in performance when those layers were turned off for the given methods. As mentioned in Sec. \ref{sec:igmm-data}, for the memory-free scenarios, the extractors were pre-trained on either 10, 20, 100 or 200 classes of CIFAR100 and IMAGENET200. For this setting we trained all the models for 20 epochs per class.

Results for the offline model were either obtained by us (learned from scratch for IMAGENET20A, IMAGENET20B and fine-tuned models for IMAGENET200), or by referring to other publications \cite{Korycki:2021ercd, Patacchiola:2020coarse}.

\section{Appendix}
\label{sec:igmm-appx-b}

\subsection{Additional visualizations}
\label{sec:igmm-appx-b-vis}

Fig. \ref{fig:igmm-mp} presents an example of a single-component class-incremental mixture model learned with the inter-contrastive loss. Fig. \ref{fig:igmm-mpr} demonstrates the effectiveness of training a multi-component model with the intra-contrastive loss and regionalization.

As mentioned in the main document, the CE loss can often achieve similar predictive performance even if its mixture models are not really fitting the data (Fig. \ref{fig:igmm-cemp}). We can see it when compared with MC for $K$=1 or MCR for both $K$ (Fig. \ref{fig:igmm-mp} and \ref{fig:igmm-mpr}). Furthermore, the model produced for MC with $K$=3 clearly shows that it is incapable of effectively utilizing multiple components for the same class. Please notice that only the Gaussians in the middle actually cover some data points, while the remaining components are completely unrelated to the observed data. These are examples of the degenerate solutions. While for FASHION this loss could still, analogously to CE, provide similar performance as MCR (the components in the middle are fitted to the data and they are sufficient to model it), the observed desynchronization of components results in its weaknesses for more complex problems. The MCR loss can provide high quality of predictive performance and of the produced mixture models.

\subsection{Additional configurations}

\paragraph{Number of components:} Tab. \ref{tab:igmm-ks} presents how many components were required to obtain the best solutions per each dataset for the given settings. We can observe that for simpler datasets (MNIST, FASHION) using a single component per class for sufficient and that introducing additional ones led to slightly worse performance, most likely due to the fact of fitting to simple concepts and overcomplicating the optimization problem. On the other hand, more complex benchmarks (SVHN, CIFAR10, IMAGENET10) preferred access to more components per class, which could provide significant improvements, e.g., for SVHN the difference between $K$=1 and $K$=10 was almost 0.3. While for these experiments we set the learning rate slightly higher for the GMM model (0.001) than for the extractor (0.0001), we observed that when the former used rate lower than the latter (as suggested by the results for learning rates that will be presented below), the optimal $K$ tended to be lower on average. It is possible that if GMM is dominant it prefers having more flexibility (components), while when the extractor has a higher learning rate it may be more effective in adjusting representations to lower numbers of components.

\begin{table}[h]
	\caption[Average incremental accuracy for MIX using different numbers of components]{Average incremental accuracy for MIX using different numbers of components $K$.}
	\centering
	\scalebox{0.78}{
		\begin{tabular}[H]{l >{\centering\arraybackslash} m{1.4cm} >{\centering\arraybackslash} m{1.4cm} >{\centering\arraybackslash} m{1.4cm} >{\centering\arraybackslash} m{1.4cm} >{\centering\arraybackslash} m{1.4cm}}
			\toprule	
			\textit{Config} & MNIST & FASHION & SVHN & CIFAR10 & IMGNET10\\
			\midrule
			$K$=1 & \textbf{0.9885} & \textbf{0.8859} & 0.4862 & 0.4282 & 0.6466 \\
			$K$=3 & 0.9875 & 0.8782 & 0.5978 & 0.5407 & 0.6584 \\
			$K$=5 & 0.9463 & 0.8562 & 0.6994 & 0.5522 & \textbf{0.6604} \\
			$K$=10 & 0.9393 & 0.8577 & \textbf{0.7438} & \textbf{0.5620} & 0.6252 \\
			$K$=20 & 0.9521 & 0.8517 & 0.6868 & 0.5532 & 0.4270 \\
			\toprule
		\end{tabular}
	}
	\label{tab:igmm-ks}
\end{table}

\paragraph{Covariance:} Results presented in Tab. \ref{tab:igmm-covs}, unequivocally show that our gradient-based MIX can much better adapt to data if it maintains only the variance of the covariance matrix (better by almost 0.3 when compared with full covariance). It is not surprising since previous publications related to the gradient-based GMMs for offline settings suggested a similar thing \cite{Gepperth2021:grad}. Most likely, working with a full covariance matrix leads to less stable loss values, and many more free parameters (especially if the feature space is high-dimensional) likely cause problems with convergence.

\begin{table}[h]
	\caption[Average incremental accuracy for MIX with diagonal and full covariance]{Average incremental accuracy for MIX with diagonal and full covariance.}
	\centering
	\scalebox{0.78}{
		\begin{tabular}[H]{l >{\centering\arraybackslash} m{1.4cm} >{\centering\arraybackslash} m{1.4cm} >{\centering\arraybackslash} m{1.4cm} >{\centering\arraybackslash} m{1.4cm} >{\centering\arraybackslash} m{1.4cm}}
			\toprule	
			\textit{Config} & MNIST & FASHION & SVHN & CIFAR10 & IMGNET10 \\
			\midrule
			FULL & 0.7304 & 0.6577 & 0.2931 & 0.3298 & 0.3255\\
			VAR & \textbf{0.9888} & \textbf{0.8849} & \textbf{0.6393} & \textbf{0.5777} & \textbf{0.6865}\\
			\toprule
		\end{tabular}
	}
	\label{tab:igmm-covs}
\end{table}

\begin{figure*}[tb]
	\centering
	\begin{subfigure}{0.19\linewidth}
		\centering
		\includegraphics[width=\linewidth]{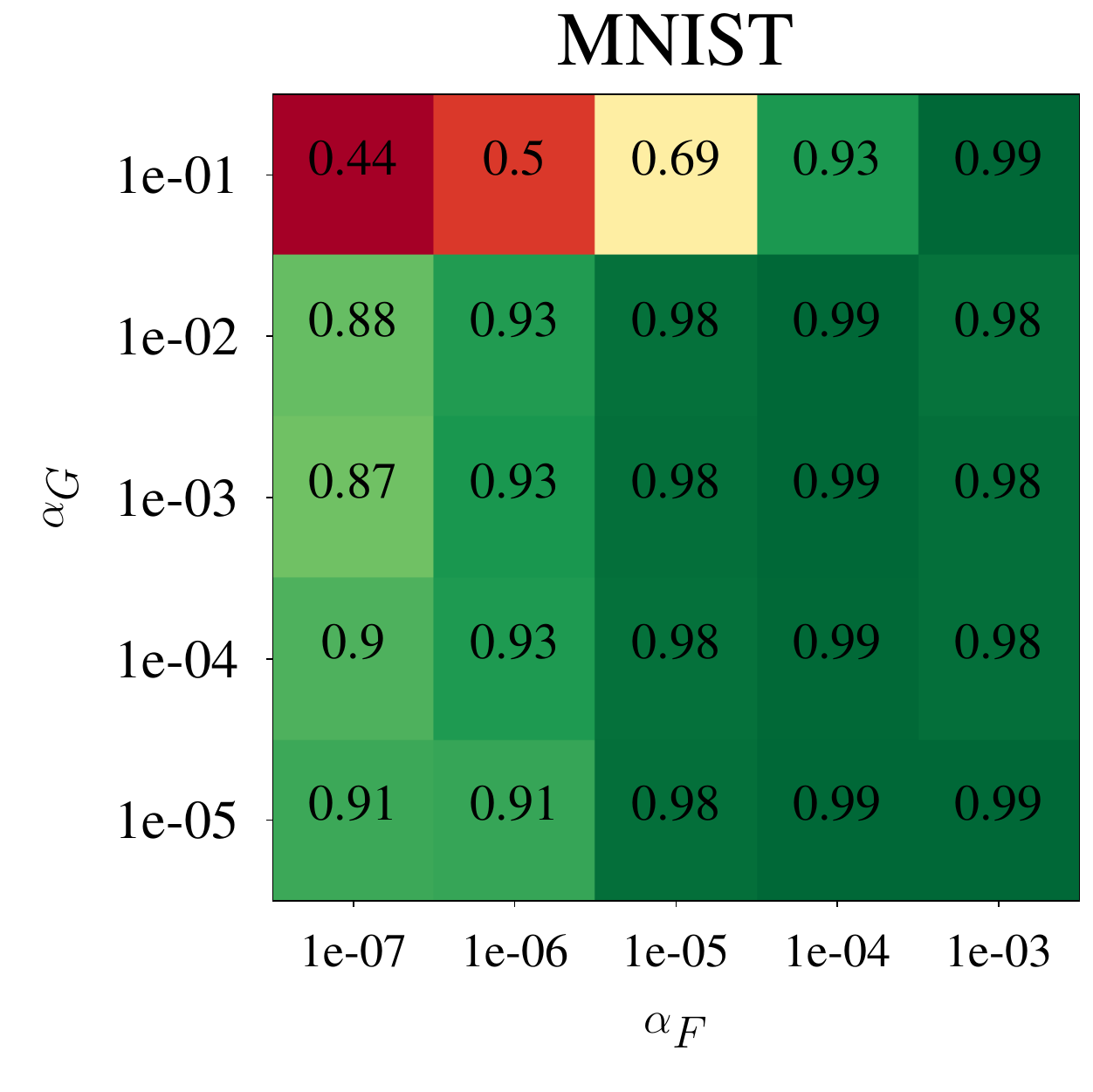}
	\end{subfigure}
	\begin{subfigure}{0.19\linewidth}
		\centering
		\includegraphics[width=\linewidth]{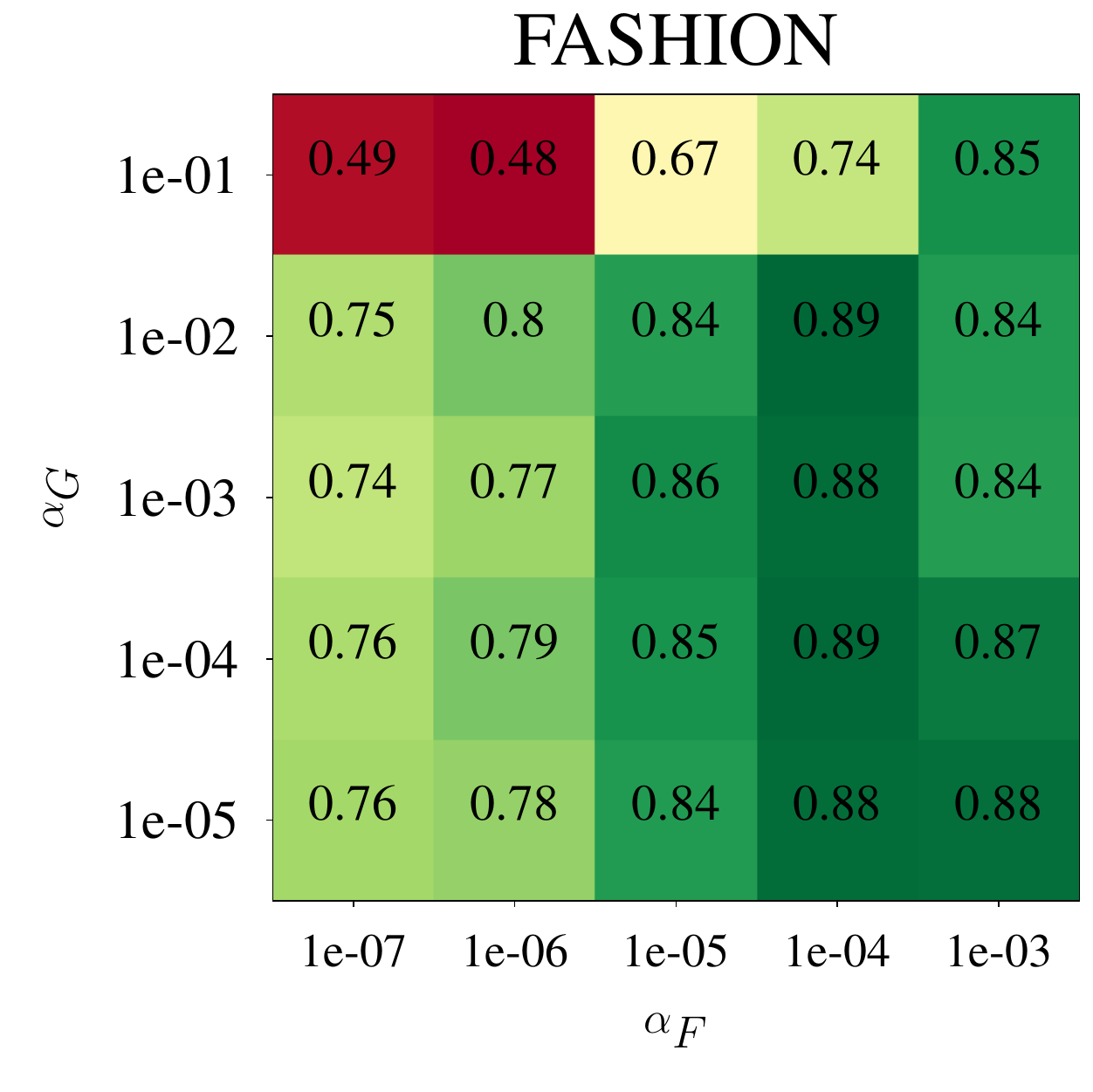}
	\end{subfigure}
	\begin{subfigure}{0.19\linewidth}
		\centering
		\includegraphics[width=\linewidth]{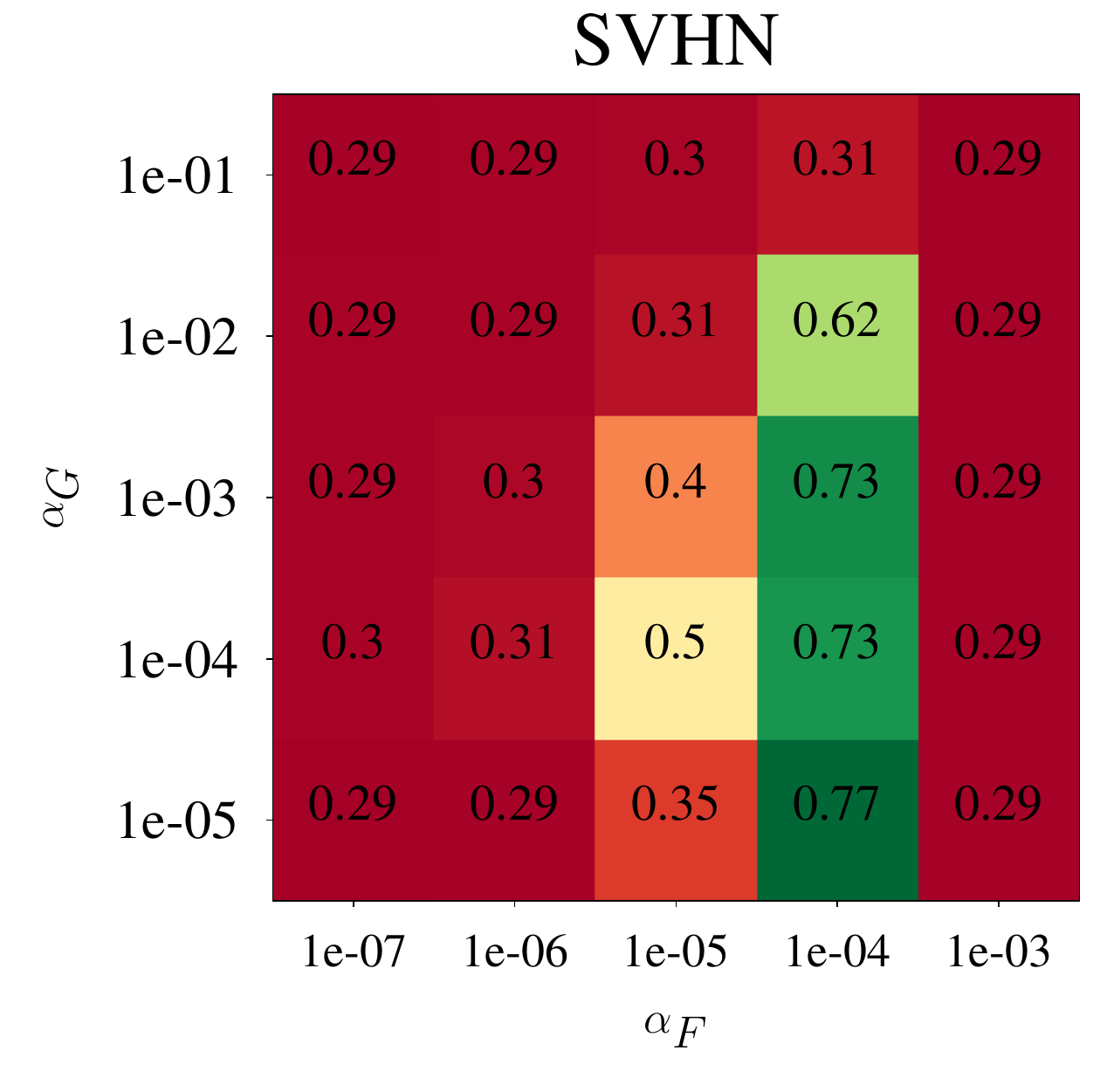}
	\end{subfigure}
	\begin{subfigure}{0.19\linewidth}
		\centering
		\includegraphics[width=\linewidth]{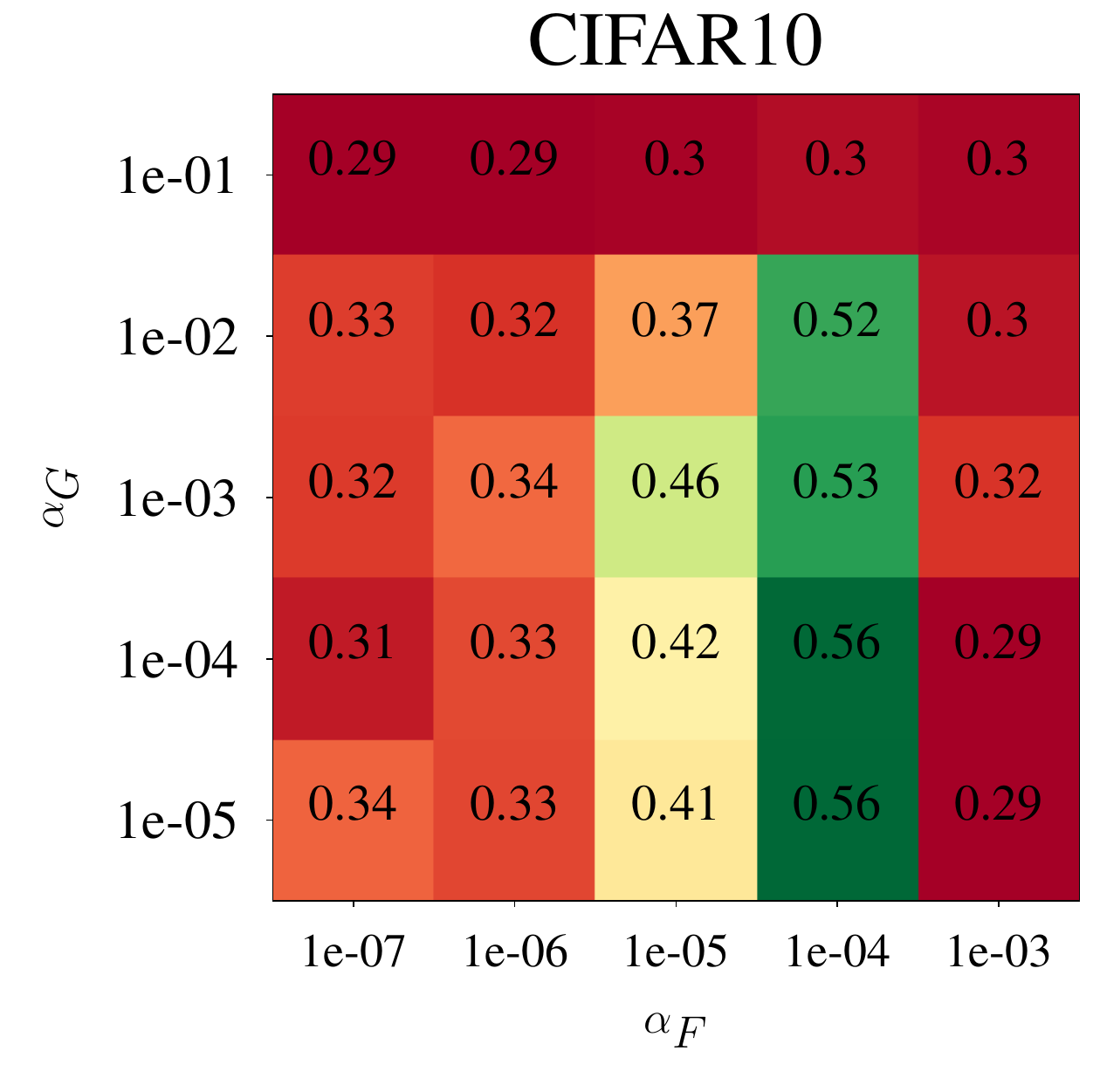}
	\end{subfigure}
	\begin{subfigure}{0.19\linewidth}
		\centering
		\includegraphics[width=\linewidth]{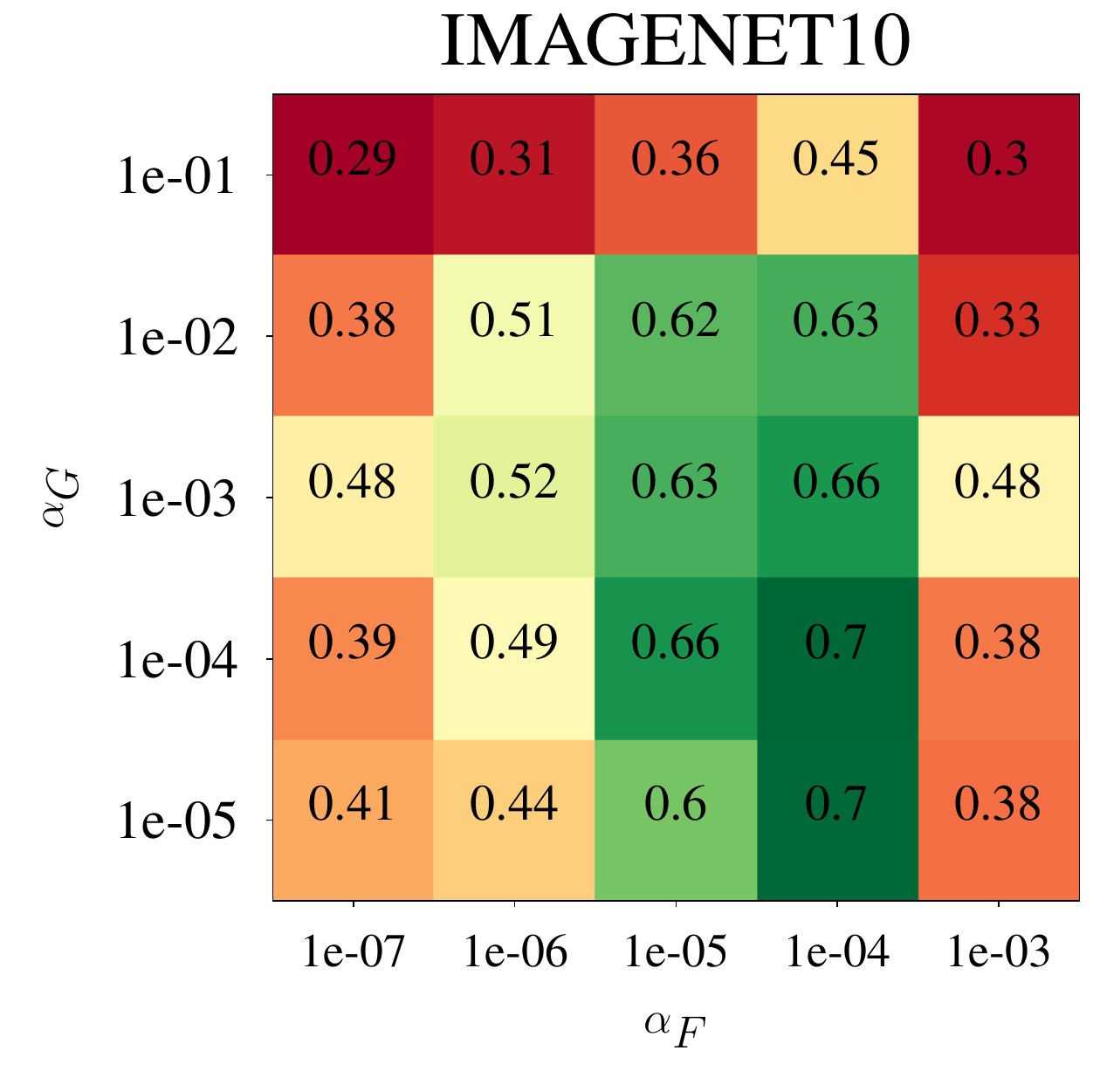}
	\end{subfigure}
	\caption[Average incremental accuracy for different learning rates]{Average incremental accuracy for different learning rates.}
	\label{fig:igmm-lrs}
\end{figure*}

\paragraph{Learning rates:} Analogously to the experiments for tightness, in Fig. \ref{fig:igmm-lrs} we presented the grid of results for different extractor (horizontal) and mixture (vertical) learning rates. The obtained results suggest that the former part is more important -- once the optimal rate is set (0.0001 for the given settings) tuning the latter seems less significant, although overall it should be set to a similar or slightly lower value.

\begin{figure*}[tb]
	\centering
	\begin{subfigure}{0.19\linewidth}
		\centering
		\includegraphics[width=\linewidth]{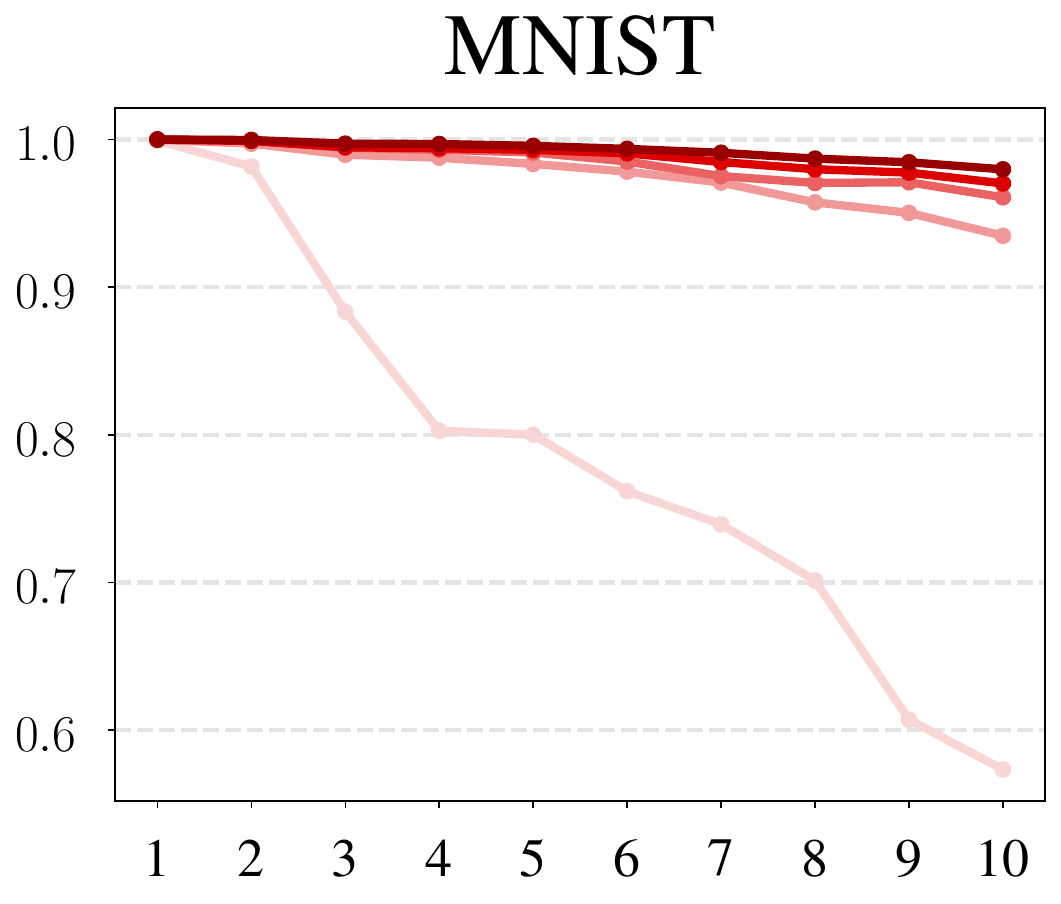}
	\end{subfigure}
	\begin{subfigure}{0.19\linewidth}
		\centering
		\includegraphics[width=\linewidth]{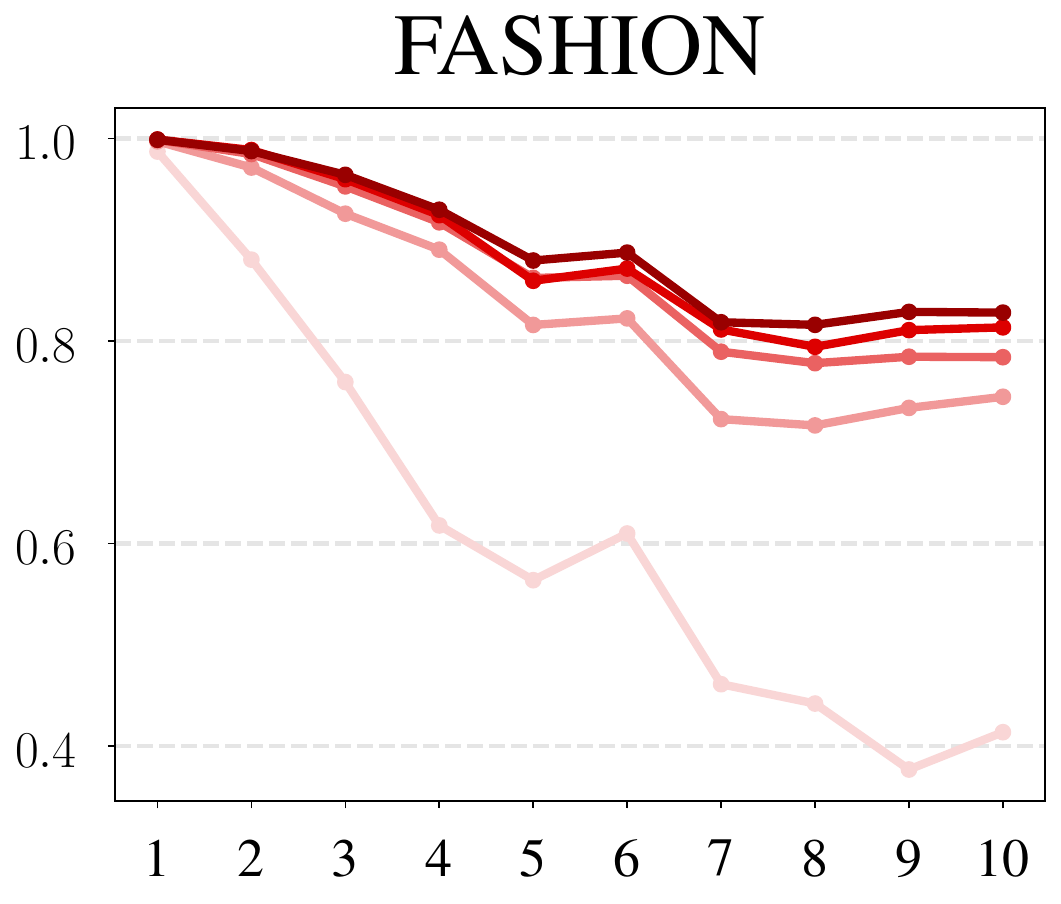}
	\end{subfigure}
	\begin{subfigure}{0.19\linewidth}
		\centering
		\includegraphics[width=\linewidth]{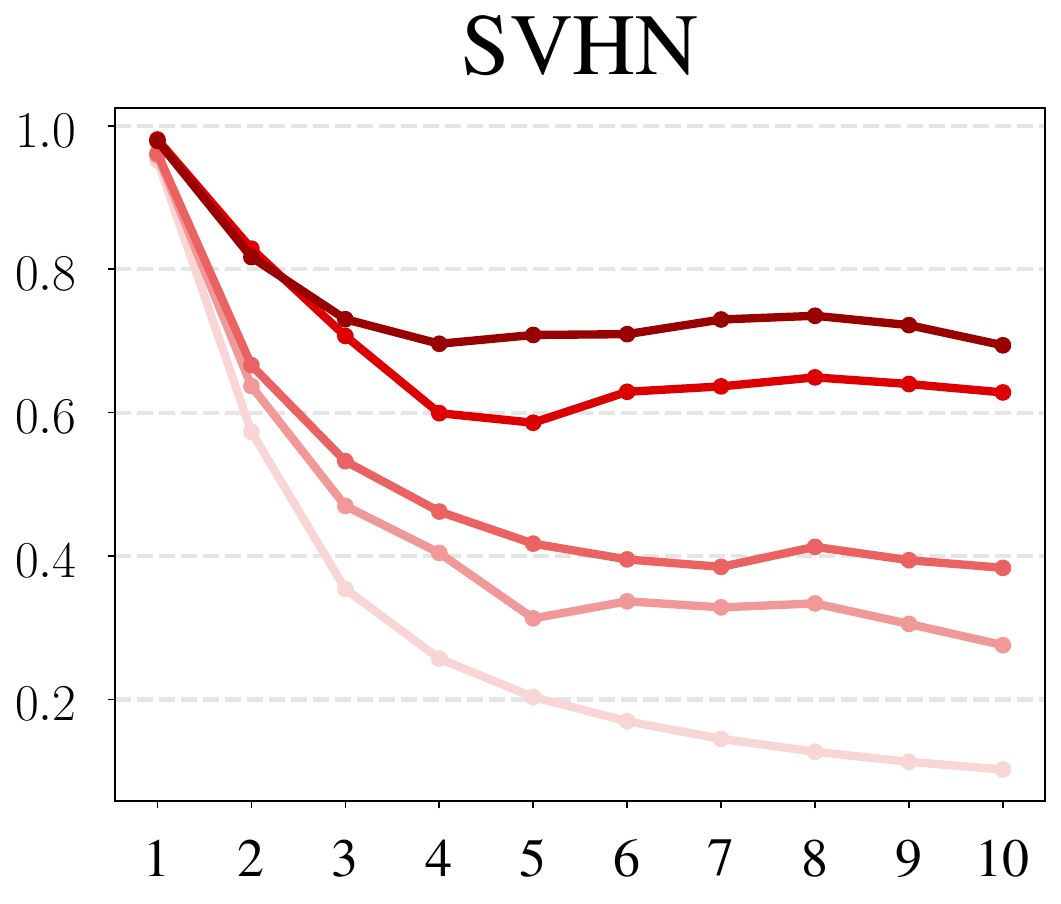}
	\end{subfigure}
	\begin{subfigure}{0.19\linewidth}
		\centering
		\includegraphics[width=\linewidth]{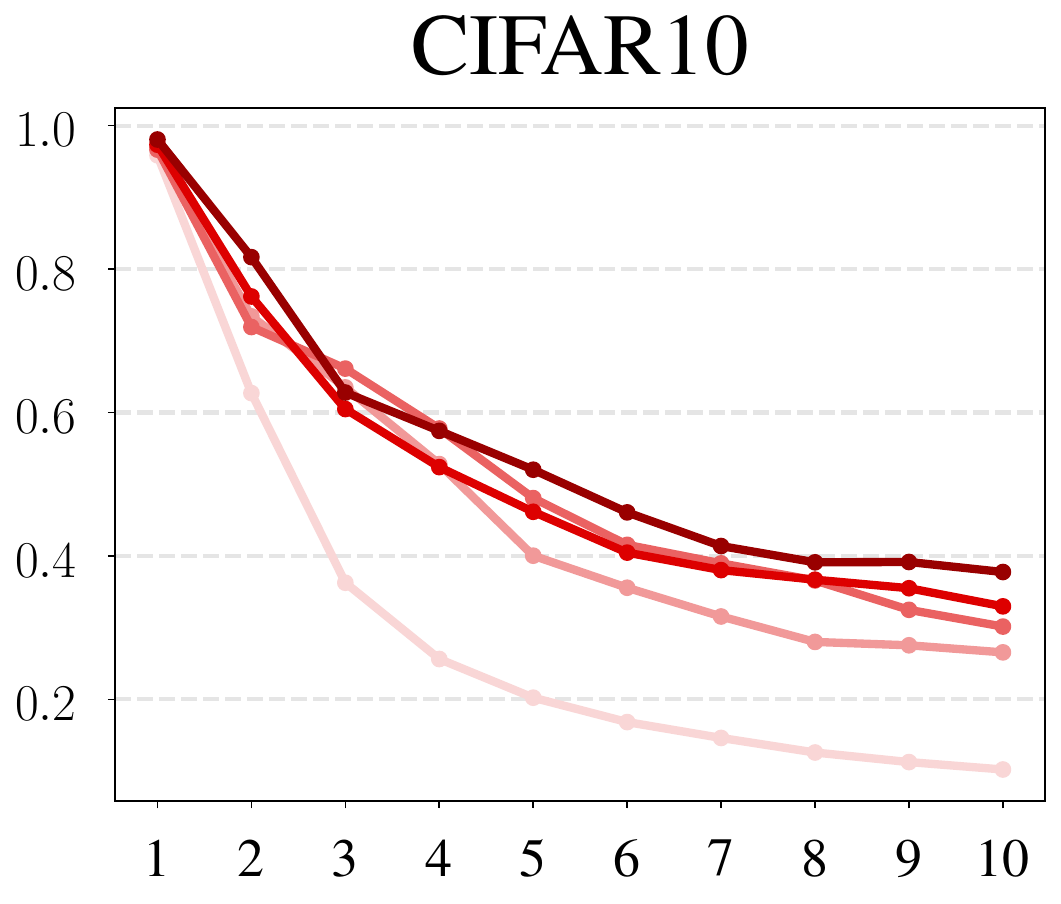}
	\end{subfigure}
	\begin{subfigure}{0.19\linewidth}
		\centering
		\includegraphics[width=\linewidth]{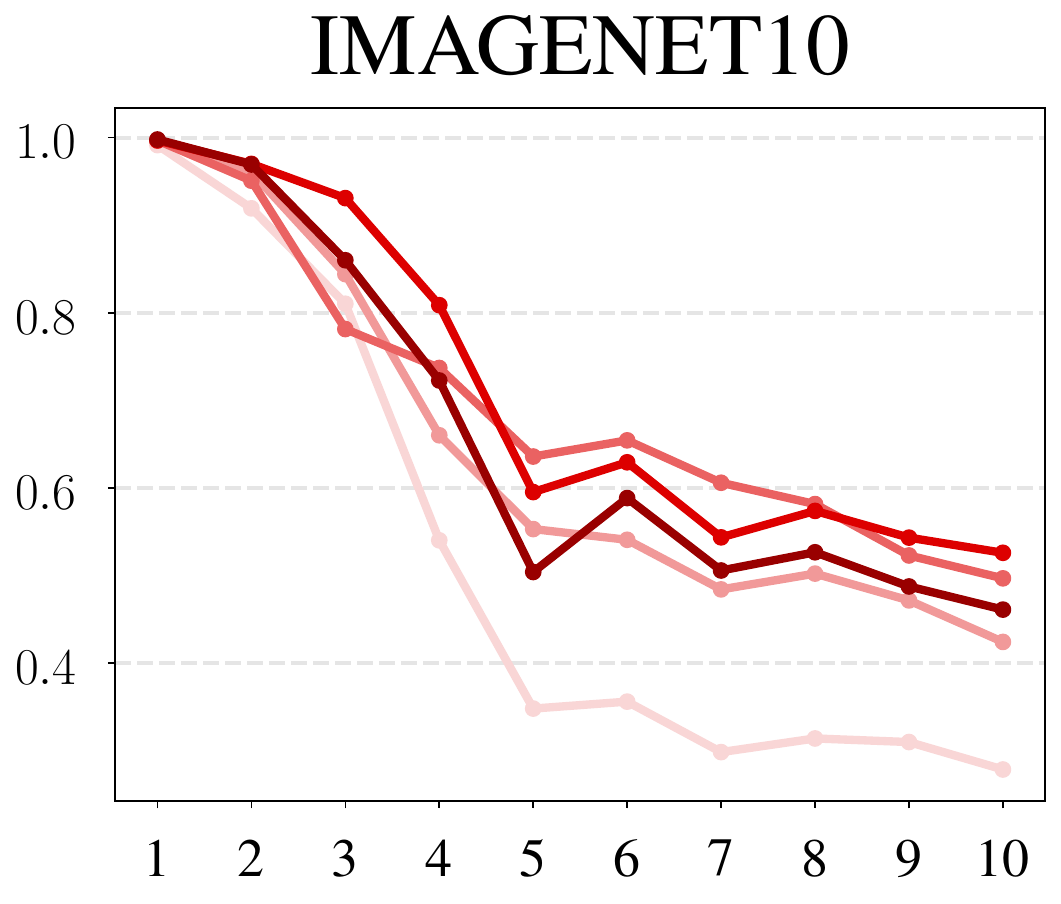}
	\end{subfigure}
	\caption[Incremental accuracy after each class batch for different sizes of the replay buffer]{Incremental accuracy after each class batch for different sizes of the replay buffer.}
	\label{fig:igmm-rbs}
\end{figure*}

\paragraph{Memory size:} Finally, if we look at the results of class-incremental learning using different memory sizes, given in Fig. \ref{fig:igmm-rbs}, we will see that MIX can effectively utilize larger buffers and that it seems to be quite memory-dependent, especially for SVHN where the difference between subsequent sizes ranged from 0.1 to 0.2. Still, the gap was much smaller for all of the remaining datasets. While this characteristic of the algorithm may be problematic (the fewer examples we need, the better), it is still valid that if we can use a pre-trained extractor, the whole model does not need to use the memory buffer at all.

\subsection{Lessons learned}

Based on the theoretical and empirical analysis presented for this work we can conclude the following.

\begin{enumerate}
	
	\item \textbf{Class-incremental learner.} Regardless of many combined challenges, it is possible to successfully hybridize the gradient-based mixture models on top of convolutional feature extractors, and use them in class-incremental end-to-end continual learning scenarios. The presented results show that MIX is capable of providing competitive results when compared with well-known incremental baselines.
	
	\item \textbf{Dedicated losses.} It has been shown that the training of the mixture models combined with dynamic feature extractors requires the inter-contrastive loss to effectively distinguish components of different classes from each other. In addition to that, to ensure diversity among same-class components and avoid degenerate solutions, such techniques as regionalization combined with the intra-contrastive loss are required. We showed that not only do the proposed approaches deliver what was intended, but also that they can translate into significant performance gains for more complex datasets. Finally, although the more generic high-level cross-entropy loss may provide good solutions in many cases, only the most advanced variant (MIX-MCR) delivers both high predictive performance and high quality of generated mixture models, which may be important from the perspective of interpretability or potential Gaussian-based extensions.
	
	\item \textbf{Effective tightness.} The tightness bound plays a crucial role in stabilizing the mixture learning procedure. Setting the optimal values of inter- and intra-tightness leads to striking a balance between pushing different components from each other and actually fitting them to the data. Intuitively, the inter-tightness prefers slightly lower values than intra-tightness.
	
	\item \textbf{Recommended configurations.} By analyzing other different hyperparameter settings and combinations of our methods we could observe that: (i) the CE loss works much better with the softmax classification method, while MC and MCR should be combined with the max-component approach, (ii) different numbers of components may be required for different data and different learning rates may also affect the optimal number, (iii) maintaining only the diagonal of the covariance matrices leads to more stable optimization and better results, (iv) the learning rate for the feature extractor dominates over the one for the mixture model, and that (v) MIX is quite memory-dependent in general end-to-end scenarios.
	
	\item \textbf{Memory-free scenarios.} At the same time, MIX is capable of learning without a memory buffer if we use a fixed pre-trained extractor and disable the contrastive loss that is not needed in this case. Our method stands out as the best model for such class-incremental scenarios which can be very important if there are any data privacy concerns or strict memory limits.
	
\end{enumerate}

\end{document}